\documentclass{article} 

\pdfoutput=1

\PassOptionsToPackage{numbers, compress, sort}{natbib}

\usepackage{times}  
\usepackage{helvet}  
\usepackage{courier}  
\usepackage{url}  
\usepackage{graphicx}  
\usepackage{natbib}
\usepackage[utf8]{inputenc} 
\usepackage[T1]{fontenc}    
\usepackage{hyperref}       
\usepackage{url}            
\usepackage{booktabs}       
\usepackage{amsfonts}       
\usepackage{amsmath}  
\usepackage{amsthm}
\usepackage{amssymb}  
\usepackage{nicefrac}       
\usepackage{microtype}      
\usepackage{bbm}
\usepackage{xcolor}         
\usepackage{adjustbox}
\usepackage{geometry}
\usepackage{pdflscape}

\usepackage{tikz}
\usetikzlibrary{arrows}
\usetikzlibrary{positioning}
\tikzset{
  treenode/.style = {align=center, inner sep=0pt, text centered,
    font=\sffamily},
  arn_n/.style = {treenode, circle, black, font=\sffamily\bfseries, draw=black,
    fill=white, text width=1.5em},
  arn_r/.style = {treenode, circle, black, font=\sffamily\bfseries, draw=black,
    fill=white, text width=1.0em},
  arn_x/.style = {treenode, rectangle, draw=black,
    minimum width=0.5em, minimum height=0.5em}
}
\usepackage{thmtools,thm-restate}
\usepackage[skip=6pt plus 2pt, indent=12pt]{parskip}
\usepackage{authblk}
\usepackage[linesnumbered,ruled,noend]{algorithm2e}
\usepackage{algorithmic}
\usepackage{mathtools}

\usepackage{floatrow}
\newfloatcommand{capbtabbox}{table}[][\FBwidth]

\usepackage{basic}
\hypersetup{
  pdffitwindow=true,
  pdfstartview={FitH},
  pdfnewwindow=true,
  colorlinks,
  linktocpage=true,
  linkcolor=blue,
  urlcolor=blue,
  citecolor=blue
}

\theoremstyle{plain}
\newtheorem{theorem}{Theorem}[section]

\theoremstyle{definition}
\newtheorem{definition}[theorem]{Definition}

\theoremstyle{remark}

\usepackage{xspace}

\newcommand{\X}[0]{\mathcal{X}}

\newcommand{\D}[0]{\mathcal{D}}

\newcommand{\R}[0]{\mathbb{R}}

\newif\ifsinglecolumn

\newcommand{\F}[0]{\mathcal{F}}

\newcommand{\A}[0]{\mathcal{A}}

\newcommand{\Leval}[1]{L_{\text{eval}, #1}}

\newcommand{\name}[0]{\textsc{Skill-it}\xspace}

\newcommand{\strainset}[0]{\mathcal{S}_{\text{train}}}
\newcommand{\sevalset}[0]{\mathcal{S}_{\text{eval}}}
\newcommand{\sprereq}[0]{\mathcal{S}_{\text{prereq}}}

\newcommand{\strain}[1]{s_{\text{train}, #1}}
\newcommand{\seval}[1]{s_{\text{eval}, #1}}

\usepackage{etoolbox}

\makeatletter
\newcommand{\changeoperator}[1]{%
  \csletcs{#1@saved}{#1@}%
  \csdef{#1@}{\changed@operator{#1}}%
}
\newcommand{\changed@operator}[1]{%
  \mathop{%
    \mathchoice{\textstyle\csuse{#1@saved}}
               {\csuse{#1@saved}}
               {\csuse{#1@saved}}
               {\csuse{#1@saved}}%
  }%
}
\makeatother

\let\oldnl\nl
\newcommand{\nonl}{\renewcommand{\nl}{\let\nl\oldnl}}

\title{Skill-it! A Data-Driven Skills Framework for Understanding and Training Language Models}

\author[1]{Mayee~F.~Chen\thanks{Corresponding author: \url{mfchen@cs.stanford.edu}.}}
\author[2]{Nicholas~Roberts}
\author[1]{Kush~Bhatia}
\author[3]{Jue~Wang}
\author[3, 4]{Ce~Zhang}
\author[2]{Frederic~Sala}
\author[1]{Christopher~R\'e}

\affil[1]{Department of Computer Science, Stanford University}
\affil[2]{Department of Computer Sciences, University of Wisconsin-Madison}
\affil[3]{Together AI}
\affil[4]{Department of Computer Science, University of Chicago}

\begin{document}

\maketitle

\begin{abstract}
The quality of training data impacts the performance of pre-trained large language models (LMs). Given a fixed budget of tokens, we study how to best select data that leads to good downstream model performance across tasks.
We develop a new framework based on a simple hypothesis: just as humans acquire interdependent skills in a deliberate order, language models also follow a natural order when learning a set of skills from their training data. If such an order exists, it can be utilized for improved understanding of LMs and for data-efficient training. Using this intuition, our framework formalizes the notion of a skill and of an ordered set of skills in terms of the associated data.
First, using both synthetic and real data, we demonstrate that these ordered skill sets exist, and that their existence enables more advanced skills to be learned with less data when we train on their prerequisite skills.
Second, using our proposed framework, we introduce an online data sampling algorithm, \name, over mixtures of skills for both continual pre-training and fine-tuning regimes, where the objective is to efficiently learn multiple skills in the former and an individual skill in the latter. 
On the LEGO synthetic in the continual pre-training setting, \name obtains $36.5$ points higher accuracy than random sampling. On the Natural Instructions dataset in the fine-tuning setting, \name reduces the validation loss on the target skill by $13.6$\% versus training on data associated with the target skill itself. 
We apply our skills framework on the recent RedPajama dataset to continually pre-train a 3B-parameter LM, achieving higher accuracy on the LM Evaluation Harness with 1B tokens than the baseline approach of sampling uniformly over data sources with 3B tokens.
\end{abstract}

\section{Introduction}

Large language models (LMs) exhibit remarkable capabilities, including producing creative content~\cite{stevenson2022putting},  writing source code~\cite{chen2021evaluating}, or chatting with users~\cite{brown2020language}.
A key ingredient in enabling models to perform such tasks is the data on which the models are trained~\cite{palm2techreport, gururangan2020dont, touvron2023llama}. 
A natural way to unlock particular capabilities is to improve this training data. However, it is unclear how to select data from a large corpus for these capabilities given a fixed budget of training tokens, as 
data selection methods for current state-of-the-art LMs mostly rely on heuristics for filtering and mixing together different datasets~\cite{lee2022deduplicating, touvron2023llama}. 
We lack a formal framework for capturing how data influences the model's capabilities and how to utilize this data effectively for improving LM performance. 

\begin{figure*}[t]
    \centering
    \includegraphics[width=0.85\textwidth]{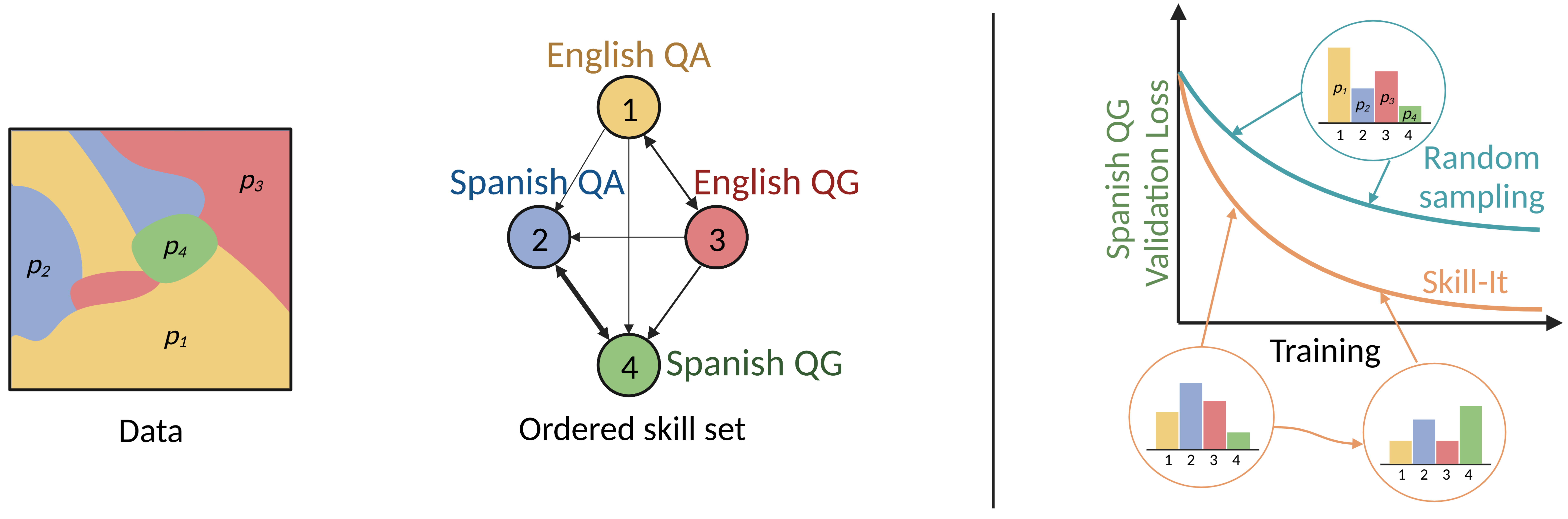}
    \caption{Inspired by how humans acquire knowledge, we hypothesize that LMs best learn skills in a particular order and that this can help improve our understanding and training of LMs. We show that these ordered skill sets exist in real data, which enables skills to be learned with less data given that we train on their prerequisite skills. We then propose \name, an online data selection algorithm that learns skills quickly by exploiting their ordering.}
    \label{fig:banner}
\end{figure*}

To develop such a framework, we take inspiration from how humans acquire knowledge. 
A classic idea in education literature is the concept of \textit{skills} that form a learning hierarchy~\cite{white1973research}. 
For example, one study found that students learned mathematical and scientific skills most quickly when these skills were presented in a particular order~\cite{gagne1962acquisition}.
We seek to understand the extent that similar skill-based orderings characterize LM training.
Such orderings, if they exist, may provide a better understanding of LMs as well as a mechanism for data-efficient training. 
For instance, to train an LM for Spanish question generation, we wish to know if training first on related but simpler tasks, such as Spanish grammar and English question generation, helps.

We study if the idea of skill orderings can help us build a framework that relates data to LM training and behavior.
This requires addressing two challenges revolving around the connection between skills and data. 
First, in order to show that there exist sets of skills that the LM learns most efficiently in some particular order, \textit{an operational definition of LM skill and skill ordering must be developed and validated on data}.
In initial experiments, we investigated if semantic groupings of data, such as metadata attributes or embedding clusters, were sufficient to represent a skill and characterize how models learn. For instance, we partitioned the Alpaca dataset~\cite{alpaca} by instruction type---a technique used to capture dataset diversity~\cite{wang2022selfinstruct}---but we found that sampling based on instruction types and random sampling resulted in similar model performance, suggesting that not just any existing notion of data groups can characterize skills.

Second, \textit{these definitions of skills must be used to construct sampling distributions to actually improve model training.} 
To develop criteria for a data selection algorithm that learns skills efficiently, we identify challenges that naive selection approaches face. The standard approach of random uniform sampling over data fails to learn skills optimally due to not accounting for skill imbalance and ordering. Skills can be distributed unevenly in the data, with more complex skills being rare---for instance, Spanish and question generation (QG) are $5\%$ and $4\%$ of the Natural Instructions dataset~\cite{wang2022supernaturalinstructions}, respectively, but Spanish QG is only $0.2\%$.
Random sampling also provides no mechanism for taking into account a particular training order and dependency structure on skills. More sophisticated techniques like curriculum learning account for sample-level ordering, but not skills or their dependencies.
Our goal framework must account for these issues of imbalance and ordering.

\textbf{Skill-based framework}  
We define a \emph{skill} as a unit of behavior that a model can learn using an associated slice of data (Definition~\ref{def:skill}). An \emph{ordered skill set} is a collection of skills with a directed \textit{skills graph} that is neither complete nor empty, where an edge from a prerequisite skill to a skill exists if the amount of training it takes to learn the skill
can be reduced if the prerequisite skill is also learned (Definition~\ref{def:oss}, Figure~\ref{fig:banner} left, center).
We show that ordered skill sets exist in synthetic and real datasets using this operational definition. 
Interestingly, the existence of these ordered skill sets unveils 
that one can learn a skill quickly not by training solely on that skill, but on a mixture of that skill and prerequisite skills. For instance, in Figure~\ref{fig:ni_lego_examples} we observe that Spanish QG can be learned more efficiently when the model also learns English QG and Spanish---we can achieve $4\%$ lower validation loss than training on only Spanish QG over a fixed budget of overall training steps.

Next, given an ordered skill set to train on, we use our framework to propose methods for how to select data so that the LM learn skills faster: skill-stratified sampling and an online generalization, \name. 
We address the issue of unevenly distributed skills in datasets by proposing skill-stratified sampling, a simple approach that allows us to explicitly optimize for learning skills by uniformly sampling relevant skills (such as a target skill and its prerequisite skills in fine-tuning). 
Skill-stratified sampling uses the construction of the ordered skill set but is static, which does not incorporate the ordering as training proceeds and results in oversampling skills that may be already learned early on in training.
We address this issue by proposing an online data selection algorithm, \name, for selecting mixtures of training skills that allocates more weight towards learning skills that are not yet learned or towards prerequisite influential skills (Figure~\ref{fig:banner} right). \name is derived from an online optimization problem over the training skills for minimizing loss on a set of evaluation skills given a fixed budget of data and the skills graph. \name is inspired by online mirror descent and can be adapted for continual pre-training, fine-tuning, or out-of-domain evaluation depending on the relationship between the evaluation skill set and the training skill set.

We evaluate \name on synthetic and real datasets at two model scales, 125M and 1.3B parameters. 
For the continual pre-training setting, we show on the LEGO synthetic~\cite{zhang2022unveiling} that we obtain a $35.8$ point improvement in accuracy over randomly selecting training data and curriculum learning~\cite{bengio2009curriculum}. For the fine-tuning setting, we show that on the widely-used Natural Instructions dataset~\cite{naturalinstructions, wei2021finetuned}, our algorithm over a mixture of skills is able to achieve up to 13.6\% lower loss on that skill than solely training on that skill, given the same overall training budget. For the out-of-domain setting when our training skills do not align perfectly with evaluation skills, our algorithm is able to achieve the lowest loss on 11 out of 12 evaluation skills corresponding to task categories in the Natural Instructions test tasks dataset over random and skill-stratified sampling on the training data. We finally apply our framework to a case study on the recent RedPajama 1.2 trillion token dataset~\cite{redpajama}. We use the data mixture produced by \name to continually pre-train a 3B parameter model. We find that \name achieves higher accuracy with 1B tokens than uniform sampling over data sources with 3B tokens.
\section{Skills framework}

\vspace{-0.5em}

First, we propose definitions of skills and ordered skill sets in order to formalize our intuition around how models learn skills, and we demonstrate that not just any existing notion of data groups can characterize an ordered skill set in the dataset. Then, we demonstrate the existence of ordered skill sets on synthetic and real data, which show how viewing data through a skills-based framework can help with training and understanding model performance. Finally, we explore unsupervised skill recovery from data, finding that embedding-based approaches do not adequately recover synthetic skills.
\vspace{-0.5em}

\subsection{Definitions}\label{sec:def}

\begin{figure*}[t]
    \centering
    \includegraphics[height=1.4in]{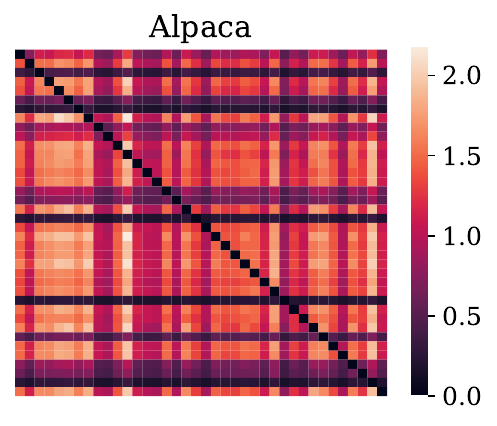}
    \includegraphics[height=1.4in]{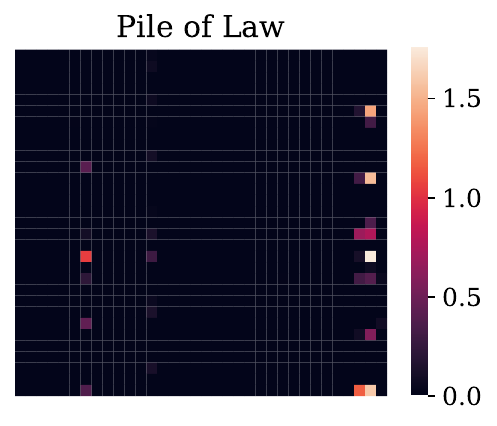}
    \includegraphics[height=1.4in]{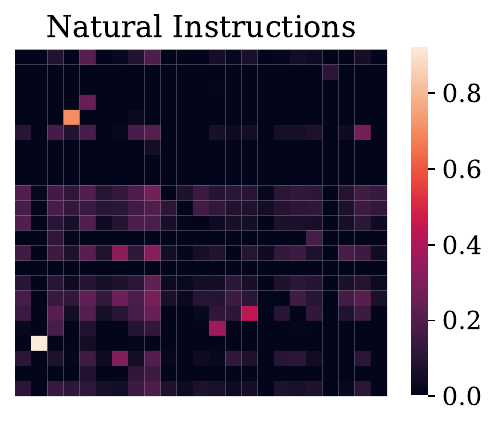}
    \caption{Heatmaps of adjacency matrices we compute for skill graphs for Alpaca, Pile of Law, and Natural Instructions. Negative elements and diagonals are thresholded to $0$ for clarity. See Appendix~\ref{supp:skill_graphs} for descriptions of how they were constructed and larger versions.}
    \label{fig:heatmaps}
\end{figure*}

We first present a definition of an individual skill. 
Let the input space of all possible text data be $\X$, where $x \in \X$ is an individual text sample that a next-token-prediction LM $f \in \F: \X \rightarrow \X $ is trained on.
We quantify learning via a metric $L: \F \times \X \rightarrow \R$, which maps from a model and evaluation data to a scalar quantity. In our setup, we use the cross-entropy validation loss applied over next-token predictions as our metric $L$.

\begin{definition}[Skill]\label{def:skill}
A \emph{skill} $s$ is a unit of behavior with associated data $\X_s \subseteq \X$ such that if $f$ is trained on an dataset $\D_s \subset \X_s$, then $f$ has improved metric $L$ on samples belonging to $\X_s \backslash \D_s$ on average.
\end{definition}

This definition of a skill is flexible---it simply means that given a training dataset associated with the skill, a model $f$ has an improved metric (e.g., decreasing validation loss) when evaluated on validation data associated with this skill. Under this definition, a skill could be a granular task, such as Spanish question generation for a subset of Wikipedia articles, or can be defined over a data source, such as next-token prediction of legal data from tax court rulings.
However, our next definition, the ordered skill set, has a more specific construction and provides a framework for how models learn across dependent skills.

\begin{definition}[Ordered skill set, skills graph] \label{def:oss}
An \emph{ordered skill set} for $f$ is a collection of skills $\mathcal{S} = \{s_1, \dots, s_k\}$ over which there is a directed \textit{skills graph} $G = (\mathcal{S}, E)$ on the skill set 
that is neither complete or empty, 
where $(s_i, s_j) \in E$ if the amount of data needed to learn $s_j$ when uniformly sampling from $\D_{s_i} \cup \D_{s_j}$ is no more than the amount of data needed when sampling only from $\D_{s_j}$. 
We equate learning a skill $s_j$ to $f$ attaining a certain value of $L$ or lower on average over $\X_{s_j} \backslash \D_{s_j}$.
\end{definition}
 This definition isolates complete and empty graphs as extrema that do not capture meaningful sets of skills. We discuss the three types of skill graphs---complete, empty, intermediate---and their implications for data selection. 
In particular, we discuss how several initial attempts of defining skills over datasets via semantic groupings resulted in the extrema cases (see Appendix~\ref{supp:skill_graphs} for full results):
 
\begin{itemize}[itemsep=0.05pt,topsep=0pt,leftmargin=*]
    \item The complete graph demonstrates that all skills influence each other. A random partition is an example of a skill set that yields a complete graph. This graph suggests that the best approach for learning any skill or set of skills is random sampling on the dataset. This is not a setting where we can gain much with skill-based sampling. For example, using instruction types as skills on the Alpaca dataset results in a nearly complete estimated skills graph ($97.4\%$ dense), and we find that stratified sampling on these skills only improves validation loss per skill by $0.007$ points over random sampling on average (Figure~\ref{fig:heatmaps} left), suggesting that utilizing skills does not improve model performance in this case.
    \item The empty graph demonstrates that each skill is independent. 
    This can occur if skills are too granular; for instance, learning Spanish math problems is unlikely to help with English poem generation. This graph suggests that the best approach for learning an individual skill is to train on the skill itself. We see that empty graphs exist in real data; in Figure~\ref{fig:heatmaps} (center), using data sources as skills on the Pile of Law~\cite{hendersonkrass2022pileoflaw} results in a nearly empty skills graph ($3.9\%$ dense).
    \item Graphs that are neither empty nor complete thus suggest a nontrivial order of how skill influence each other. \textit{This is the setting in which we expect that identifying skills and exploiting their ordering will help the most}. In Figure~\ref{fig:heatmaps} right, we use task categories, which capture broader reasoning patterns, as skills on Natural Instructions and find that the estimated graph has intermediate density ($42.7\%$ dense). We show concrete examples of how skills can be learned more efficiently on Natural Instructions in Section~\ref{sec:skill_examples}.
\end{itemize}

While these intuitive groupings result in ordered skill sets on some datasets (e.g., task categories on NI), this is not always the case (e.g., instruction types on Alpaca and sources on Pile of Law).
Even though these groupings capture some notion of diversity in the dataset, our findings suggest that not just any semantic grouping induces an ordered skill set.
We now empirically demonstrate that our definition of ordered skill sets aligns with how models learn and can be exploited for more data-efficient training.
\vspace{-0.5em}

\subsection{Examples of skills and ordered skill sets} \label{sec:skill_examples}

\begin{figure*}[t]
    \centering
    \includegraphics[height=1.4in]{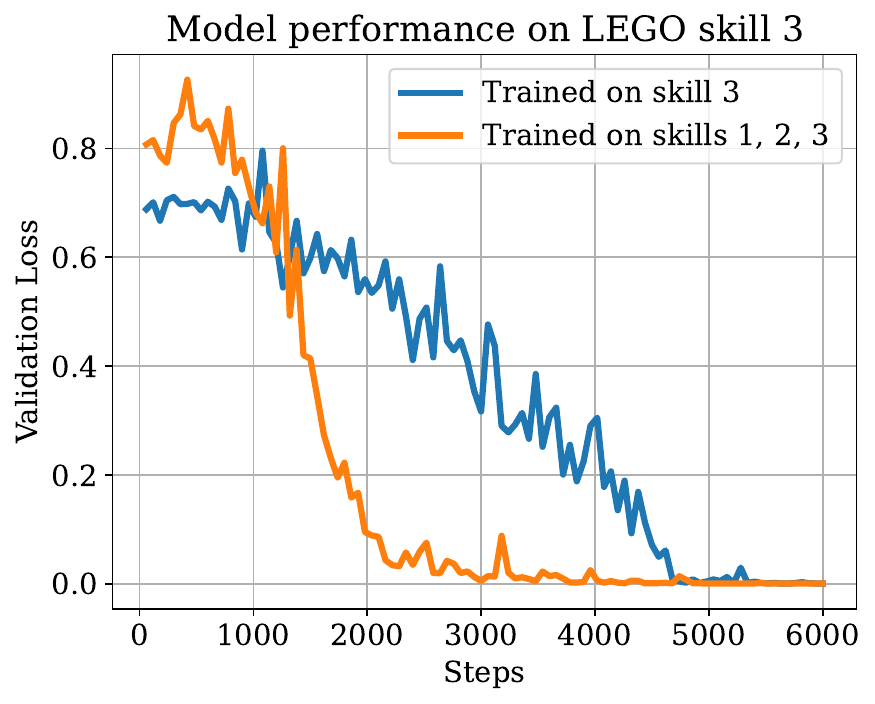}
    \includegraphics[height=1.4in]{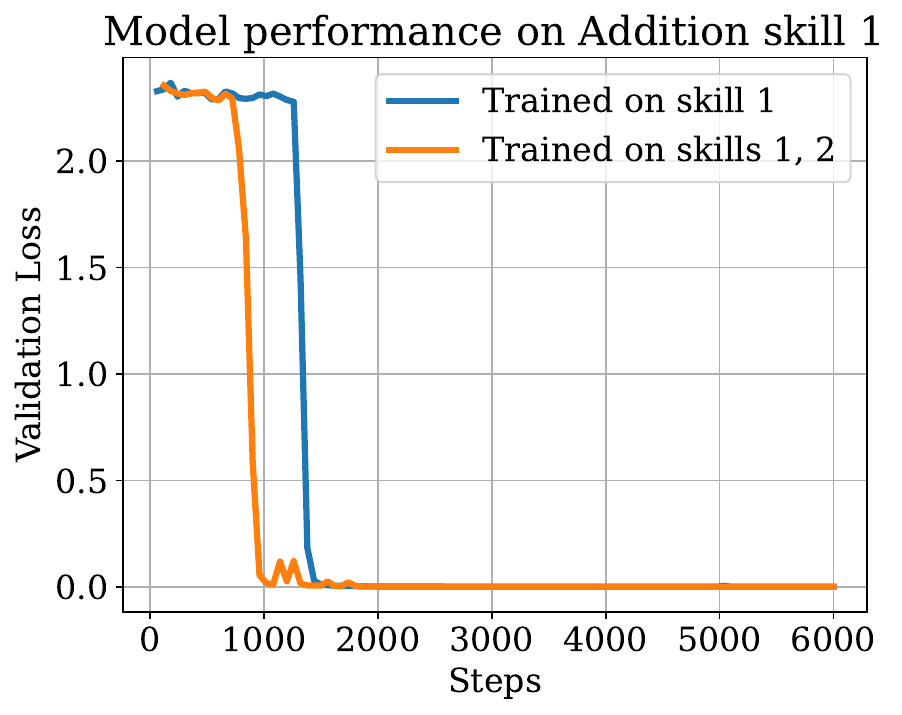}
    \includegraphics[height=1.4in]{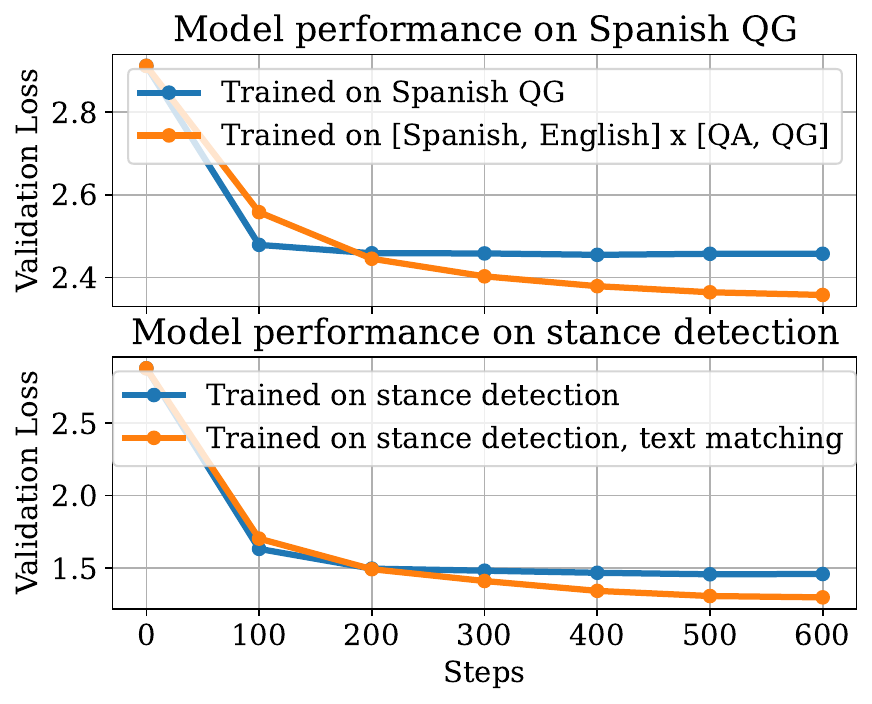}
    \caption{On the LEGO synthetic, 3-digit addition, and Natural Instructions, we identify examples of ordered skill sets in which training on a mixture of skills helps learn an individual skill faster than just training on that skill itself, given a fixed training budget.}
    \label{fig:ni_lego_examples}
\end{figure*}

We provide examples of ordered skill sets on the LEGO synthetic dataset, an addition synthetic dataset, and subsets of the Natural Instructions dataset. On these datasets, we find that certain skills are better learned when trained along with their prerequisite skills rather than in isolation.

\textbf{LEGO skills} \;
The LEGO synthetic, first introduced in~\cite{zhang2022unveiling}, can evaluate a model's ability to follow a chain of reasoning. In this synthetic, the letters of the alphabet, $\A$, are variables each with some binary label in $\{0, 1\}$. An individual sample consists of $k$ clauses for some fixed $k$ across the dataset, each of the form $a = gx$ where $a, x \in \A$ and $g$ is either a negation (``not'') or assertion (``val''), e.g. we assign $a$ to the value of $x$, or we assign $a$ to the opposite label. At the end of the sentence, we prompt the model for what the value of one of these variables is. Two samples $x \in \X$ are given below for $k = 5$:

\begin{quote}
    Input: b = not y, r = val 1, m = val b, q = val m, y = not r. Output: b = 1.  
    
    Input: c = val x, p = val f, x = val k, f = not c, k = val 0. Output: k = 0.
\end{quote}

These samples each correspond to a chain of reasoning; for instance the first sample has the chain $r, y, b, m, q$, where knowing $q$'s label requires the most reasoning steps. 
We define the $i$th skill $s_i$ as the model's ability to know the $i$th variable of the chain. From our example above, the first sample belongs to $\X_{s_3}$ and the second sample belongs to $\X_{s_1}$. To demonstrate the existence of ordered skill sets, we continually pre-train the 125M parameter GPT-Neo model~\cite{gao2020pile, gpt-neo} over various mixtures of LEGO skills with $k = 5$. In Figure~\ref{fig:ni_lego_examples} (left), we find that 
in 35.9\% fewer training steps, training on a balanced mixture of $\X_{s_1}, \X_{s_2}$, and $\X_{s_3}$ resulted in the same validation loss of $0.01$ as training solely on $\X_{s_3}$.
This suggests that $s_1, s_2$ helped unlock performance on $s_3$ and that there exist edges from $s_1$ or $s_2$ to $s_3$ in the skill graph.
Additional observations are available in Appendix~\ref{supp:lego_exp}, where we examine other edges as well as more complex reasoning chains, and the full skills graph corresponding to the ordered skill set for LEGO with $k=5$ is in Figure~\ref{fig:lego_heatmap}.

\textbf{Addition skills} \;
We consider a variant of a synthetic 5-digit addition dataset analyzed in \cite{nanda2023progress}.
We show the existence of ordered skill sets for a simplified 3-digit addition dataset where we treat each digit prediction as a skill---the outputs, in this case, are the integers $\{0, 1, ..., 9\}$. 
Examples are of the following form:
\begin{quote}
    Input: A = 1 0 6 + 0 7 1 , A 0 = ? Output: 7 $\quad$ Input: A = 6 0 6 + 8 7 9 , A 2 = ? Output: 4 
\end{quote}
where `A 0' refers to the ones digit of the output ($s_1$) and `A 2' refers to the hundreds digit ($s_3$).
In Figure~\ref{fig:ni_lego_examples} (center), we find that in $32\%$ fewer training steps, training on a balanced mixture of $\X_{s_1}$, and $\X_{s_2}$ resulted in the same validation loss of $0.01$ as training solely on $\X_{s_1}$.
That is, the ones digit addition skill can be improved by simultaneously learning the tens digit addition skill, even though the former should not require information from the latter---this is in line with observations from prior work that models do not always learn the ones digit addition first~\cite{nanda2023progress}. The full skills graph corresponding to the ordered skill set over 3-digit addition is in Figure~\ref{fig:addition_heatmap}.

\textbf{Natural Instructions (NI) skills} \;
We show that ordered skill sets exist in NI~\cite{wang2022supernaturalinstructions} when we treat task categories as skills. 
\begin{itemize}[itemsep=0.05pt,topsep=0pt,leftmargin=*]
    \item In Figure~\ref{fig:ni_lego_examples} (top right), we show that ordered skill sets exist over crosslingual task categories. Training on Spanish question generation (QG) along with equal parts of English QG, Spanish question answering (QA), and English QA results in $4.1\%$ lower validation loss than training only on Spanish QG. Remarkably, the former only uses $25\%$ of the latter's Spanish QG data. 
    This suggests that there are edges from Spanish QA, English QA, and English QG to Spanish QG.
    \item In Figure~\ref{fig:ni_lego_examples} (bottom right), we see that training on the task category Text Matching along with Stance Detection helps decrease the loss on Stance Detection by $11\%$. This suggests that these categories, which both involve understanding the relationship between two input texts, share an edge.
\end{itemize}

The full skills graphs corresponding to the ordered skill sets over these task categories are in Figure~\ref{fig:ni_ft_heatmap}. While equating task categories to skills may be noisy, these examples suggest that there is signal within real data that suggests that ordered skill sets can improve data efficiency. 

\vspace{-0.5em}
\subsection{Skill recovery} \label{sec:skill_recovery}

\vspace{-0.5em}
A final component of characterizing skills is unsupervised recovery of ordered skill sets. We consider embedding-based clustering approaches and a loss-based clustering approach for recovering LEGO skills. When clustering data using various trained and pre-trained embeddings, we find that they were unable to achieve above $39\%$ accuracy on LEGO. Instead, we find that taking $10$ random training runs and clustering data by their \textit{loss} per timestep per run recovers the skills with $61\%$ accuracy (Table~\ref{tab:skill_recovery}). The intuition behind this method is that the validation losses on points from the same skill have similar trajectories as models learn. We discuss this approach more in Appendix~\ref{supp:skill_clustering}. 

\vspace{-0.5em}

\section{Skills-based data selection}

\vspace{-0.5em}

Now that we have established the existence of ordered skill sets, we discuss how to use them for data selection.
We state the data selection problem for learning across skills in Section~\ref{sec:problem_statement}.
We discuss how to learn the skills graph that will be exploited in our data selection methods in Section~\ref{sec:graph_learning}.
We then introduce two sampling methods that utilize the graph, a simple skill-stratified sampling method and the online sampling method \name, in Section~\ref{sec:sampling}.

\subsection{Problem statement} \label{sec:problem_statement}

We are given an ordered training skill set $\strainset = \{\strain{1}, \dots, \strain{k}\}$ on the training data, each with associated support set $\X_{\strain{1}}, \dots \X_{\strain{k}}$, and an ordered evaluation skill set $\sevalset = \{\seval{1}, \dots, \seval{m}\}$ of $m$ evaluation skills on a separate evaluation dataset.
We aim to select $n$ samples from $\strainset$ via a mixture of training skills, $p \in \Delta^{k-1}$, to achieve three goals depending on how $\sevalset$ is constructed:
\begin{itemize}[itemsep=0.05pt,topsep=0pt,leftmargin=*]
    \item \textbf{Continual pre-training}: when $\sevalset = \strainset$, our goal is select a mixture of training skills to learn all of them.
    \item \textbf{Fine-tuning}: when $\sevalset \subset \strainset$, our goal is to select a mixture of training skills to learn an individual target skill or subset of these skills. 
    \item \textbf{Out-of-domain}: when $\sevalset \cap \strainset = \emptyset$, our goal is to select a mixture of training skills to learn a disjoint set of evaluation skills we cannot train on. This can arise when we have a separate downstream validation dataset or the skills identified in the training dataset are noisy.

\end{itemize}
Furthermore, we have a skills graph $G = (\strainset \cup \sevalset, E)$, where $E \subseteq \strainset \times \sevalset$ and $A \in \R^{k \times m}$ is a weighted adjacency submatrix, where $A_{ij}$ describes the strength of the edge from $\strain{i}$ to $\seval{j}$. In Table~\ref{tab:settings}, we summarize how the three different settings are constructed and how $A$ varies across them. Next, we discuss how $A$ can be estimated from the data. 

\begin{table*}
\footnotesize
\caption{Summary of three settings---continual pre-training, fine-tuning, and out-of-domain. These settings are determined by how $\sevalset$ is defined and result in different skills graphs used for our sampling methods.}
\begin{tabular}{c|c|c}
    \toprule
    Setting & $\sevalset$ & Skills graph \\
    \midrule 
    Continual pre-training & $\sevalset = \strainset$ & $A \in \R^{k \times k}$, edges among all $\strainset$  \\
    Fine-tuning & $\sevalset \subset \strainset$ & $A \in \R^{k \times m}$, edges from all training skills to target skill subset \\
    Out-of-domain & $\sevalset \cap \strainset = \emptyset$ & $A \in \R^{k \times m}$, edges from all training skills to separate evaluation skill set \\
    \bottomrule
\end{tabular}
\label{tab:settings}
\end{table*}

\subsection{Skills graph learning} \label{sec:graph_learning}
The skills graph is important for determining how to sample from the ordered skill set for training efficiently.
We present two approaches for learning the skills graph---brute-force and linear approximation. Algorithms are provided in Appendix~\ref{supp:graph_learning}. By definition~\ref{def:oss}, the brute-force way of identifying edges involves fixing an overall training budget of $H$ steps and 1) training and evaluating the model on each $s_i$ and 2) training the model on each pair of $(s_i, s_j)$ and evaluating on $s_i$ and $s_j$. If the loss on $s_j$ when trained on both $s_i$ and $s_j$ is lower, there exists an edge from $s_i$ to $s_j$. This approach has runtime $\mathcal{O}(H k^2)$, which is feasible for small $k$. 
When $k$ is large, we can approximate this approach in linear time by training on each $s_i$ for $h < H$ steps and setting $A_{ij} > 0$ if the loss on $s_j$ decreases over $h$ steps for a runtime of $\mathcal{O}(hk)$. 
This linear approach is necessary in the out-of-domain setting when $\sevalset$ and $\strainset$ are disjoint, as we do not train on data associated with $\sevalset$. In addition, both graph learning approaches can be performed on a smaller model, and the learned graph can be used for data selection for training a larger model (Appendix~\ref{supp:larger_model}).

\subsection{Skills graph-aware sampling}\label{sec:sampling}

We present two approaches for sampling over the mixture of training skills according to the skills graph: skill-stratified sampling, which samples uniformly over relevant training skills according to $A$, and \name, which is an online generalization that incorporates knowledge of how skills are being learned throughout training.

\subsubsection{Skill-stratified sampling}\label{sec:skill_stratified}

A straightforward sampling approach is to discard training skills that do not benefit the evaluation skills and sample uniformly over the set of relevant training skills, which we call \textit{skill-stratified sampling}. 
For continual pre-training, the relevant skills are the entire training skill set; for each $\strain{i} \in \strainset$, $\Pr(\strain{i}) = \frac{1}{k}$. 
This enables each skill to have sufficient training data.
For fine-tuning, the relevant skills are the target skills and prerequisite skills, which can be identified via positive entries of the $i$th column of $A$ with $\sprereq = \{\strain{i}: \exists \; \seval{j} \; \text{s.t.} \; A_{ij} > 0 \}$. 
We then set $\Pr(s) = \frac{1}{|\sprereq \cup \sevalset|}$ for $s \in \sprereq \cup \sevalset$. 
For the out-of-domain setting, skill-stratified sampling is over the set of prerequisite skills. 
For each $s \in \sprereq$, we set $\Pr(s) = \frac{1}{|\sprereq|}$. 
Next, we propose our online algorithm that exploits the graph dynamically for more efficient training.

\subsubsection{\name online data selection algorithm} \label{sec:alg}

\begin{algorithm}[tb]
   \caption{\name Online Data Selection Algorithm}
\begin{algorithmic}[1]
   \STATE {\bfseries Input:} Ordered training skill set $\strainset$, ordered evaluation skill set $\sevalset$.  Learning rate $\eta$, $T$ rounds, $n$ samples, $H$ training steps per run for graph learning, model $f_1$, window parameter $w$.
   \STATE $A \gets$ \textsc{LearnGraph}$(\strainset, \sevalset, H, f_1)$ (Alg.~\ref{alg:bruteforce_graph},~\ref{alg:approximate_graph}).
   \STATE Initialize $p_1^i = \exp(\eta \sum_{j = 1}^m A_{ij})$ for all $i \in [k]$, the softmax over $A$.
   \FOR{$t = 1, \dots, T-1$}
    \STATE Observe losses $\Leval{j}(f_t)$ for all $\seval{j} \in \sevalset$.
    \STATE Train model $f_t$ with $n/T$ samples from mixture $p_t$ over $\strainset$. Update model $f_{t+1} = \Phi(f_t, p_t)$.
    \STATE Set $p_{t+1}^i = \exp(\eta \sum_{\tau = t - w+1}^t \sum_{j = 1}^m A_{ij} \Leval{j}(f_{\tau}))$.
   \ENDFOR
\end{algorithmic}
\label{alg:skillit}
\end{algorithm}

Despite accounting for prerequisite skills, one shortcoming of skill-stratified sampling is that even if a skill has already obtained sufficiently low validation loss early during training, we will continue to allocate the same weight to that skill throughout training. Therefore, we formulate our data selection problem as an online learning problem and propose \name, which both prioritizes prerequisite skills and skills that are not yet learned. 

We are given a budget of $T$ rounds and $n$ total samples to train on.
At round $t$, we select a mixture $p_t \in \Delta^{k-1}$ from the $k$-dimensional unit simplex, and for each training skill $\strain{i} \in \strainset$, we sample from $\X_{\strain{i}}$ with proportion $p_t^i$ for a total of $\frac{n}{T}$ samples per round. Let $f_t$ be the model at at the start of round $t$. We can define $f_t$ recursively as a function of the previous round's model $f_{t-1}$ and mixture $p_{t-1}$ via a dynamics function $\Phi: \F \times \Delta^{k-1} \rightarrow \F$; that is, $f_t = \Phi(f_{t-1}, p_{t-1})$.
Let $\Leval{j}(f_{t})$ be the validation loss of $f_t$ on $\seval{j}$. Our goal is to select $p_1, \dots, p_T$ to minimize loss per evaluation skill at the end of training:
\begin{align}
    \underset{p_1, \dots, p_T \in \Delta^{k-1}}{\text{minimize}} \; \frac{1}{m} \sum_{j=1}^m \Leval{j}(f_T).\label{eq:opt}
\end{align}

This optimization problem is challenging to solve without additional assumptions. In order to make the problem tractable, we impose an explicit dynamics rule for the each evaluation skill's loss $\Leval{j}$ in terms of the current loss and data mixture. 
Assuming for simplicity that $\sevalset \subseteq \strainset$, a simple rule would be $\Leval{j}(f_t) = \Leval{j}(\Phi(f_{t-1}, p_{t-1})) := \Leval{j}(f_{t-1}) (1 - \alpha p_{t-1}^j)$ for $\alpha \in [0, 1]$. That is, we expect that allocating more data to skill $j$ should result in the validation loss on skill $j$ decreasing. However, such an expression assumes that only training on the $j$th skill will help learn the $j$th skill.  Instead, Section~\ref{sec:skill_examples} suggests that there are other skills that may help with the $j$th skill. We propose the following dynamics:
\begin{align}
    \Leval{j}(f_t)=\Leval{j}(f_{t-1})(1- A_{:, j}^\top p_{t-1}),
\end{align}

where $A_{:, j}$ is the column with weights of all skills that influence $\seval{j}$, and we absorb the scalar $\alpha$ into $A$.
The optimization problem in~\eqref{eq:opt} can thus be simplified as follows:
\begin{align} \label{eq:skillit_opt}
    &\underset{p_1, \dots, p_T \in \Delta^{k-1}}{\text{minimize}} \; \frac{1}{m}\sum_{j = 1}^m \Leval{j}(f_T) \\
    &  \text{s.t} \;\;\;  f_t = \Phi(f_{t-1}, p_{t-1}) \; \forall t = 1, \dots T \nonumber \\
    &\;\;\;\;\;\;\Leval{j}(f_t) = \Leval{j}(f_{t-1})(1 -  A_{:, j}^\top p_{t-1}) \; \forall j \in [m] \nonumber
 \end{align}

In Appendix~\ref{supp:alg}, we derive the following update rule via online mirror descent~\cite{nemirovskij1983problem} for learning rate $\eta > 0$:
\begin{align}
    p_{t+1}^{i} = p_{t}^{i} \exp \bigg(\eta \sum_{j = 1}^m A_{ij} \Leval{j}(f_{t})\bigg).\label{eq:mw_update}
\end{align}

 In addition, when equation~\ref{eq:mw_update} is expanded, we have that $p_{t+1}^i = p_1^{i} \exp \Big(\eta \sum_{\tau = 1}^{t} \sum_{j = 1}^m A_{ij} \Leval{j}(f_{\tau})\Big)$. Since this summation over $\tau$ results in diminishing strength of updates, we change it to a moving window of size $w$. Our full method is in Algorithm~\ref{alg:skillit}.

 Intuitively, at each step we adjust the weight on skill $i$ based on the losses of skills that $i$ influences, with the assumption that more training data helps decrease loss. Note that when we use our algorithm with a complete graph or empty graph, we achieve expected behavior discussed in Section~\ref{sec:def}. For the complete graph, our algorithm reduces to stratified sampling. When we have a skill set with an empty graph, the update rule reduces to sampling proportional to each skill's validation loss.

\section{Experimental results}\label{sec:exp}

Given an ordered skill set, we aim to validate \name's ability to select data for efficiently learning skills in the continual pre-training, fine-tuning, and out-of-domain settings. 
We provide full tables of results in Appendix~\ref{supp:full} and results where we learn the skills graph on the 125M model and use it for the 1.3B parameter model in Appendix~\ref{supp:larger_model}. Skills graphs are in Appendix~\ref{supp:skill_graphs}, weight trajectories for \name are in Appendix~\ref{supp:weights}, and ablations on the graph and online components of \name are in Appendix~\ref{supp:ablations}.

\subsection{Continual pre-training}

\noindent \textbf{Setup} \; We evaluate the ability of \name to select data for efficiently learning over all skills. We measure average validation loss per skill after a fixed number of training steps. We construct the LEGO synthetic and addition synthetic with $k = 5$ and $3$, respectively, and an imbalanced dataset over the skills. On the Natural Instructions dataset, we use $23$ of the task categories as skills.

\noindent \textbf{Baselines}\; We compare \name against three baselines that do not account for skills: random sampling, curriculum learning, and anticurriculum learning. Random sampling is a standard procedure for selecting samples given no additional information. Curriculum learning~\cite{bengio2009curriculum} and anticurriculum learning~\cite{wu2020curricula} score the samples from easiest to hardest and vice versa, respectively, and sample over an expanding set of the lowest scored samples at every epoch; we use the pre-trained model's loss to rank points. We evaluate skill-stratified sampling, which uses knowledge of the skills but is not online, and include an additional skills curriculum baseline in Appendix~\ref{supp:full}

\begin{figure*}[t]
    \centering
    \includegraphics[width=0.56\textwidth]{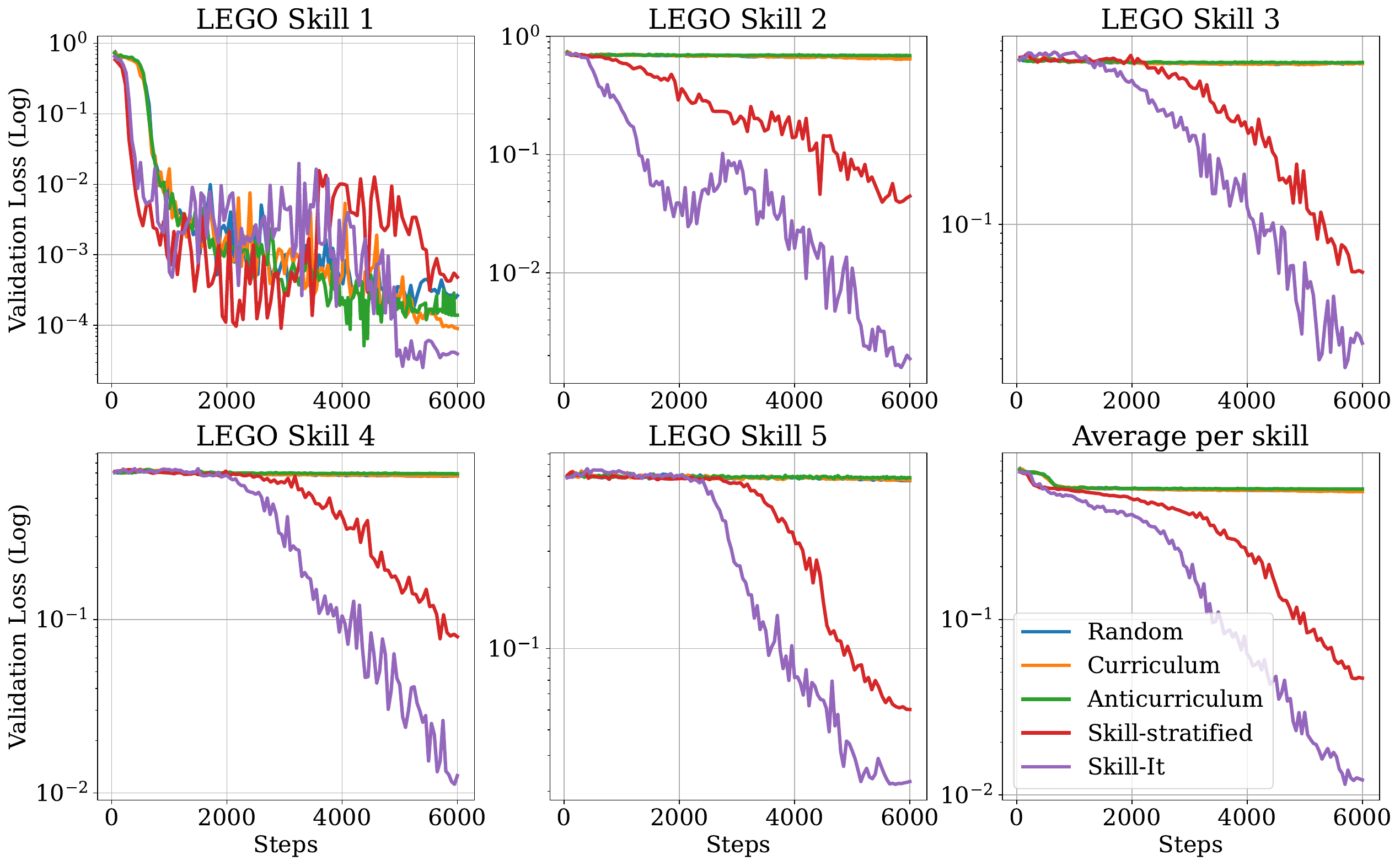}
    \includegraphics[width=0.38\textwidth]{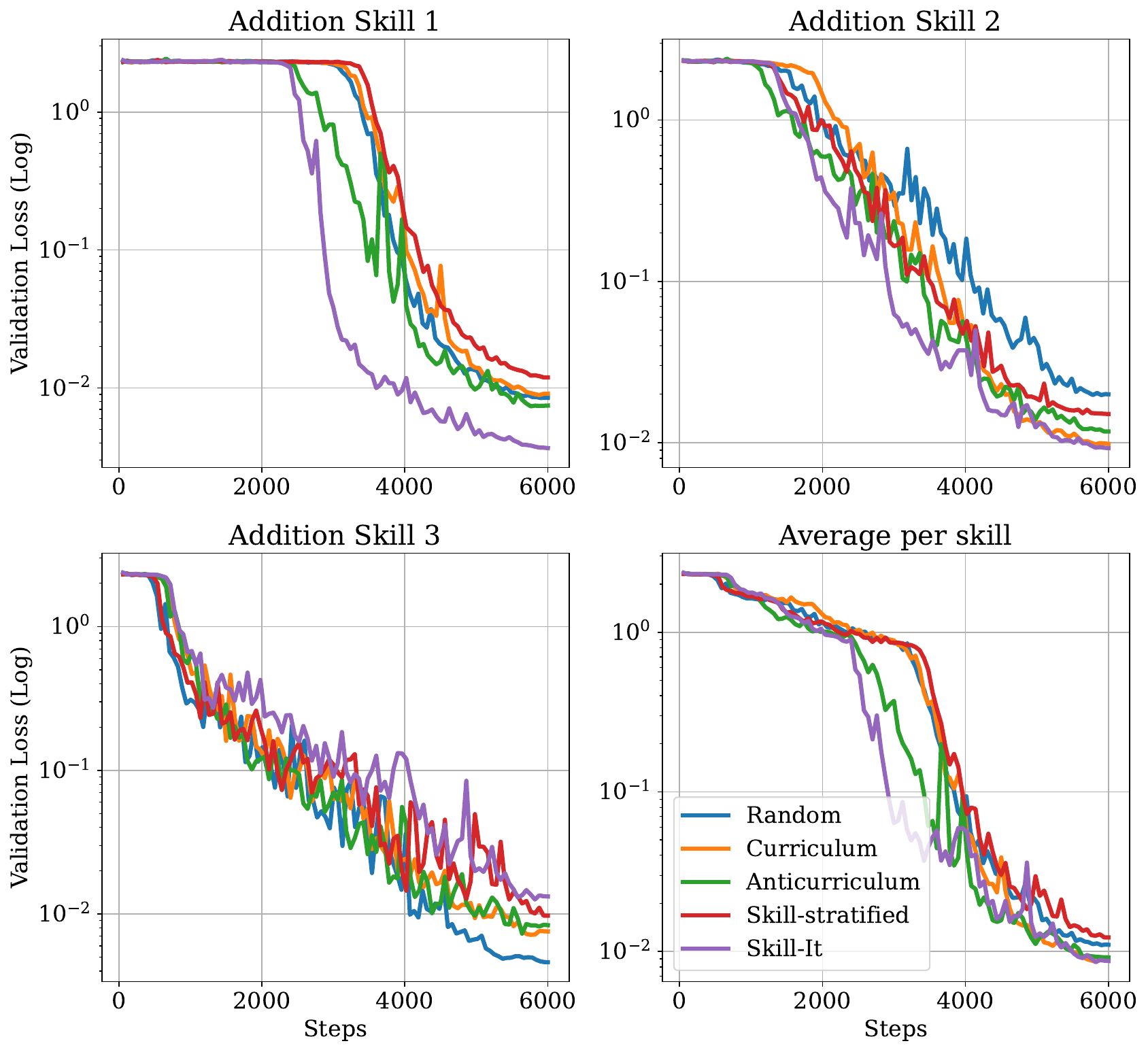}
    \caption{Performance of \name on each skill in the continual pre-training setting (learning over all skills in the ordered training skill set) on the LEGO synthetic (left) and addition synthetic (right).}
    \label{fig:lego_mw}
\end{figure*}

\noindent \textbf{Analysis} \; Our results are shown in Figure~\ref{fig:lego_mw}. Across our experiments we find that \name outperforms baselines that do not use skills as well as skill-stratified sampling. 
On the LEGO dataset, all three baselines that do not utilize a notion of skills exhibit plateauing loss on four of the skills. Both skill-stratified sampling and \name are able to significantly reduce loss on all skills, but the former is slower. Halfway through training, \name exhibits an accuracy improvement between $9.9$ and $25.9$ points over other approaches, reaching a final accuracy of $99.4$ (Figure~\ref{fig:lego_accs}).
\name outperforms skill-stratified sampling by initially allocating more weight to prerequisite skills and eventually allocating more weight to skills that are learned more slowly (Figure~\ref{fig:lego_weights}).
On the addition synthetic with $k = 3$, \name converges to near-zero validation loss faster than the baselines on skills 1 and 2. 
While the random baseline may seem competitive at first glance, it fails to learn skill 1 (adding together the ones digits), which hurts its average loss per skill. 
On NI, the validation loss from \name is $3.2\%$ lower than from random sampling (Table~\ref{tab:ni_pretraining}). Our results suggest that exploiting the construction and ordering of skills is critical to learning skills quickly.

\subsection{Fine-tuning}

\begin{figure}[h]
    \centering
    \includegraphics[width=0.24\columnwidth]{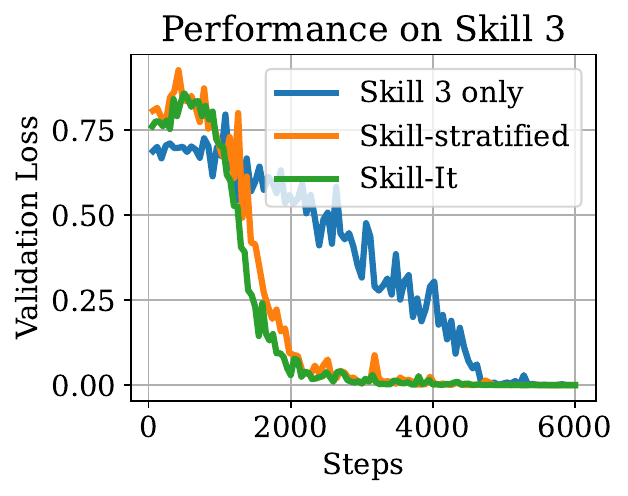}
    \includegraphics[width=0.22\columnwidth]{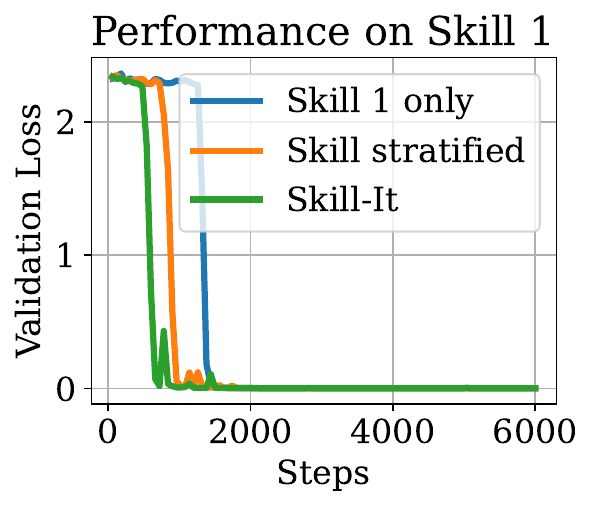}
    \includegraphics[width=0.23\columnwidth]{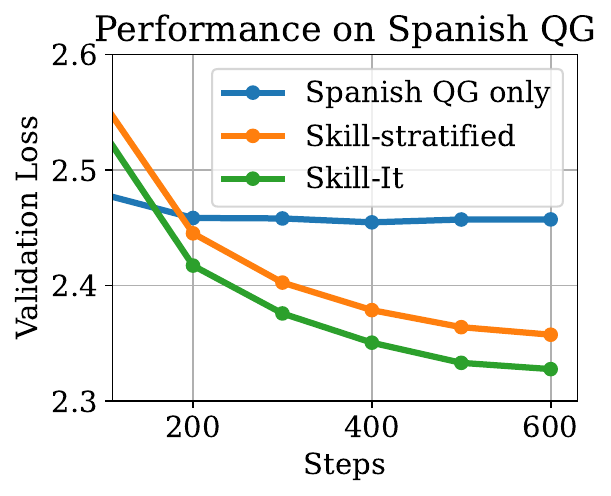}
    \includegraphics[width=0.24\columnwidth]{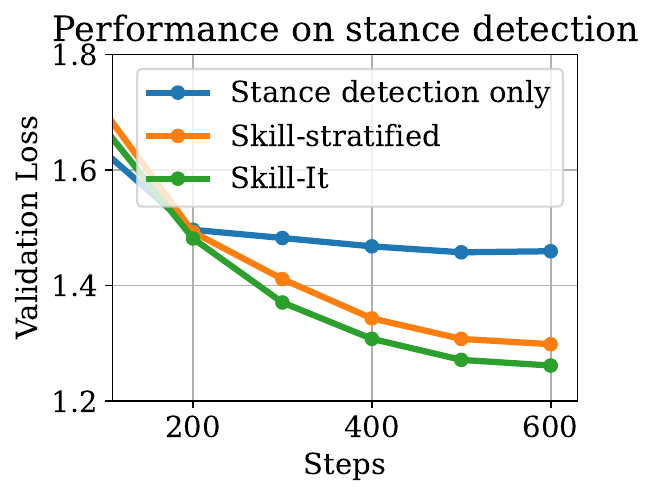}
    \caption{Performance of \name in the fine-tuning setting (learning a target skill using the ordered training skill set) on LEGO, addition, and NI.
    }
    \label{fig:ni_targeted}
\end{figure}

\noindent \textbf{Setup} \; We evaluate the ability of \name to select data from an ordered training skill set for learning a target skill. Mirroring Figure~\ref{fig:ni_lego_examples}, we evaluate on LEGO target skill 3 (third in reasoning chain), on the addition synthetic's skill 1 (ones place digit addition), and on NI's Spanish QG and Stance Detection.

\noindent \textbf{Baselines} \; We compare \name against training on the target skill only and skill-stratified sampling over prerequisite skills and the target skill. The skill-stratified sampling approach uses the ordered skill set to identify prerequisite skills, but does not exploit them dynamically. 

\noindent \textbf{Analysis} \; Our results are shown in Figure~\ref{fig:ni_targeted}. 
On LEGO, \name results in the same validation loss of $0.01$ as training only on the target skill in $38.1\%$ fewer steps.
We observe a similar trend on addition, with \name converging to a validation loss of $0.01$ in $59\%$ fewer steps required to do so when training only on the target skill.
Finally, on NI, \name improves validation loss on Spanish question generation by $5.3\%$ and Stance Detection by $13.6\%$ over just training on the respective target skill only.  
In this setting, a significant portion of the improvement over training only on the target skill comes from identification of prerequisite skills through the learned graph in the skill-stratified sampling method. \name is further able to improve performance with finer-grained dynamic weighting on prerequisite skills.

\subsection{Out-of-domain setting}

\begin{figure}
    \centering
    \includegraphics[width=4.8in]{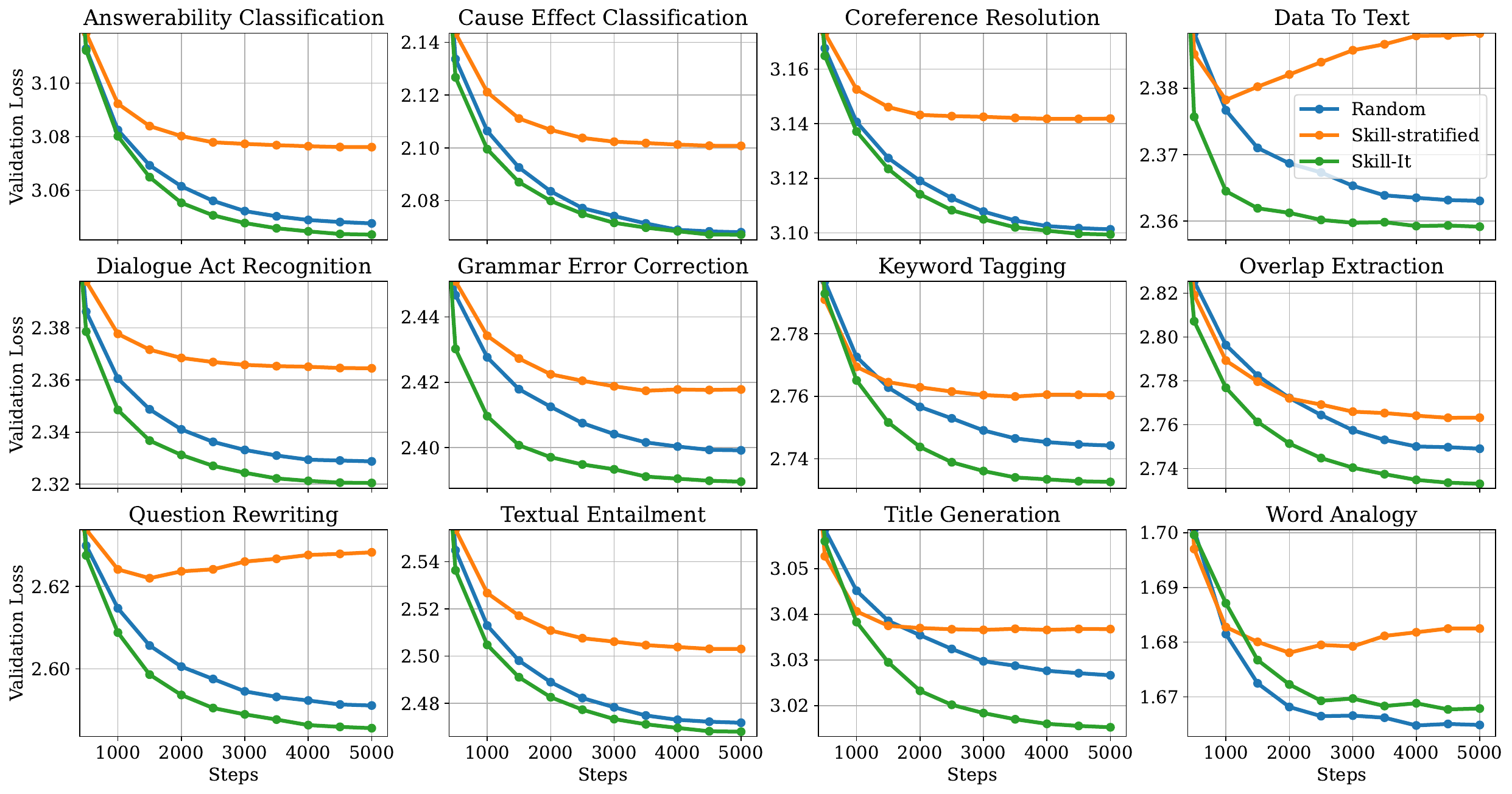}
    \caption{Performance of \name in the out-of-domain setting for the NI test task split. \name uses the graph between the train and evaluation skills to produce an online mixture on the training dataset.}
    \label{fig:ni_test}
\end{figure}

\noindent \textbf{Natural Instructions} \;
We evaluate the ability of \name to select data from a set of training skills for learning a disjoint set of evaluation skills that we cannot train on. We use all $59$ task categories in the NI train tasks split as the training skills and the $12$ task categories in the test tasks split as our evaluation skills.
We compare \name against random and skill-stratified sampling, both of which do not exploit the relationships between training skills and evaluation skills.
\name achieves the lowest loss on 11 out of 12 task categories over random and skill-stratified sampling (Figure~\ref{fig:ni_test}, tables in Appendix). 

\begin{figure}
\begin{center}
\begin{minipage}{.3\linewidth}
\begin{center}
    \includegraphics[height=1in]{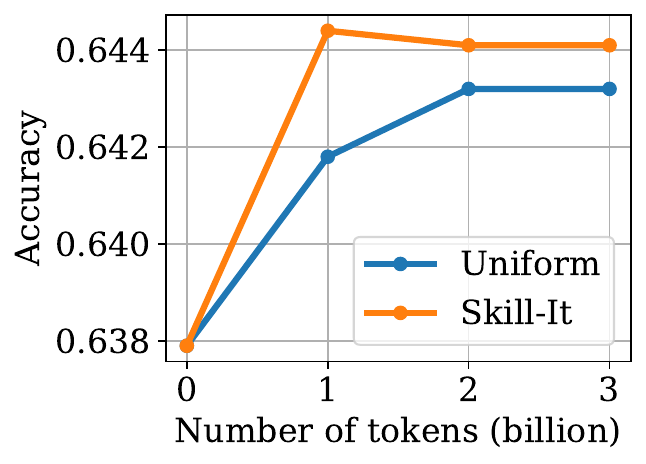}
\end{center}
\end{minipage} \;
\begin{minipage}{.3\linewidth} 
\begin{center}
\scriptsize
    \begin{tabular}{c | c}
        RedPajama source & \name mixture \\
        \toprule 
        ArXiv & 0.1370 \\
        Books & 0.0437 \\
        C4 & 0.4195 \\
        CommonCrawl & 0.0732 \\
        GitHub & 0.189 \\
        StackExchange & 0.0892 \\
        Wikipedia & 0.0484
    \end{tabular}
\end{center}
\end{minipage}
\end{center}
\caption{Left: Accuracy on LM Evaluation Harness for continual pre-training of a 3B parameter model using \name on the RedPajama dataset. We achieve higher accuracy at 1B additional tokens than uniform at 3B tokens. Right: \name mixture over RedPajama sources.}
\label{fig:redpajama}
\end{figure}

\noindent \textbf{RedPajama} \;
We use \name to produce a data mixture on the RedPajama dataset. The training skills are the data sources comprising the dataset, and the evaluation skills are several tasks from the Language Model Evaluation Harness~\cite{eval-harness}. \name with $T = 1$ (i.e. a static, graph-based mixture) yields the mixture in Figure~\ref{fig:redpajama} (right). We continually pre-train a 3B parameter model trained on one trillion tokens for three billion additional tokens using this mixture, and see that it outperforms uniform sampling over the data sources (Figure~\ref{fig:redpajama} left). In particular, \name achieves higher accuracy with 1B additional tokens than uniform with 3B additional tokens.

\section{Related work}
\vspace{-0.5em}

\paragraph{Data selection for LMs}
There have been several studies of large-scale data selection for LMs. 
Data deduplication~\cite{lee2022deduplicating, abbas2023semdedup, hern2022scaling}, in which identical or nearly identical samples are removed, is a method that enables LMs to be trained on smaller, cleaned datasets and has been increasingly used as a pre-processing step for training data~\cite{touvron2023llama, biderman2023pythia, zhang2022opt}. Other methods applied at scale involve ensuring high quality of data by explicitly filtering out samples or comparing the training dataset with a cleaned reference dataset~\cite{brown2020language, touvron2023llama, laurenon2023bigscience}. Importance reweighting approaches have also been proposed for identifying training data from a large corpus that best approximates a smaller target distribution~\cite{xie2023data}, and influence functions have been used to select a subset of training data to improve performance on downstream tasks~\cite{wang2023farewell}. These approaches can identify data pertaining to a particular target distribution or filter out low quality data according to some heuristic, while our work aims to understand how the choice of data is related to the numerous skills that LMs learn. 

Recent development of LMs has shifted focus from emphasizing the scale of the model to prioritizing the training data utilized. For example, models like Alpaca~\cite{alpaca}, Vicuna~\cite{vicuna2023}, and Koala~\cite{koala_blogpost_2023} are all based on the LLaMA model combined with instruction data generated by an existing LM. Palm 2's technical report states that the data mixture was a critical component of the final model~\cite{palm2techreport}, and Mosaic ML's recent MPT model was trained on a hand-engineered mixture of the RedPajama dataset~\cite{mpt}. However, these works lack rigorous explanation for why their training datasets were constructed in this way. 

Finally, perhaps most related to our approach is the contemporary work DoReMi~\cite{xie2023doremi}, which uses group distributionally robust optimization on a smaller LM to select data source mixtures for training a larger LM. Their approach focuses on selecting data at the data source level for optimizing worst-case performance across the training data sources, rather than at the more general skills level for a variety of target skill sets. Furthermore, we focus on understanding how skills are related to each other and induce some order in how LMs learn by explicitly modeling skill graph structure, which we find to be important for data-efficient LM training (see ablations in Appendix~\ref{supp:ablations}).

\paragraph{Data selection methods}

Many data selection methods have been proposed for supervised, task-specific settings. In this setting, the most typical objective is dataset condensation, which aims to identify a small subset of data that captures the larger dataset's properties with respect to the model. Some approaches include constructing coresets~\cite{langberg2010universal, phillips2016coresets}, identifying samples that the model forgets during training~\cite{toneva2018empirical}; identifying samples with the largest gradients~\cite{paul2021deep} or gradients that approximate the overall gradient~\cite{mirzasoleiman2019coresets}; clustering in embedding space and selecting points farthest from cluster centers~\cite{sorscher2022neural}; and selecting samples with the highest uncertainty or entropy~\cite{lewis1995sequential}. These approaches have also been shown to transfer from smaller models to larger models~\cite{coleman2019selection}. Unlike these methods, we study how to select data for learning one or many skills at the mixture level for LMs instead of the instance level.

Another area of interest is data selection for domain adaptation and multitask learning. For domain adaptation, there are a wide range of methods that select data to best match the target distribution. For example, the Moore-Lewis method matches data based on the difference in cross-entropy using a model trained on the target versus a model trained on the source data~\cite{moore2010intelligent}. Several other approaches suggest training a model to distinguish between source and target and selecting points with high uncertainty~\cite{ruder2017data}, or selecting points based on some divergence in an embedding space~\cite{ruder2017learning}. In comparison to these approaches, our work focuses on learning one or many skills and also finds that embedding-based heuristics do not fully identify skills. 

\paragraph{Data attribution} Another perspective on understanding training data is data attribution, which seeks to identify what data is responsible for particular model behaviors. Influence functions~\cite{koh2017understanding} and shapley values~\cite{ghorbani2019data} are two ways to quantify the role of individual samples. Datamodels~\cite{ilyas2022datamodels} fit a model to predict behavior given a subset of training data, providing a framework for understanding individual samples as well as dataset counterfactuals. Simfluence~\cite{guu2023simfluence} fits a Markov process to a set of training trajectories for finer-grained understanding of how data impacts training. We focus on understanding how groups of data associated with skills elicit broader model capabilities, and utilize this understanding to select data for more efficient training.

\paragraph{Curriculum learning} Curriculum learning~\cite{bengio2009curriculum} proposes to show the model data in order from easy samples to hard ones. Various criteria have been used to determine hardness, and anticurriculum as well as various pacing functions and mixing rates have been explored~\cite{soviany2022curriculum}. Curriculum learning can also be performed at the group level~\cite{varshney2022let}. More sophisticated approaches include parametrizing each sample with a dynamic importance~\cite{saxena2019data}, and also accounting for irrelevant and noisy data~\cite{mindermann2021prioritized}. Our approach similarly utilizes a curriculum, but it is defined over a skills graph and does not necessarily align with training on easiest to hardest skills.

\paragraph{How LMs learn} Many different explanations for how LMs learn from data have been proposed. One hypothesis is that there exist discrete, universal building blocks of LM knowledge called quanta, and power law scaling emerges from a learning over a particular distribution of quanta in the right order~\cite{michaud2023quantization}. Another is that chain of thought reasoning emerges due to local clusters of latent variables that influence each other, which can be validated by studying the LM's ability to do conditional inference given intermediate variables~\cite{prystawski2023think}. Others have provided theoretical analysis of how transformers learn topics by studying co-occurrences of words in the training data~\cite{li2023transformers}.
Empirically, how models learn is still a mystery---for instance, models trained on code are found to perform fairly well at commensense reasoning~\cite{madaan2022language}. Our work initiates a study on how LMs learn various skills and how to exploit this for better data selection. 

\paragraph{Task selection} In multitask auxiliary learning, the goal is to train a model to perform well on a target task(s) by selecting the most beneficial source tasks to train on. One can use feature similarity to select tasks~\cite{kung-etal-2021-efficient}, but we find in our synthetics that feature similarity does not always recover skills. In Taskonomy~\cite{zamir2018taskonomy}, a hypergraph over a set of tasks is learned and used to select tasks. The methods used to develop the taxonomy can be applied to further expand our graph learning (e.g., studying transitive and higher-order properties). However, their focus is on task selection in computer vision rather than data selection for LMs to learn skills. Lastly, the contemporary work of TaskWeb~\cite{kim2023taskweb} builds a graph among $22$ common NLP tasks in order to determine what the best source tasks are for a target task. Their definition of an edge in the task graph is less strict than ours (their comparison is on if training on additional data from $s_i$ helps with $s_j$, while we fix the overall amount of training data over both $s_i$ and $s_j$). Overall, our approach is similar in use of the skills graph, but we incorporate it into a dynamic sampling algorithm. Furthermore, we look more broadly at skills, rather than tasks, and characterize when we expect using the skills graph to improve model performance.

\paragraph{Education} The notion of skill has been studied in education. Classical
research on learning hierarchies~\cite{white1974past} identify sets of skills that make up subordinate capabilities for students. For instance,~\cite{gagne1961abilities} identified that in order for students to solve linear equations, there were many prerequisite skills, ranging from the simplest being symbol recognition to the most complex being the ability to add, subtract, multiple, and divide from both sides of the equation. More recently, decision-making over lesson sequences based on skills, e.g., what the student already knows versus what the lesson teaches, has become an area of interest in personalized learning~\cite{reddy2016latent}.

\section{Conclusion}

Given a fixed budget of data, knowing what data to train on to induce various capabilities in an LM is challenging. As LMs continue to improve, it will become increasingly important to extract as much signal as possible from the data and to direct that signal towards acquiring a broad variety of capabilities. In this paper, we introduce a skills-based framework for understanding how LMs learn and for selecting training data. We hope our study invites others to build on such a notion of skill and further explore how to align skills with data.

\section*{Acknowledgements}

We thank Together Computer (\url{https://together.xyz/}) for providing portions of the compute used to train models in this paper. We thank Sabri Eyuboglu, Karan Goel, Arjun Desai, Neel Guha, Michael Zhang, Vishnu Sarrukai, Simran Arora, Ben Spector, Brandon Yang, Gautam Machiraju, and Sang Michael Xie for their helpful feedback and discussion.

We gratefully acknowledge the support of NIH under No. U54EB020405 (Mobilize), NSF under Nos. CCF1763315 (Beyond Sparsity), CCF1563078 (Volume to Velocity), and 1937301 (RTML); US DEVCOM ARL under No. W911NF-21-2-0251 (Interactive Human-AI Teaming); ONR under No. N000141712266 (Unifying Weak Supervision); ONR N00014-20-1-2480: Understanding and Applying Non-Euclidean Geometry in Machine Learning; N000142012275 (NEPTUNE); NXP, Xilinx, LETI-CEA, Intel, IBM, Microsoft, NEC, Toshiba, TSMC, ARM, Hitachi, BASF, Accenture, Ericsson, Qualcomm, Analog Devices, Google Cloud, Salesforce, Total, the HAI-GCP Cloud Credits for Research program,  the Stanford Data Science Initiative (SDSI), and members of the Stanford DAWN project: Facebook, Google, and VMWare. FS is supported by NSF CCF2106707 and the Wisconsin Alumni Research Foundation (WARF). The U.S. Government is authorized to reproduce and distribute reprints for Governmental purposes notwithstanding any copyright notation thereon. Any opinions, findings, and conclusions or recommendations expressed in this material are those of the authors and do not necessarily reflect the views, policies, or endorsements, either expressed or implied, of NIH, ONR, or the U.S. Government. 

\newpage

\bibliography{references}

\begin{thebibliography}{10}

\bibitem{abbas2023semdedup}
Amro Abbas, Kushal Tirumala, Dániel Simig, Surya Ganguli, and Ari~S. Morcos.
\newblock Semdedup: Data-efficient learning at web-scale through semantic
  deduplication, 2023.

\bibitem{bai2022training}
Yuntao Bai, Andy Jones, et~al.
\newblock Training a helpful and harmless assistant with reinforcement learning
  from human feedback, 2022.

\bibitem{bengio2009curriculum}
Yoshua Bengio, J{\'e}r{\^o}me Louradour, Ronan Collobert, and Jason Weston.
\newblock Curriculum learning.
\newblock In {\em Proceedings of the 26th annual international conference on
  machine learning}, pages 41--48, 2009.

\bibitem{biderman2023pythia}
Stella Biderman, Hailey Schoelkopf, Quentin Anthony, Herbie Bradley, Kyle
  O'Brien, Eric Hallahan, Mohammad~Aflah Khan, Shivanshu Purohit, USVSN~Sai
  Prashanth, Edward Raff, Aviya Skowron, Lintang Sutawika, and Oskar van~der
  Wal.
\newblock Pythia: A suite for analyzing large language models across training
  and scaling, 2023.

\bibitem{gpt-neo}
Sid Black, Gao Leo, Phil Wang, Connor Leahy, and Stella Biderman.
\newblock {GPT-Neo: Large Scale Autoregressive Language Modeling with
  Mesh-Tensorflow}, March 2021.
\newblock {If you use this software, please cite it using these metadata.}

\bibitem{bommasani2021opportunities}
Rishi Bommasani, Percy Liang, et~al.
\newblock On the opportunities and risks of foundation models, 2021.

\bibitem{brown2020language}
Tom~B. Brown, Benjamin Mann, Nick Ryder, Melanie Subbiah, Jared Kaplan,
  Prafulla Dhariwal, Arvind Neelakantan, Pranav Shyam, Girish Sastry, Amanda
  Askell, et~al.
\newblock Language models are few-shot learners.
\newblock {\em arXiv preprint arXiv:2005.14165}, 2020.

\bibitem{chen2021evaluating}
Mark Chen, Jerry Tworek, et~al.
\newblock Evaluating large language models trained on code, 2021.

\bibitem{vicuna2023}
Wei-Lin Chiang, Zhuohan Li, Zi~Lin, Ying Sheng, Zhanghao Wu, Hao Zhang, Lianmin
  Zheng, Siyuan Zhuang, Yonghao Zhuang, Joseph~E. Gonzalez, Ion Stoica, and
  Eric~P. Xing.
\newblock Vicuna: An open-source chatbot impressing gpt-4 with 90\%* chatgpt
  quality, March 2023.

\bibitem{coleman2019selection}
Cody Coleman, Christopher Yeh, Stephen Mussmann, Baharan Mirzasoleiman, Peter
  Bailis, Percy Liang, Jure Leskovec, and Matei Zaharia.
\newblock Selection via proxy: Efficient data selection for deep learning,
  2019.

\bibitem{gagne1962acquisition}
Robert~M Gagne.
\newblock The acquisition of knowledge.
\newblock {\em Psychological review}, 69(4):355, 1962.

\bibitem{gagne1961abilities}
Robert~M Gagne and Noel~E Paradise.
\newblock Abilities and learning sets in knowledge acquisition.
\newblock {\em Psychological Monographs: General and Applied}, 75(14):1, 1961.

\bibitem{gao2020pile}
Leo Gao, Stella Biderman, Sid Black, Laurence Golding, Travis Hoppe, Charles
  Foster, Jason Phang, Horace He, Anish Thite, Noa Nabeshima, et~al.
\newblock The pile: An 800gb dataset of diverse text for language modeling.
\newblock {\em arXiv preprint arXiv:2101.00027}, 2020.

\bibitem{eval-harness}
Leo Gao, Jonathan Tow, Stella Biderman, Sid Black, Anthony DiPofi, Charles
  Foster, Laurence Golding, Jeffrey Hsu, Kyle McDonell, Niklas Muennighoff,
  Jason Phang, Laria Reynolds, Eric Tang, Anish Thite, Ben Wang, Kevin Wang,
  and Andy Zou.
\newblock A framework for few-shot language model evaluation, September 2021.

\bibitem{koala_blogpost_2023}
Xinyang Geng, Arnav Gudibande, Hao Liu, Eric Wallace, Pieter Abbeel, Sergey
  Levine, and Dawn Song.
\newblock Koala: A dialogue model for academic research.
\newblock Blog post, April 2023.

\bibitem{ghorbani2019data}
Amirata Ghorbani and James Zou.
\newblock Data shapley: Equitable valuation of data for machine learning.
\newblock In {\em International Conference on Machine Learning}, pages
  2242--2251. PMLR, 2019.

\bibitem{palm2techreport}
{Google}.
\newblock Palm2 technical report.
\newblock Technical report, 2023.

\bibitem{gupta2020}
Anupam Gupta.
\newblock Advanced algorithms: Notes for cmu 15-850 (fall 2020), 2020.

\bibitem{gururangan2020dont}
Suchin Gururangan, Ana Marasović, Swabha Swayamdipta, Kyle Lo, Iz~Beltagy,
  Doug Downey, and Noah~A. Smith.
\newblock Don’t stop pretraining: Adapt language models to domains and tasks.
\newblock {\em Proceedings of the 58th Annual Meeting of the Association for
  Computational Linguistics}, 2020.

\bibitem{guu2023simfluence}
Kelvin Guu, Albert Webson, Ellie Pavlick, Lucas Dixon, Ian Tenney, and Tolga
  Bolukbasi.
\newblock Simfluence: Modeling the influence of individual training examples by
  simulating training runs, 2023.

\bibitem{hendersonkrass2022pileoflaw}
Peter Henderson*, Mark~S. Krass*, Lucia Zheng, Neel Guha, Christopher~D.
  Manning, Dan Jurafsky, and Daniel~E. Ho.
\newblock Pile of law: Learning responsible data filtering from the law and a
  256gb open-source legal dataset, 2022.

\bibitem{hern2022scaling}
Danny Hernandez, Tom Brown, Tom Conerly, Nova DasSarma, Dawn Drain, Sheer
  El-Showk, Nelson Elhage, Zac Hatfield-Dodds, Tom Henighan, Tristan Hume,
  Scott Johnston, Ben Mann, Chris Olah, Catherine Olsson, Dario Amodei,
  Nicholas Joseph, Jared Kaplan, and Sam McCandlish.
\newblock Scaling laws and interpretability of learning from repeated data,
  2022.

\bibitem{ilyas2022datamodels}
Andrew Ilyas, Sung~Min Park, Logan Engstrom, Guillaume Leclerc, and Aleksander
  Madry.
\newblock Datamodels: Predicting predictions from training data, 2022.

\bibitem{kim2023taskweb}
Joongwon Kim, Akari Asai, Gabriel Ilharco, and Hannaneh Hajishirzi.
\newblock Taskweb: Selecting better source tasks for multi-task nlp, 2023.

\bibitem{kirk2021bias}
Hannah~Rose Kirk, Yennie Jun, Filippo Volpin, Haider Iqbal, Elias Benussi,
  Frederic Dreyer, Aleksandar Shtedritski, and Yuki Asano.
\newblock Bias out-of-the-box: An empirical analysis of intersectional
  occupational biases in popular generative language models.
\newblock In M.~Ranzato, A.~Beygelzimer, Y.~Dauphin, P.S. Liang, and J.~Wortman
  Vaughan, editors, {\em Advances in Neural Information Processing Systems},
  volume~34, pages 2611--2624. Curran Associates, Inc., 2021.

\bibitem{kitaev-etal-2019-multilingual}
Nikita Kitaev, Steven Cao, and Dan Klein.
\newblock Multilingual constituency parsing with self-attention and
  pre-training.
\newblock In {\em Proceedings of the 57th Annual Meeting of the Association for
  Computational Linguistics}, pages 3499--3505, Florence, Italy, July 2019.
  Association for Computational Linguistics.

\bibitem{kitaev-klein-2018-constituency}
Nikita Kitaev and Dan Klein.
\newblock Constituency parsing with a self-attentive encoder.
\newblock In {\em Proceedings of the 56th Annual Meeting of the Association for
  Computational Linguistics (Volume 1: Long Papers)}, pages 2676--2686,
  Melbourne, Australia, July 2018. Association for Computational Linguistics.

\bibitem{koh2017understanding}
Pang~Wei Koh and Percy Liang.
\newblock Understanding black-box predictions via influence functions, 2017.

\bibitem{kung-etal-2021-efficient}
Po-Nien Kung, Sheng-Siang Yin, Yi-Cheng Chen, Tse-Hsuan Yang, and Yun-Nung
  Chen.
\newblock Efficient multi-task auxiliary learning: Selecting auxiliary data by
  feature similarity.
\newblock In {\em Proceedings of the 2021 Conference on Empirical Methods in
  Natural Language Processing}, pages 416--428, Online and Punta Cana,
  Dominican Republic, November 2021. Association for Computational Linguistics.

\bibitem{langberg2010universal}
Michael Langberg and Leonard~J. Schulman.
\newblock {\em Universal approximators for integrals}, pages 598--607.

\bibitem{laurenon2023bigscience}
Hugo Laurençon, Lucile Saulnier, et~al.
\newblock The bigscience roots corpus: A 1.6tb composite multilingual dataset,
  2023.

\bibitem{lee2022deduplicating}
Katherine Lee, Daphne Ippolito, Andrew Nystrom, Chiyuan Zhang, Douglas Eck,
  Chris Callison-Burch, and Nicholas Carlini.
\newblock Deduplicating training data makes language models better.
\newblock {\em Proceedings of the 60th Annual Meeting of the Association for
  Computational Linguistics (Volume 1: Long Papers)}, 2022.

\bibitem{lewis1995sequential}
David~D Lewis.
\newblock A sequential algorithm for training text classifiers: Corrigendum and
  additional data.
\newblock In {\em Acm Sigir Forum}, volume~29, pages 13--19. ACM New York, NY,
  USA, 1995.

\bibitem{li2023transformers}
Yuchen Li, Yuanzhi Li, and Andrej Risteski.
\newblock How do transformers learn topic structure: Towards a mechanistic
  understanding, 2023.

\bibitem{liang2021towards}
Paul~Pu Liang, Chiyu Wu, Louis-Philippe Morency, and Ruslan Salakhutdinov.
\newblock Towards understanding and mitigating social biases in language
  models.
\newblock In Marina Meila and Tong Zhang, editors, {\em Proceedings of the 38th
  International Conference on Machine Learning}, volume 139 of {\em Proceedings
  of Machine Learning Research}, pages 6565--6576. PMLR, 18--24 Jul 2021.

\bibitem{madaan2022language}
Aman Madaan, Shuyan Zhou, Uri Alon, Yiming Yang, and Graham Neubig.
\newblock Language models of code are few-shot commonsense learners, 2022.

\bibitem{michaud2023quantization}
Eric~J. Michaud, Ziming Liu, Uzay Girit, and Max Tegmark.
\newblock The quantization model of neural scaling, 2023.

\bibitem{mindermann2021prioritized}
Sören Mindermann, Muhammed Razzak, Winnie Xu, Andreas Kirsch, Mrinank Sharma,
  Adrien Morisot, Aidan~N. Gomez, Sebastian Farquhar, Jan Brauner, and Yarin
  Gal.
\newblock Prioritized training on points that are learnable, worth learning,
  and not yet learned (workshop version), 2021.

\bibitem{mirzasoleiman2019coresets}
Baharan Mirzasoleiman, Jeff Bilmes, and Jure Leskovec.
\newblock Coresets for data-efficient training of machine learning models,
  2019.

\bibitem{naturalinstructions}
Swaroop Mishra, Daniel Khashabi, Chitta Baral, and Hannaneh Hajishirzi.
\newblock Cross-task generalization via natural language crowdsourcing
  instructions.
\newblock In {\em ACL}, 2022.

\bibitem{moore2010intelligent}
Robert~C. Moore and William Lewis.
\newblock Intelligent selection of language model training data.
\newblock In {\em Proceedings of the {ACL} 2010 Conference Short Papers}, pages
  220--224, Uppsala, Sweden, July 2010. Association for Computational
  Linguistics.

\bibitem{mpt}
MosaicML.
\newblock Introducing mpt-7b: A new standard for open-source, commercially
  usable llms, 2023.

\bibitem{nadeem2021stereoset}
Moin Nadeem, Anna Bethke, and Siva Reddy.
\newblock Stereoset: Measuring stereotypical bias in pretrained language
  models.
\newblock {\em Proceedings of the 59th Annual Meeting of the Association for
  Computational Linguistics and the 11th International Joint Conference on
  Natural Language Processing (Volume 1: Long Papers)}, 2021.

\bibitem{nanda2023progress}
Neel Nanda, Lawrence Chan, Tom Lieberum, Jess Smith, and Jacob Steinhardt.
\newblock Progress measures for grokking via mechanistic interpretability.
\newblock In {\em The Eleventh International Conference on Learning
  Representations}, 2023.

\bibitem{nemirovskij1983problem}
Arkadij~Semenovi{\v{c}} Nemirovskij and David~Borisovich Yudin.
\newblock Problem complexity and method efficiency in optimization.
\newblock 1983.

\bibitem{paul2021deep}
Mansheej Paul, Surya Ganguli, and Gintare~Karolina Dziugaite.
\newblock Deep learning on a data diet: Finding important examples early in
  training, 2021.

\bibitem{phillips2016coresets}
Jeff~M. Phillips.
\newblock Coresets and sketches, 2016.

\bibitem{prystawski2023think}
Ben Prystawski and Noah~D. Goodman.
\newblock Why think step-by-step? reasoning emerges from the locality of
  experience, 2023.

\bibitem{reddy2016latent}
Siddharth Reddy, Igor Labutov, and Thorsten Joachims.
\newblock Latent skill embedding for personalized lesson sequence
  recommendation, 2016.

\bibitem{ruder2017data}
Sebastian Ruder, Parsa Ghaffari, and John~G. Breslin.
\newblock Data selection strategies for multi-domain sentiment analysis, 2017.

\bibitem{ruder2017learning}
Sebastian Ruder and Barbara Plank.
\newblock Learning to select data for transfer learning with bayesian
  optimization.
\newblock {\em Proceedings of the 2017 Conference on Empirical Methods in
  Natural Language Processing}, 2017.

\bibitem{saxena2019data}
Shreyas Saxena, Oncel Tuzel, and Dennis DeCoste.
\newblock Data parameters: A new family of parameters for learning a
  differentiable curriculum.
\newblock In H.~Wallach, H.~Larochelle, A.~Beygelzimer, F.~d\textquotesingle
  Alch\'{e}-Buc, E.~Fox, and R.~Garnett, editors, {\em Advances in Neural
  Information Processing Systems}, volume~32. Curran Associates, Inc., 2019.

\bibitem{sorscher2022neural}
Ben Sorscher, Robert Geirhos, Shashank Shekhar, Surya Ganguli, and Ari~S.
  Morcos.
\newblock Beyond neural scaling laws: beating power law scaling via data
  pruning, 2022.

\bibitem{soviany2022curriculum}
Petru Soviany, Radu~Tudor Ionescu, Paolo Rota, and Nicu Sebe.
\newblock Curriculum learning: A survey.
\newblock {\em International Journal of Computer Vision}, 130(6):1526–1565,
  Apr 2022.

\bibitem{stevenson2022putting}
Claire Stevenson, Iris Smal, Matthijs Baas, Raoul Grasman, and Han van~der
  Maas.
\newblock Putting gpt-3's creativity to the (alternative uses) test, 2022.

\bibitem{alpaca}
Rohan Taori, Ishaan Gulrajani, Tianyi Zhang, Yann Dubois, Xuechen Li, Carlos
  Guestrin, Percy Liang, and Tatsunori~B. Hashimoto.
\newblock Stanford alpaca: An instruction-following llama model.
\newblock \url{https://github.com/tatsu-lab/stanford_alpaca}, 2023.

\bibitem{redpajama}
Together.
\newblock Redpajama-data: An open source recipe to reproduce llama training
  dataset, 2023.

\bibitem{toneva2018empirical}
Mariya Toneva, Alessandro Sordoni, Remi~Tachet des Combes, Adam Trischler,
  Yoshua Bengio, and Geoffrey~J. Gordon.
\newblock An empirical study of example forgetting during deep neural network
  learning, 2018.

\bibitem{touvron2023llama}
Hugo Touvron, Thibaut Lavril, Gautier Izacard, Xavier Martinet, Marie-Anne
  Lachaux, Timothée Lacroix, Baptiste Rozière, Naman Goyal, Eric Hambro,
  Faisal Azhar, Aurelien Rodriguez, Armand Joulin, Edouard Grave, and Guillaume
  Lample.
\newblock Llama: Open and efficient foundation language models, 2023.

\bibitem{varshney2022let}
Neeraj Varshney, Swaroop Mishra, and Chitta Baral.
\newblock Let the model decide its curriculum for multitask learning, 2022.

\bibitem{wang2023farewell}
Xiao Wang, Weikang Zhou, Qi~Zhang, Jie Zhou, Songyang Gao, Junzhe Wang, Menghan
  Zhang, Xiang Gao, Yunwen Chen, and Tao Gui.
\newblock Farewell to aimless large-scale pretraining: Influential subset
  selection for language model, 2023.

\bibitem{wang2022selfinstruct}
Yizhong Wang, Yeganeh Kordi, Swaroop Mishra, Alisa Liu, Noah~A. Smith, Daniel
  Khashabi, and Hannaneh Hajishirzi.
\newblock Self-instruct: Aligning language model with self generated
  instructions, 2022.

\bibitem{wang2022supernaturalinstructions}
Yizhong Wang, Swaroop Mishra, et~al.
\newblock Super-naturalinstructions: Generalization via declarative
  instructions on 1600+ nlp tasks, 2022.

\bibitem{wei2021finetuned}
Jason Wei, Maarten Bosma, Vincent~Y. Zhao, Kelvin Guu, Adams~Wei Yu, Brian
  Lester, Nan Du, Andrew~M. Dai, and Quoc~V. Le.
\newblock Finetuned language models are zero-shot learners, 2021.

\bibitem{white1973research}
Richard~T White.
\newblock Research into learning hierarchies.
\newblock {\em Review of Educational Research}, 43(3):361--375, 1973.

\bibitem{white1974past}
Richard~T. White and Robert~M. Gagné.
\newblock Past and future research on learning hierarchies.
\newblock {\em Educational Psychologist}, 11(1):19--28, 1974.

\bibitem{wu2020curricula}
Xiaoxia Wu, Ethan Dyer, and Behnam Neyshabur.
\newblock When do curricula work?, 2020.

\bibitem{xie2023doremi}
Sang~Michael Xie, Hieu Pham, Xuanyi Dong, Nan Du, Hanxiao Liu, Yifeng Lu, Percy
  Liang, Quoc~V. Le, Tengyu Ma, and Adams~Wei Yu.
\newblock Doremi: Optimizing data mixtures speeds up language model
  pretraining, 2023.

\bibitem{xie2023data}
Sang~Michael Xie, Shibani Santurkar, Tengyu Ma, and Percy Liang.
\newblock Data selection for language models via importance resampling, 2023.

\bibitem{zamir2018taskonomy}
Amir~R. Zamir, Alexander Sax, William Shen, Leonidas Guibas, Jitendra Malik,
  and Silvio Savarese.
\newblock Taskonomy: Disentangling task transfer learning.
\newblock {\em 2018 IEEE/CVF Conference on Computer Vision and Pattern
  Recognition}, Jun 2018.

\bibitem{zhang2022opt}
Susan Zhang, Stephen Roller, Naman Goyal, Mikel Artetxe, Moya Chen, Shuohui
  Chen, Christopher Dewan, Mona Diab, Xian Li, Xi~Victoria Lin, Todor Mihaylov,
  Myle Ott, Sam Shleifer, Kurt Shuster, Daniel Simig, Punit~Singh Koura, Anjali
  Sridhar, Tianlu Wang, and Luke Zettlemoyer.
\newblock Opt: Open pre-trained transformer language models, 2022.

\bibitem{zhang2022unveiling}
Yi~Zhang, Arturs Backurs, Sébastien Bubeck, Ronen Eldan, Suriya Gunasekar, and
  Tal Wagner.
\newblock Unveiling transformers with lego: a synthetic reasoning task, 2022.

\end{thebibliography}
\bibliographystyle{plain}

\newpage
\appendix

\section{Broader Impacts and Limitations}

\paragraph{Broader Impacts} 
As more LMs are developed, a key criteria for their adoption and utility is if they exhibit a wide array of useful capabilities, such as generating harmless content, summarizing essays, and being conversational with the user. While improvements in other parts of the LM development pipeline such as training and architecture are important, many recent advances in building LMs with a wide array of useful capabilities have come from the data itself~\cite{palm2techreport, alpaca, vicuna2023, koala_blogpost_2023, mpt}. Our work is fundamental in investigating how LMs learn and how to select data to learn skills more efficiently. However, we recognize that data selection methods can always be utilized to optimize for particular skills that may be considered malicious or negatively target or exclude specific groups~\cite{bai2022training}. Furthermore, pre-trained LMs have been found to have various biases~\cite{kirk2021bias, nadeem2021stereoset, liang2021towards, bommasani2021opportunities}.

\paragraph{Limitations} The skills graph can either be provided (e.g., using a knowledge graph) or learned. Our work learns the skills graph using Algorithm~\ref{alg:bruteforce_graph} or Algorithm~\ref{alg:approximate_graph}, which requires initial training runs on pairs of skills or each skill, respectively. This can be made more efficient by performing these training runs on a smaller model and for fewer number of steps, but tradeoffs here have yet to be thoroughly investigated. \name also assumes that the ordered skill set is provided; as discussed in sections~\ref{sec:def} and~\ref{sec:skill_recovery}, it is challenging to recover ordered skill sets simply via metadata attributes or embedding clustering. Otherwise, the best way to sample over collections of skills that form a complete or empty graph is random or stratified sampling with no ordering to exploit. Our loss-based clustering approach presented in section~\ref{sec:skill_recovery} demonstrates that grouping by losses can provide an explanation for how skills are defined over data. An important direction for future work is to use such a clustering approach or other unsupervised algorithms in an end-to-end pipeline for skill discovery, skill graph learning, and data selection based on such skills.
\section{Additional Algorithmic Details}\label{supp:alg}

\subsection{Derivation of \name Update Rule}
First, we provide the derivation of our update rule from online mirror descent using the proximal point view~\cite{gupta2020}. We restate our optimization problem from~\eqref{eq:skillit_opt}:
\begin{align}\label{eq:skillit_opt_again}
    &\underset{p_1, \dots, p_T \in \Delta^{k-1}}{\text{minimize}} \; \frac{1}{m}\sum_{j = 1}^m \Leval{j}(f_T) \\
    &\text{s.t} \;\;\; \Leval{j}(f_t) = \Leval{j}(f_{t-1})(1 - \alpha A_{:, j}^\top p_{t-1}) \; \forall j \in [m], t = 1, \dots, T \nonumber \\
    & \;\;\;\;\;\; f_t = \Phi(f_{t-1}, p_{t-1}) \; \forall t = 1, \dots T \nonumber 
 \end{align}
 
Let $\bar{L}_t(p) = \frac{1}{m} \sum_{i = j}^m \Leval{j}(f_{t+1}) = \frac{1}{m} \sum_{i = j}^m \Leval{j}(\Phi(f_{t}, p))$; that is, $p$ is the mixture we must choose at time $t$ and $\bar{L}_t$ is the average loss per skill of the model after it is trained on $p$ at round $t$.
A greedy approximation of~\eqref{eq:skillit_opt_again} is $\underset{p \in \Delta^{k-1}}{\text{minimize}} \; \bar{L}_t(p)$, given the model and mixtures at previous rounds.
A linear approximation of $\bar{L}_{t}(p)$ is
\begin{align}\label{eq:approx}
    \bar{L}_{t}(p) \approx \bar{L}_{t}(p_{t-1}) + \langle \triangledown \bar{L}_{t-1}(p_{t-1}), p - p_{t-1}\rangle
\end{align}

Then, the problem of minimizing $\bar{L}_t(p)$ can be approximated as
\begin{align}
    \text{argmin}_{p\in \Delta^{k-1}} \langle \eta \triangledown \bar{L}_{t-1}(p_{t-1}), p \rangle
\end{align}

after we drop terms from~\eqref{eq:approx} that do not depend on $p$. Note that the $\eta$ is a constant and does not impact the solution. The optimal solution to this problem is selecting the $p$ that has the most weight on the slice with the largest gradient (e.g., a folow-the-leader sort of algorithm). To improve stability and prevent overfitting, we introduce regularization via a Bregman divergence $D_h(p || p_{t-1}) = h(p) - h(p_{t-1}) - \langle \triangledown h(p_{t-1}), p - p_{t-1} \rangle$. After dropping terms that do not contain $p$, our problem is now
\begin{align}
    \text{argmin}_{p\in \Delta^{k-1}} \langle \eta \triangledown \bar{L}_{t-1}(p_{t-1}), p \rangle + h(p) - \langle \triangledown h(p_{t-1}), p \rangle 
\end{align}

Taking the gradient and setting it equal to $0$ gives us
\begin{align}
    \eta \triangledown \bar{L}_{t-1}(p_{t-1}) + \triangledown h(p)-\triangledown h(p_{t-1}) = 0
\end{align}

Similar to in standard multiplicative weights, we set $h(p) = \sum_i p_i \ln p_i$ and $\triangledown h(p) =  [\ln p_i + 1]_i$. Then,
\begin{align}
    \ln p^{i}= \ln p_{t-1}^{i} - \eta \triangledown_i L_{t-1} (p_{t-1}) \nonumber \\
    \Rightarrow p^{i}_{t+1} = p_{t}^{i} \exp( -\eta\triangledown_i \bar{L}_{t} (p_{t}))\label{eq:gradient_1}
\end{align}

where $\triangledown_i$ is the $i$th element of the gradient. 
Now we wish to compute $\triangledown_i \bar{L}_{t}(p_{t}) = \frac{1}{m} \sum_{j = 1}^m \triangledown_i [\Leval{j}(f_{t+1})]  = \newline \frac{1}{m} \sum_{j = 1}^m \triangledown_i [\Leval{j}(\Phi(f_t, p_t))]$. Recall the dynamics model for $L_{\text{eval}}$:
\begin{align}
    \Leval{j}(f_{t+1})=\Leval{j}(f_t)(1- A_{:, j}^\top p_{t}),
\end{align}

The gradient of this model with respect to each training skill $s_i$ is
\begin{align}
    &\triangledown_i \Leval{j}(f_{t+1}) = - A_{ij}\Leval{j}(f_t) \\
    \Rightarrow & \triangledown_i \bar{L}_{t}(p_t) = \frac{1}{m} \sum_{j = 1}^m - A_{ij} \Leval{j}(f_t)  \nonumber 
\end{align}

Plugging this back into~\eqref{eq:gradient_1},
\begin{align}
    p_{t+1}^i= p_t^i \exp \bigg(\eta \sum_{j = 1}^m A_{ij} \Leval{j}(f_t)\bigg),
\end{align}

where we can absorb the $\frac{1}{m}$ into $\eta$.

\subsection{Graph Learning Method}\label{supp:graph_learning}

\begin{algorithm}[tb]
   \caption{\textsc{LearnGraph} (Brute-Force)}
\begin{algorithmic}[1]
   \STATE {\bfseries Input:} Ordered skill set $\mathcal{S} = \{s_1, \dots, s_k\}$. Number of training steps $H$, base model $f$.
   \FOR{$j \in [k]$}
        \STATE Train $f$ on samples from $\X_{s_j}$ for $H$ steps and denote $f_{H, j}$ to be the model after training.
        \STATE Observe change in loss, $\delta_j^j = \Leval{j}(f) - \Leval{j}(f_{H, j})$.
   \ENDFOR
   \FOR{$i, j \in [k]$}
        \STATE Train $f$ on samples from $\X_{s_i} \cup \X_{s_j}$ for $H$ steps and denote $f_{H, i, j}$ to be the model after training.
        \STATE Observe change in loss, $\delta_j^{i, j} = \Leval{j}(f) - \Leval{j}(f_{H, i, j})$. 
        \IF{$\delta_j^{ij} > \delta_j^j$}
            \STATE Draw edge $s_i \rightarrow s_j$ and set $A_{ij} > 0$. 
        \ENDIF
   \ENDFOR
   \STATE \textbf{Return} Adjacency matrix $A \in \R^{k \times k}$
\end{algorithmic}
\label{alg:bruteforce_graph}
\end{algorithm}

\begin{algorithm}[tb]
   \caption{\textsc{LearnGraph} (Approximate)}
\begin{algorithmic}[1]
   \STATE {\bfseries Input:} Ordered skill sets $\strainset$ and $\sevalset$. Number of training steps $H$, base model $f$.
   \FOR{$i \in [k]$}
        \STATE Train $f$ on samples from $\X_{\strain{i}}$ for $H$ steps and denote $f_{H, i}$ to be the model after training.
        \FOR{$j \in [m]$}
        \STATE Observe change in loss, $\delta_j^i = \Leval{j}(f) - \Leval{j}(f_{H, i})$.
        \STATE If $\delta_j^i > 0$, draw edge $\strain{i} \rightarrow \seval{j}$ and set $A_{ij} > 0$.
        \ENDFOR
   \ENDFOR
   \STATE \textbf{Return} Bipartite adjacency submatrix $A \in \R^{k \times m}$
\end{algorithmic}
\label{alg:approximate_graph}
\end{algorithm}

We provide algorithms for learning the graph over an ordered skill set. In Algorithm~\ref{alg:bruteforce_graph}, we discuss the brute-force approach for learning the adjacency matrix. This approach only works when $\sevalset \subseteq \strainset$ (e.g. pre-training and fine-tuning cases), so we denote $\mathcal{S} = \strainset$ in the algorithm box. In Algorithm~\ref{alg:approximate_graph}, we discuss the linear approach for learning the adjacency matrix. This approach works even in the out-of-domain case when $\sevalset$ and $\strainset$ are disjoint.

In both approaches, the exact value of $A_{ij}$ can vary, but we can typically set it proportional to $\delta_{j}^{i, j} - \delta_j^j$, the difference between the changes in loss, in the brute-force case or $\delta_j^i$, the change in loss itself, in the approximate case. The exact constructions and methods for learning each $A$ in our experiments are in Appendix~\ref{supp:skill_graphs}.
\section{Additional Experimental Details} \label{supp:details}

\subsection{Datasets}

We present details about each dataset used, including information on the skills and the validation dataset. A summary is presented in Table~\ref{tab:datasets}.

\begin{table}[h]
\small
    \caption{We list each dataset used as well as its corresponding skill. We include the number of skills in the training dataset, as well as details on how the validation dataset is constructed.}
    \centering
    \begin{tabular}{c|c |c | c}
        \toprule 
        Dataset & Skill & $\#$ skills & Validation data  \\
        \midrule 
        Alpaca & Instruction type & $38$ &  $50$ samples per skill \\
        Pile of Law & Legal data source &  $31$ & $645$ samples per skill \\
        LEGO & Reasoning chain depth & $5$ & $100$ samples per skill \\
        Addition & Digit & $3$ &  $100$ samples per skill \\
        NI (pre-training) & Task category & $23$  & $50$ samples per task \\
        NI (Spanish QG) & Task category $\times$ language & $4$  & $100$ samples per task  \\
        NI (stance detection) & Task category & $2$  & $50$ samples per task \\
        NI (out-of-domain) & Task category & $59, 12$ & $400$ samples per task \\
        RedPajama & Data source & $7$ & LM eval harness \\
    \end{tabular}
    \label{tab:datasets}
\end{table}

\begin{itemize}[itemsep=0.05pt,topsep=0pt,leftmargin=*]
    \item Alpaca dataset~\cite{alpaca}: the Alpaca dataset consists of $52$K instruction examples that were generated from text-davinci-003. We applied the Berkeley Neural Parser~\cite{kitaev-etal-2019-multilingual, kitaev-klein-2018-constituency} to each instruction, keeping $40777$ samples it was able to parse successfully. If the sample began with a question, we annotated it with the skill ``question'', and otherwise we annotated it with the verb identified from the parser. We grouped the data into a total of $38$ skills, such as "list", "edit", "calculate", "describe" and "identify".

    \item Pile of Law~\cite{hendersonkrass2022pileoflaw}: the Pile of Law dataset consists of various sources of legal and administrative data, ranging from tax rulings to the world's constitutions. We evaluate on a subset of the Pile of Law validation dataset consisting of $13883$ samples, where we selected max($645$, source size) samples per source. We truncated each sample to be no more than $100$K characters.

    \item LEGO~\cite{zhang2022unveiling}: for the LEGO synthetic, we set $k = 5$ and sample $192000$ points across the skills. Our validation dataset consisted of $100$ samples per skill.

    \item Addition: for the $3$-digit addition synthetic, we set $k = 3$ and sample $192000$ points across the skills. We use a validation dataset of $100$ samples per skill.

    \item Natural Instructions~\cite{wang2022supernaturalinstructions, naturalinstructions}: the Natural Instructions dataset is a large collection of tasks and their definitions in natural language. For the pre-training setting, we used a set of $23$ task categories that had the largest degree (in-degree + out-degree) in the learned skills graph, for a total of $1,232,437$ samples and $425$ tasks to select from. We evaluated on $50$ samples per task. 

    For the fine-tuning setting with Spanish question generation, we select data over $4$ skills (Spanish question generation, Spanish question answering, English question generation, English question answering) for a total of $513210$ samples and $212$ tasks to select from. We evaluated on $100$ samples per task.

    For the fine-tuning setting with stance detection, we select data over $2$ skills (stance detection, text matching) for a total of $50990$ samples and $19$ tasks to select from. We evaluated on $50$ samples per task. 

    For the out-of-domain setting, we select data over all $59$ task categories for a total of $2,417,867$ samples and $753$ tasks to select from. The test split consisted of $12$ task categories and $119$ tasks, and we evaluated on min($400$, task size) samples per task.

    \item RedPajama~\cite{redpajama}: the RedPajama dataset is a $1$-trillion token dataset that aims to reproduce the LLaMA~\cite{touvron2023llama} training dataset. We select over the $7$ data sources and evaluate using the LM evaluation harness~\cite{eval-harness}.
\end{itemize}

\subsection{Graph Learning Details} \label{supp:skill_graphs}

\begin{figure}
    \centering
    \includegraphics[width=4in]{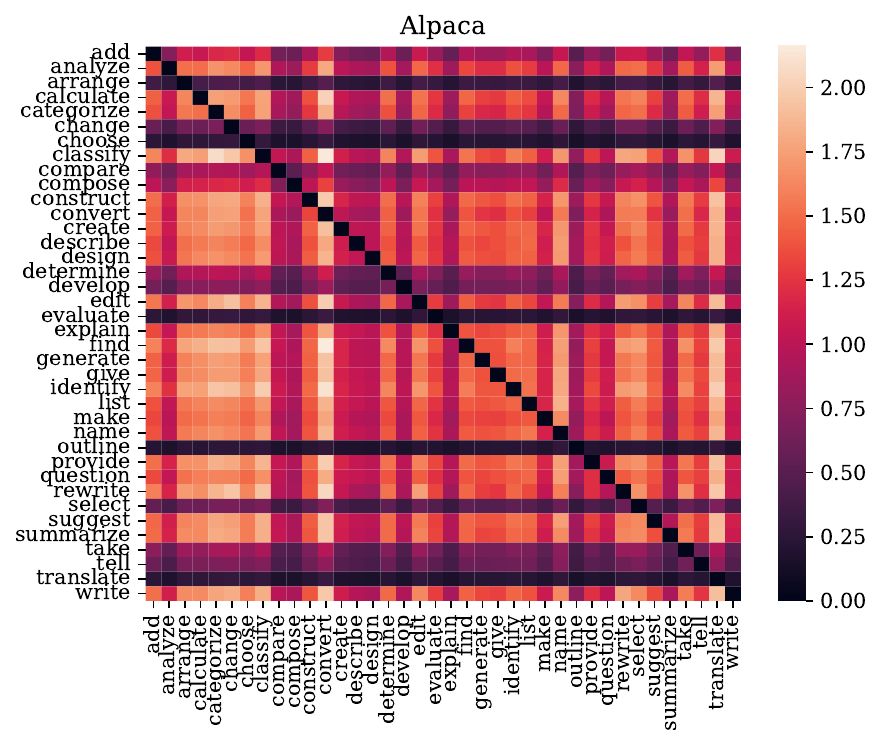}
    \caption{Alpaca heatmap where $i, j$th entry is $\max(0, \delta_j^i)$ (the change in loss on $s_j$ after training on $s_i$ for $150$ steps). Diagonal entries are set to $0$ for clearer visualization.}
    \label{fig:alpaca_heatmap}
\end{figure}

\begin{figure}
    \centering
    \includegraphics[width=\textwidth]{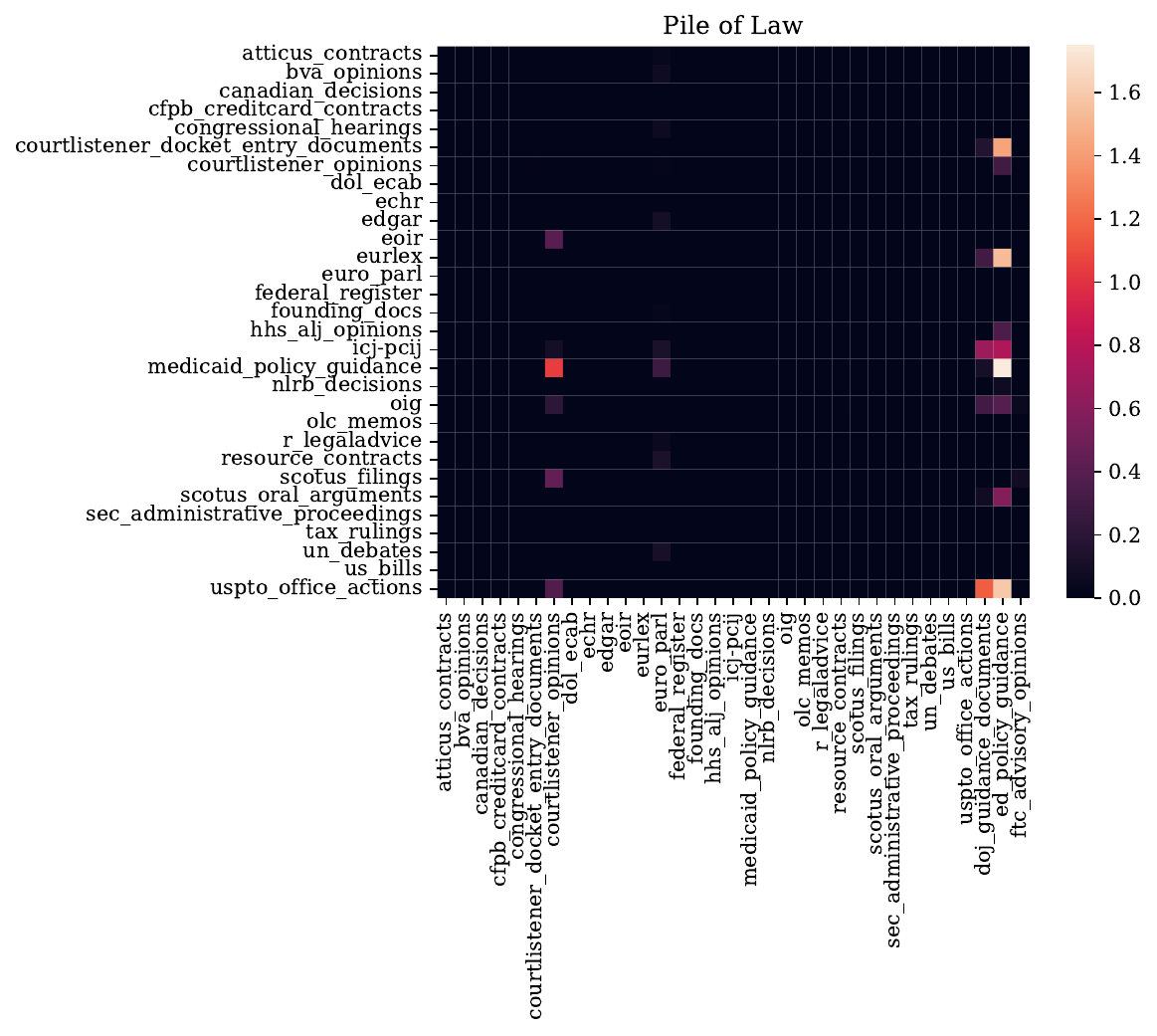}
    \caption{Pile of Law heatmap where $i, j$th entry is $\max(0, \delta_j^i)$ (the change in loss on $s_j$ after training on $s_i$ for $150$ steps). Diagonal entries are set to $0$ for clearer visualization.}
    \label{fig:pol_heatmap}
\end{figure}

\begin{figure}
    \centering
    \includegraphics[width=0.5\textwidth]{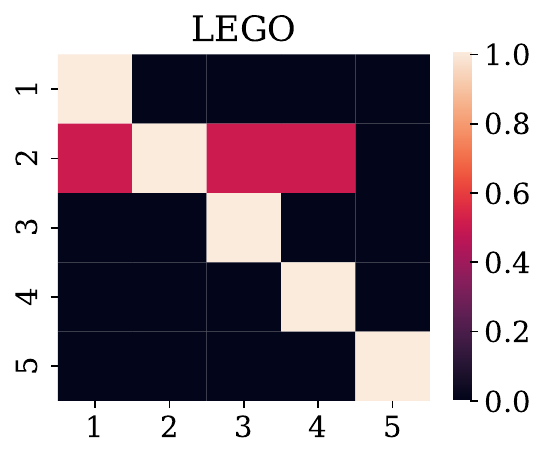}
    \caption{LEGO heatmap with $k=5$ where $i, j$th entry is set to $0.5$ if the number of steps needed to reach $0.01$ loss on skill $j$ when training on a balanced mixture of skills $i$ and $j$ is less than when training on skill $j$ only.}
    \label{fig:lego_heatmap}
\end{figure}

\begin{figure}
    \centering
    \includegraphics[width=0.4\textwidth]{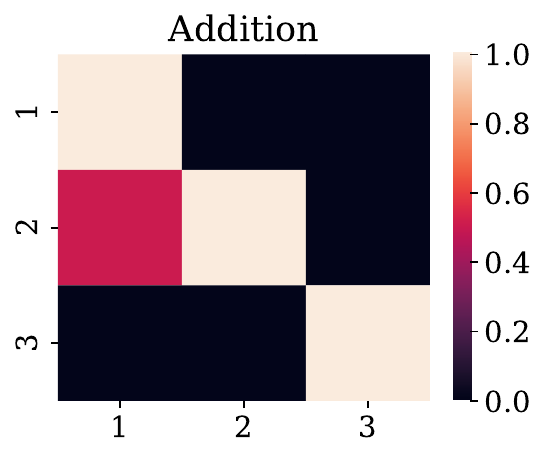}
    \caption{Addition heatmap with $k=3$ where $i, j$th entry is set to $0.5$ if the number of steps needed to reach $0.01$ loss on skill $j$ when training on a balanced mixture of skills $i$ and $j$ is less than when training on skill $j$ only.}
    \label{fig:addition_heatmap}
\end{figure}

\begin{figure}
    \centering
    \includegraphics[width=\textwidth]{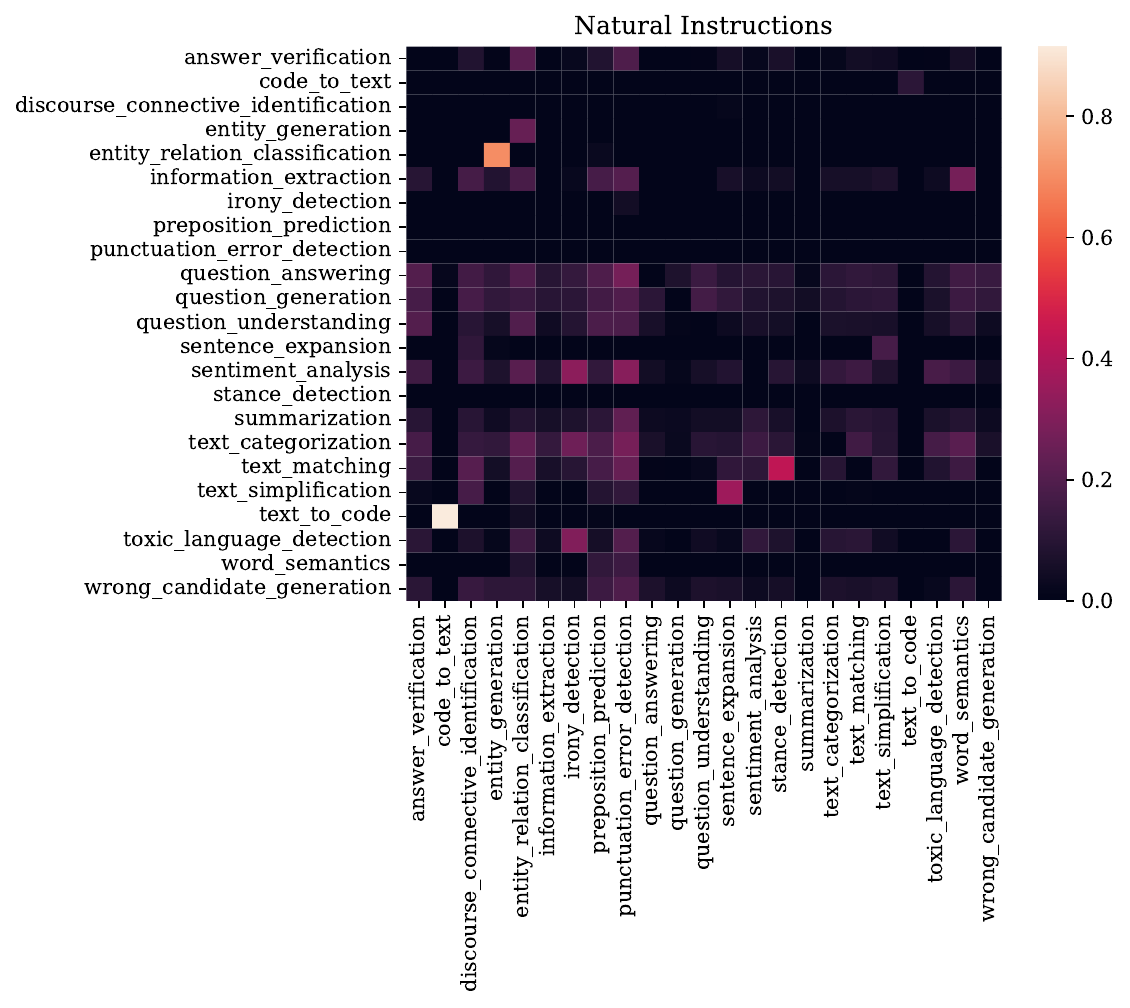}
    \caption{Natural Instructions heatmap where $i, j$th entry is $\max(0, \delta_j^i)$ (the change in loss on $s_j$ after training on $s_i$ for $100$ steps). Diagonal entries are set to $0$ for clearer visualization.}
    \label{fig:ni_heatmap}
\end{figure}

\begin{figure}
    \centering
    \includegraphics[width=0.4\textwidth]{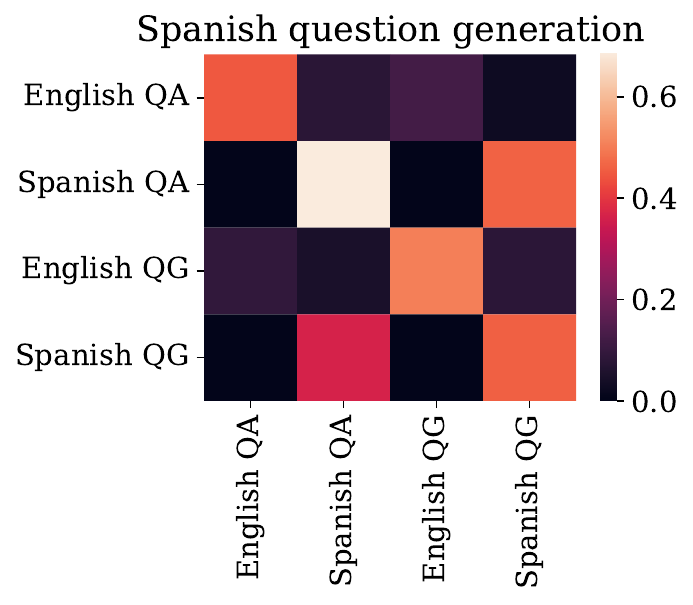}
    \includegraphics[width=0.4\textwidth]{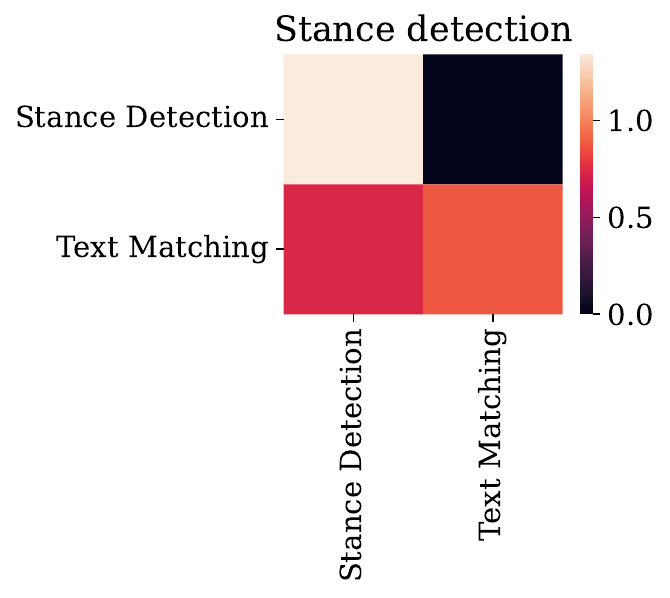}
    \caption{Spanish question generation and stance detection heatmaps where $i, j$th entry is $\max(0, \delta_j^i)$ (the change in loss on $s_j$ after training on $s_i$ for $100$ steps).}
    \label{fig:ni_ft_heatmap}
\end{figure}

\begin{figure}
    \centering
    \includegraphics[width=0.5\textwidth]{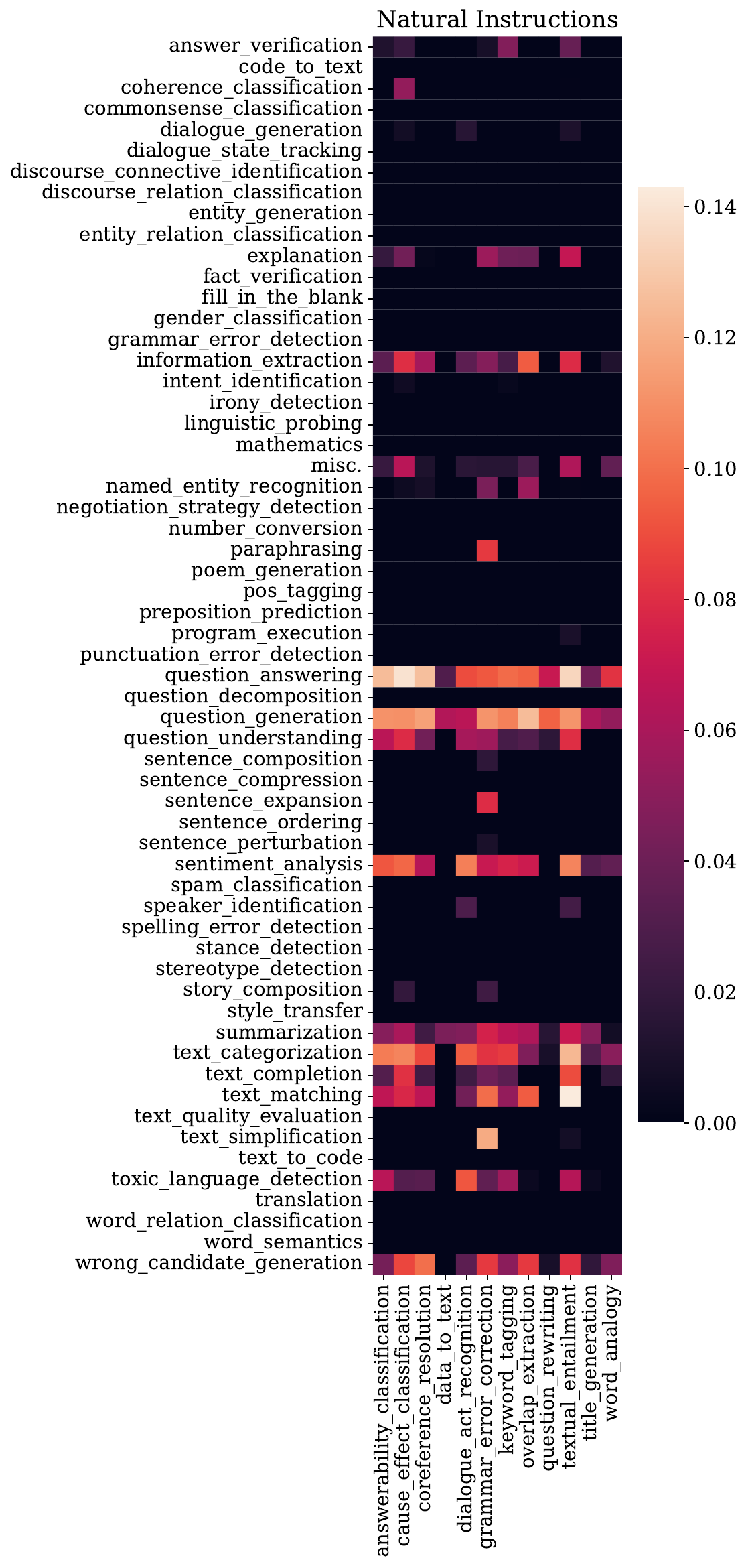}
    \caption{Natural Instructions heatmap for out-of-domain setting where rows are for the training skills and columns are for the evaluation skills. The $i, j$th entry is $\max(0, \delta_j^i)$ (the change in loss on $s_j$ after training on $s_i$ for $100$ steps). }
    \label{fig:ni_test_heatmap}
\end{figure}

\begin{figure}
    \centering
    \includegraphics[width=0.4\textwidth]{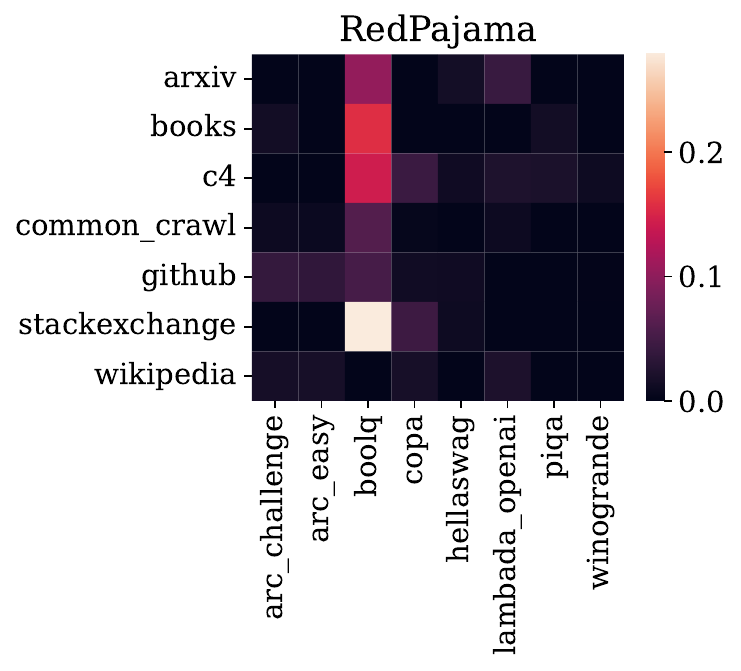}
    \caption{RedPajama heatmap for out-of-domain setting where rows are for the training skills and columns are for the evaluation skills. The $i, j$th entry is $\max(0, \delta_j^i)$ (the change in perplexity on $s_j$ after training on $s_i$ for $1$B tokens). }
    \label{fig:rp_heatmap}
\end{figure}

We describe how the skills graph was learned on each dataset. 
\begin{itemize}[itemsep=0.05pt,topsep=0pt,leftmargin=*]
    \item Alpaca (Figure~\ref{fig:alpaca_heatmap}): we use Algorithm~\ref{alg:approximate_graph} and train for $K = 150$ steps per skill. Each edge $i \rightarrow j$ has a weight of $\delta_j^i$, the difference in loss on skill $j$ before and after training on $i$. Next, we compare the average validation loss of skill-stratified sampling versus random sampling when we train for $K = 1000$ steps. We find that skill-stratified sampling only does 0.007 better than random sampling, confirming that Alpaca's dense skills graph suggests that random sampling is the best we can do.  
    \item Pile of Law (Figure~\ref{fig:pol_heatmap}): we use Algorithm~\ref{alg:approximate_graph} and train for $K = 150$ steps. Each edge $i \rightarrow j$ has a weight of $\delta_j^i$, the difference in loss on skill $j$ before and after training on $i$.
    \item LEGO (Figure~\ref{fig:lego_heatmap}): we use both Algorithm~\ref{alg:bruteforce_graph} and Algorithm~\ref{alg:approximate_graph} and train for $K=6000$ steps each. Each edge $i \rightarrow j$ has a weight of $0.5$ if the amount of data associated with skill $j$ that is needed to reach $0.01$ validation loss is less when training on $(i, j)$ than on $j$ (edges are set to $0$ if $0.01$ validation loss is not reached, even if loss is decreasing). Each edge $i \rightarrow j$ is also set to $0.5$ if training on $i$ decreases loss directly on $j$. We set each diagonal entry of $A$ to be $1$.
    \item Addition (Figure~\ref{fig:addition_heatmap}): we use Algorithm~\ref{alg:bruteforce_graph} and train for $K = 6000$ steps. Each edge $i \rightarrow j$ has a weight of $0.5$ if the amount of data associated with skill $j$ that is needed to reach $0.01$ validation loss is less when training on $(i, j)$ than on $j$ (edges are set to $0$ if $0.01$ validation loss is not reached, even if loss is decreasing). We set each diagonal entry of $A$ to be $1$.
    \item Natural Instructions (Figure~\ref{fig:ni_heatmap},~\ref{fig:ni_ft_heatmap},~\ref{fig:ni_test_heatmap}): we use Algorithm~\ref{alg:approximate_graph}. For the pre-training setting, we train for $K = 600$ steps and assign each edge $i \rightarrow j$ a weight $\delta_j^i$ equal to the change in loss on $j$ in the first $100$ steps for all $i, j \in [k]$, including diagonal entries. For the fine-tuning setting, we train for $K = 600$ steps and assign each edge $i \rightarrow j$ a weight $\delta_j^i$ equal to the change in loss before and after training. For the out-of-domain setting, we train for $K = 600$ steps and assign each edge $i \rightarrow j$ a weight $\delta_j^i$ equal to the change in loss before and after training in the first $100$ steps.
    \item RedPajama (Figure~\ref{fig:rp_heatmap}): we use Algorithm~\ref{alg:approximate_graph} and train for $1$ billion tokens per data source. We assign each edge $i \rightarrow j$ a weight $\delta_j^i$ equal to the change in perplexity on the validation datalsoa before and after training.
\end{itemize}

\subsection{Training Details}

We describe the parameters used for \name.

\paragraph{\name pre-training}

\begin{itemize}[itemsep=0.05pt,topsep=0pt,leftmargin=*]
    \item LEGO: $\eta = 0.5, T = 6, w = 3$. We train for $6000$ steps.
    \item Addition: $\eta = 0.1, T = 5, w = 3$. We train for $6000$ steps.
    \item Natural Instructions (pre-training): $\eta = 0.2, T = 1$. We train for $5000$ steps.
\end{itemize}

For the LEGO random baseline, when we selected points at random, we used an imbalanced training dataset with proportions 1:1:1:3:5.
For the addition random baseline, we used an imbalanced dataset with randomly selected proportions: 13:14:18. 
For the curriculum learning baselines, the pacing function, $g(i)$, denotes the size of the subset of the highest scoring samples that we uniformly select from in the $i$th epoch. We define our pacing function as $g(i) = \frac{iH}{M}$, where $H$ is the number of steps and $M$ is $5$ epochs for LEGO and NI, and $3$ for addition.

\paragraph{\name fine-tuning}

\begin{itemize}[itemsep=0.05pt,topsep=0pt,leftmargin=*]
    \item LEGO: $\eta = 0.5, T = 10, w = 3$. We train for $6000$ steps.
    \item Addition: $\eta = 0.1, T = 5, w = 3$. We train for $6000$ steps.
    \item Natural Instructions (Spanish QG): $\eta = 0.8, T = 6, w = 3$. We train for $600$ steps.
    \item Natural Instructions (stance detection): $\eta = 0.2, T = 6, w = 3$. We train for $600$ steps.
\end{itemize}

\paragraph{\name out-of-domain}

\begin{itemize}[itemsep=0.05pt,topsep=0pt,leftmargin=*]
    \item Natural Instructions: $\eta = 0.2, T = 10, w = 3$. We train for $5000$ steps.
    \item RedPajama: $\eta = 100, T = 1$. We train for $3$ billion tokens.
\end{itemize}

All results are computed over $5$ random seeds.

Batch sizes of $32$ and $64$ were used for the LEGO and addition synthetic on the 125M and 1.3B parameter model, respectively.
Batch sizes of $4$ and $16$ were used for the Natural Instructions experiments on the 125M and 1.3B parameter model. 

For the out-of-domain Natural Instructions experiment and Alpaca graph learning experiments, a learning rate of 5e-6 with linear scheduler and $50$ warmup steps was used. For the Natural Instructions continual pre-training experiment on the 1.3B parameter model, a learning rate of 1e-6 was used. All other experiments used a learning rate of 5e-5. All experiments used AdamW with betas = 0.9, 0.999, eps = 1e-8, and weight decay = $0.01$. A context window of $512$ was used for all experiments except LEGO and addition, which used a window of $128$.

Experiments with the Addition dataset were run using an Nvidia RTX A6000. 
Other experiments using the GPT-Neo 125M parameter model were run on an Nvidia Tesla P100. Experiments using the GPT-Neo 1.3B parameter model were run on an Nvidia Tesla A100.

\section{Additional Experimental Results}

\subsection{Additional examples of LEGO ordered skill sets} \label{supp:lego_exp}

\begin{figure}
    \centering
    \includegraphics[height=2in]{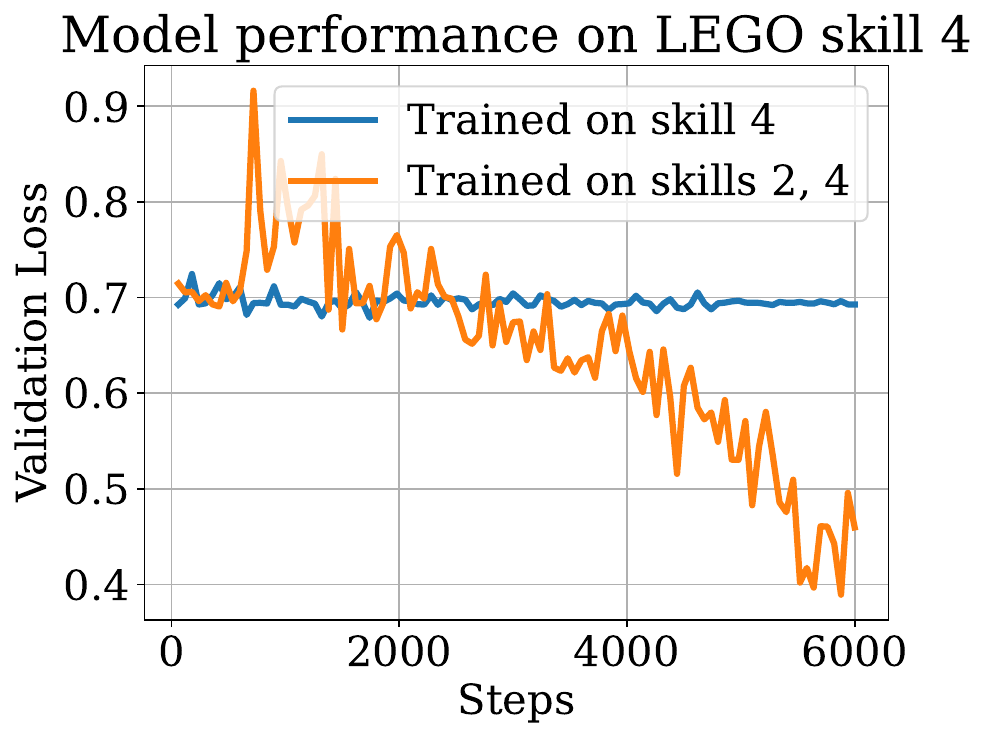}
    \includegraphics[height=2in]{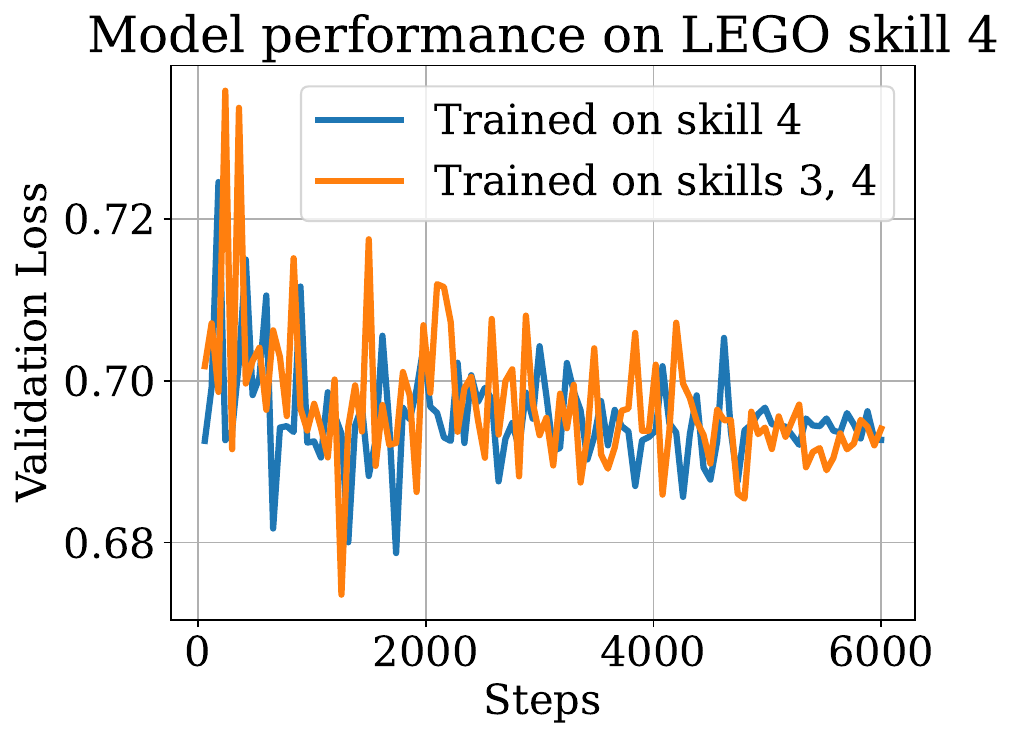}
    \caption{Performance on LEGO skill 4 when training on skill 4, skills 2 and 4, and skills 3 and 4. Even though skill $3$ and skill $4$ share an edge in the LEGO synthetic's underlying reasoning chain (i.e. a model predicting correct for the fourth variable is one extra step beyond predicting correct for the third variable), we find that training on skills $2$ and $4$ helps improve performance on skill $4$ more.}
    \label{fig:more_lego}
\end{figure}

For the LEGO synthetic, it may appear obvious that the skills graph is equivalent to the reasoning chain over the variables. However, in Figure~\ref{fig:more_lego} we see that this is not the case. Training on skills $2$ and $4$ together results in lower loss on skill $4$ than when trained on skill $4$ alone. However, training on skills $3$ and $4$ together results in roughly the same loss on skill $4$ as when training on skill $4$ alone, even though skill $3$ and skill $4$ share an edge in the LEGO synthetic's underlying reasoning chain. This suggests that our intuition for how skills influence each other does not always match how the model learns skills.

Next, we consider a slightly more complex reasoning pattern on the LEGO synthetic. Instead of a chain, we construct a tree, where two variables in the LEGO synthetic are both defined in terms of the same parent variable. For example,
\begin{quote}
    Input:  c = val 1, y = not w, v = val c,  w = not c. Output: y = 1.
\end{quote}

\begin{figure}
    \centering
    \includegraphics[height=1.8in]{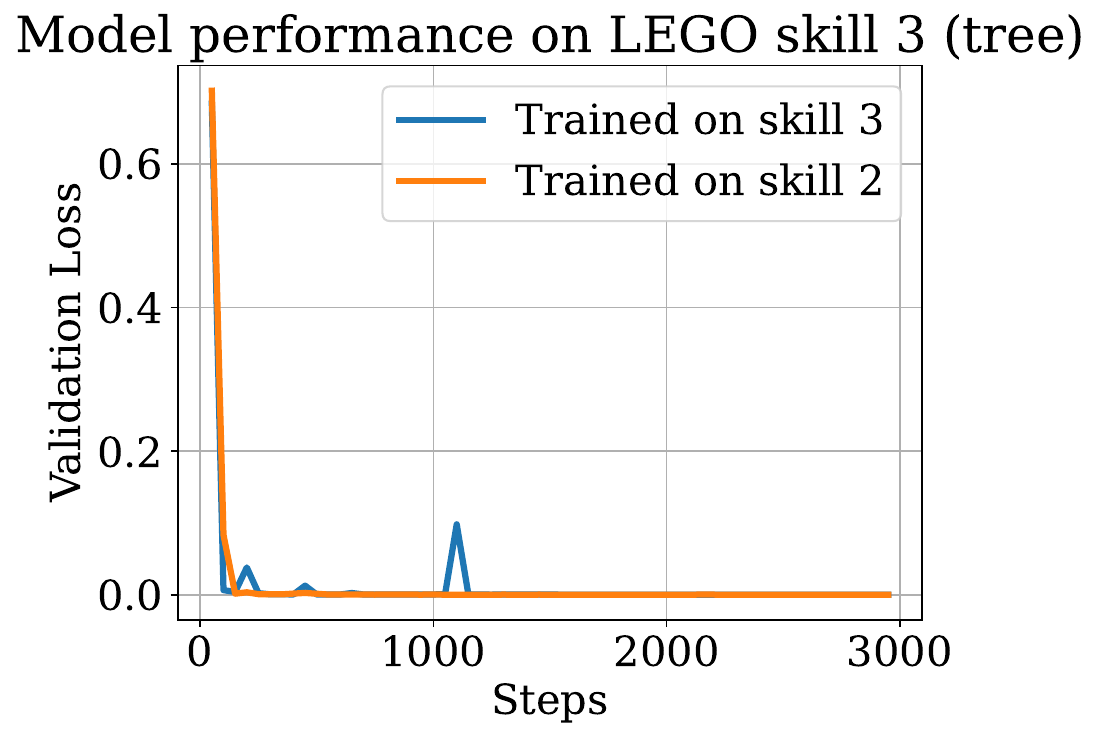}
    \includegraphics[height=1.8in]{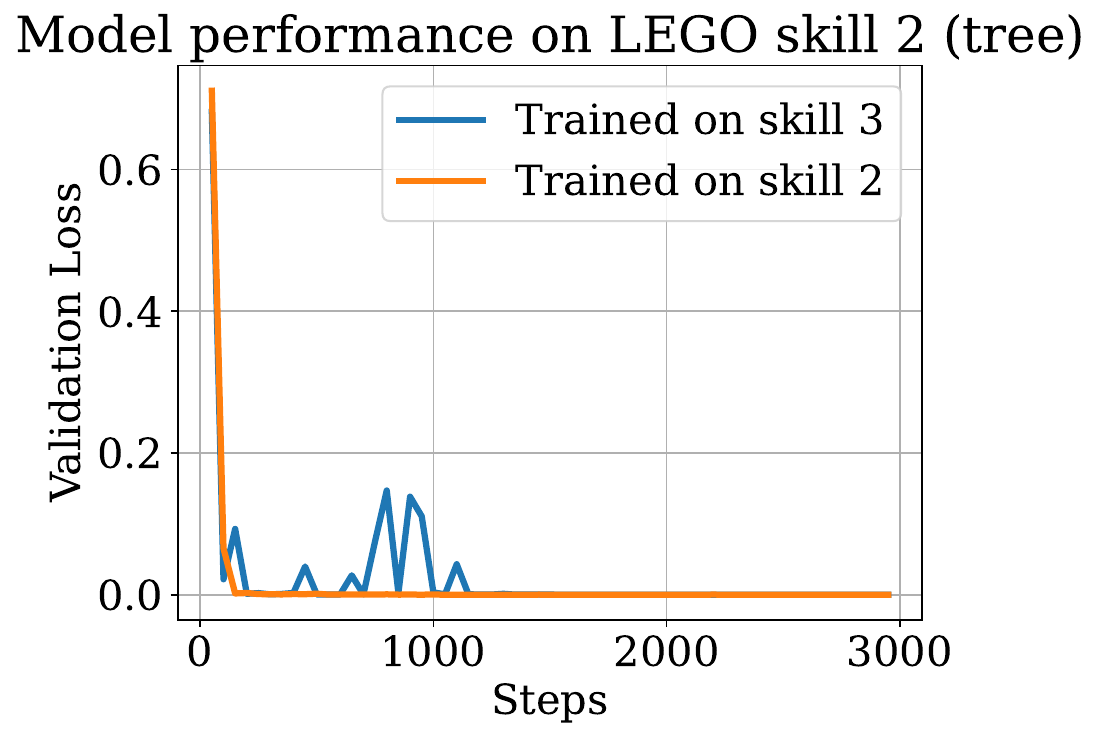}
    \caption{Performance on LEGO skill $2$ and $3$ when training on skills $2$ and $3$. The reasoning pattern is a tree rather than a chain over $k = 4$ variables. Skills $2$ and $3$ are at the same ``depth'' in the graph and both depend on skill $1$, so there is positive influence between the skills despite there being no edge between $2$ and $3$ in the LEGO reasoning graph.}
    \label{fig:lego_tree_23}
\end{figure}

\begin{figure}
    \centering
    \includegraphics[height=1.8in]{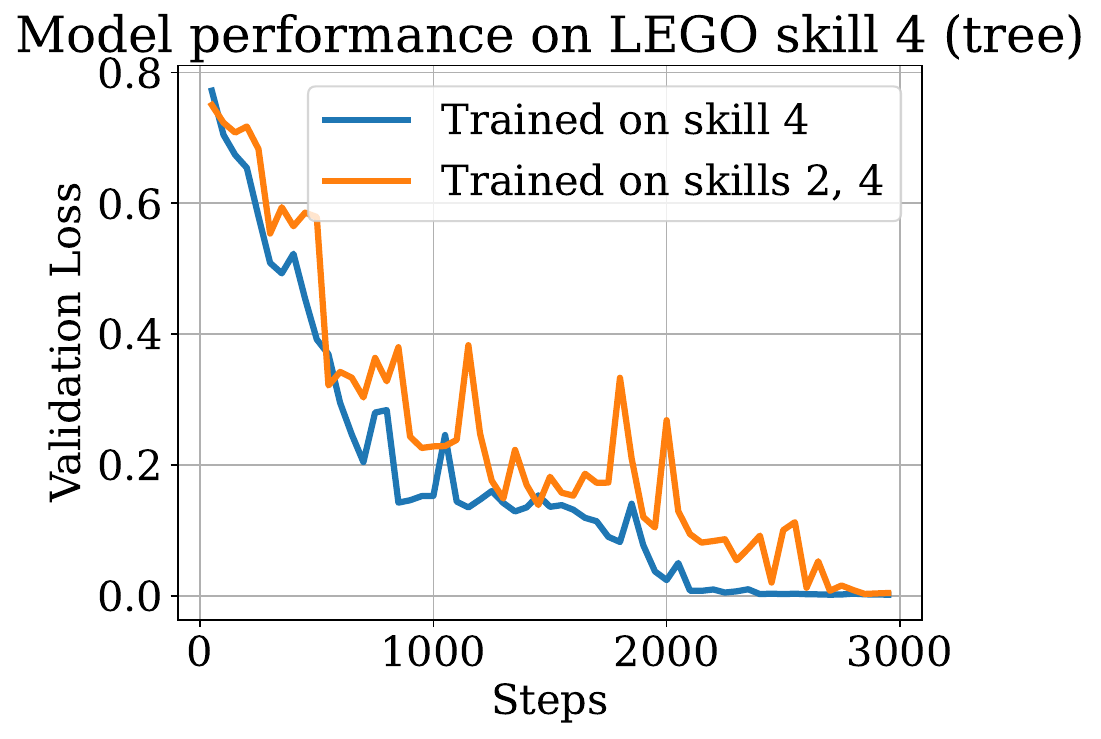}
    \includegraphics[height=1.8in]{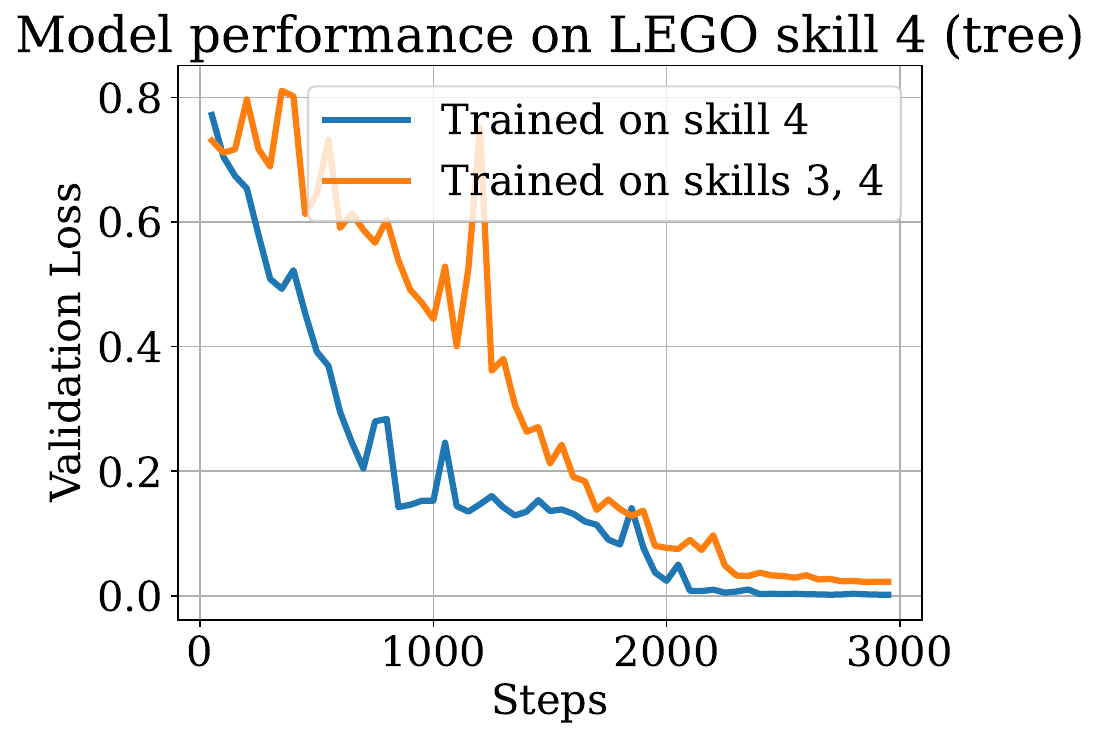}
    \caption{Performance on LEGO skill $4$ when training on skills $2, 4$ and skills $3, 4$. We find that in both cases, the benefit from training on additional skills is minor. For instance, training on $2$ and $4$ reaches $0.01$ loss in $2700$ steps, while training on $4$ only reaches it in $2100$ steps.}
    \label{fig:lego_tree_4}
\end{figure}

In this example, $k = 4$ and both $v$ and $w$ are written in terms of $c$, and the reasoning graph has edges $1 \rightarrow 2$, $1 \rightarrow 3$, $2 \rightarrow 4$. In this case, we see that training on skill $2$ or skill $3$ both improve losses on skills $2$ and $3$ (Figure~\ref{fig:lego_tree_23}). However, unlike the previous figures, training on skills $2$ and $4$ or skills $3$ and $4$ do not significantly help reduce loss on skill $4$ (Figure~\ref{fig:lego_tree_4}). Again, these measurements demonstrate that the reasoning graph does not necessarily equal the skills graph.

\subsection{Unsupervised skill recovery} \label{supp:skill_clustering}

We explore several clustering techniques for recovering the skills in the LEGO synthetic on the validation dataset. Our results are shown in Table~\ref{tab:skill_recovery}.

We first cluster based on the pre-trained model embeddings of the last token and the average token. We also report accuracies of clustering based on the trained model embedding's last token, where we train the model using random sampling for $6000$ steps, and clustering based on Sentence-BERT embeddings. Among these four methods, using the trained model embeddings has the highest accuracy of $38.4$ points. 

Next, we cluster points based on losses. In particular, we do $10$ runs, each for $6000$ steps and with a randomly sampled mixture of skills. For each run, we evaluate the model on the validation dataset at $120$ checkpoints. Then, each sample in the validation dataset has $1200$ losses associated with it, comprising a feature vector for that sample. We perform k-means clustering on these features, which has an accuracy of $61.0$ points, significantly higher than the second best accuracy of $38.4$. 

\begin{table}
    \small
    \centering
\caption{Clustering-based skill recovery methods on the LEGO dataset. The validation dataset we cluster consists of $500$ points with $k=5$, and results are reported over 10 runs of k-means.}
\begin{tabular}{|c|c|}
    \hline
    Cluster method & Accuracy \\
    \hline
    Pretrained embedding of last token & $24.8 \pm 0.5$ \\
    \hline 
    Pretrained embedding of average token & $25.2 \pm 1.1$ \\
    \hline 
    Trained model embedding of last token & $38.4 \pm 0.8$ \\ 
    \hline 
    Sentence-BERT embedding & $23.9 \pm 0.7$ \\
    \hline
    Losses over multiple runs & $\mathbf{61.0 \pm 1.6}$ \\
    \hline
\end{tabular}
\label{tab:skill_recovery}
\end{table}

\subsection{Full results for Section~\ref{sec:exp}}

\subsubsection{Per-skill performance}\label{supp:full}

In this section, we provide tables containing the per skill break-down of our results from Section~\ref{sec:exp}.

\paragraph{Continual Pre-training}

In the continual pre-training setting, we report two additional baselines that combine curriculum learning with skills. Curriculum learning has been proposed for multitask learning~\cite{varshney2022let}, in which groups of data are ranked by their average score and then trained in order of this ranking (with mixing of previously seen groups to avoid forgetting). We construct two baselines, Skill-curriculum and Skill-anticurriculum, using Algorithm 1 from~\cite{varshney2022let}. In contrast to the random baseline which has imbalanced skills, this approach has knowledge of skills and thus uses a skill-stratified training dataset to sample from. We set the fraction of the previous group to be $\text{frac}=0.4$, as we found that setting $\text{frac}=0.0$ resulted in forgetting.

\begin{figure}
    \centering
    \includegraphics[width=0.9\textwidth]{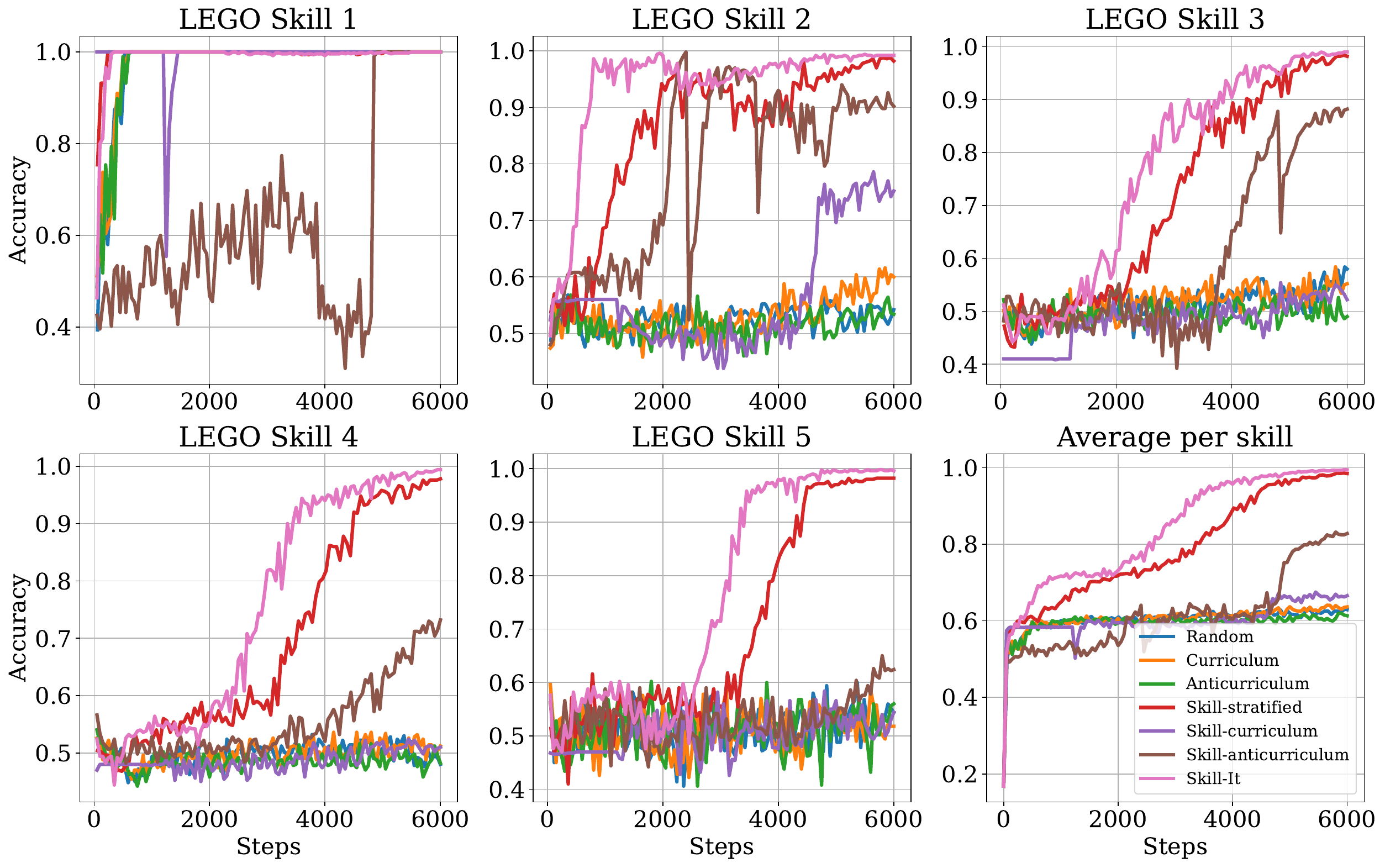}
    \caption{Accuracy of \name on each skill (binary classification) on the LEGO synthetic in the continual pre-training setting. \name attains higher accuracy more quickly than baselines that both do and do not utilize the notion of skills.}
    \label{fig:lego_accs}
\end{figure}

We report loss per skill for the LEGO synthetic in Table~\ref{tab:lego_pt_loss}, which corresponds to the results in Figure~\ref{fig:lego_mw}. We report accuracy per skill in Table~\ref{tab:lego_pt_acc} and Figure~\ref{fig:lego_accs}.
We report the loss per skill for the Addition synthetic in Table~\ref{tab:addition_pt_loss}, which also correspond to to the results in Figure~\ref{fig:lego_mw}. 
Finally, we report validation loss per task category for the Natural Instructions continual pre-training experiment in Table~\ref{tab:ni_pretraining}, where we find that \name outperforms random sampling by $3.2\%$ on average across skills.

\begin{table}[h]
\footnotesize
\caption{Results on validation loss per skill for LEGO pre-training experiment, averaged over $5$ random seeds.}
\begin{adjustbox}{center}
\begin{tabular}{c|cccccc}
\toprule
& Skill 1 & Skill 2 & Skill 3 & Skill 4 & Skill 5 & Average \\
\midrule
Random & $0_{\pm 0.000}$ & $0.675_{\pm 0.041}$ & $0.688_{\pm 0.008}$ & $0.673_{\pm 0.049}$ & $0.667_{\pm 0.056}$ & $0.541_{\pm 0.031}$ \\
Curriculum & $0_{\pm 0.000}$ & $0.645_{\pm 0.052}$ & $0.686_{\pm 0.018}$ & $0.674_{\pm 0.042}$ & $0.671_{\pm 0.0459}$ & $0.535_{\pm 0.029}$ \\
Anticurriculum & $0_{\pm 0.000}$ & $0.690_{\pm 0.003}$ & $0.695_{\pm 0.004}$ & $0.693_{\pm 0.003}$ & $0.689_{\pm 0.004}$ & $0.554_{\pm 0.001}$ \\
Skill-stratified & $0_{\pm 0.000}$ & $0.045_{\pm 0.036}$ & $0.056_{\pm 0.029}$ & $0.079_{\pm 0.044}$ & $0.050_{\pm 0.025}$ & $0.046_{\pm 0.022}$ \\
Skill-curriculum & $0_{\pm 0.000}$ & $0.484_{\pm 0.200}$ & $0.698_{\pm 0.027}$ & $0.697_{\pm 0.010}$ & $0.689_{\pm 0.007}$ & $0.514_{\pm 0.040}$ \\
Skill-anticurriculum & $0.001_{\pm 0.001}$ & $0.174_{\pm 0.118}$ & $0.245_{\pm 0.091}$ & $0.443_{\pm 0.125}$ & $0.566_{\pm 0.118}$ & $0.286_{\pm 0.060}$ \\
\name & $0_{\pm 0.000}$ & $0.002_{\pm 0.002}$ & $0.024_{\pm 0.031}$ & $0.013_{\pm 0.010}$ & $0.022_{\pm 0.021}$ & $0.012_{\pm 0.008}$ \\
\bottomrule
\end{tabular}
\end{adjustbox}
\label{tab:lego_pt_loss}
\end{table}

\begin{table}[h]
\footnotesize
\centering
\caption{Results on accuracy per skill (binary classification) for LEGO pre-training experiment, averaged over $5$ random seeds.}
\begin{adjustbox}{center}
\begin{tabular}{c|cccccc}
\toprule
& Skill 1 & Skill 2 & Skill 3 & Skill 4 & Skill 5 & Average \\
\midrule
Random & $100.0_{\pm 0.0}$ & $54.2_{\pm 5.9}$ & $58.0_{\pm 3.1}$ & $48.0_{\pm 6.3}$ & $54.4_{\pm 7.3}$ & $62.9_{\pm 3.5}$ \\
Curriculum & $100.0_{\pm 0.0}$ & $60.0_{\pm 10.6}$ & $55.2_{\pm 5.8}$ & $51.2_{\pm 6.3}$ & $51.8_{\pm 6.1}$ & $63.6_{\pm 3.6}$ \\
Anticurriculum & $100.0_{\pm 0.0}$ & $53.4_{\pm 2.3}$ & $49.0_{\pm 4.8}$ & $48.2_{\pm 6.4}$ & $56.0_{\pm 5.7}$ & $61.3_{\pm 2.2}$ \\
Skill-stratified & $100.0_{\pm 0.0}$ & $98.2_{\pm 1.8}$ & $98.2_{\pm 1.3}$ & $97.8_{\pm 1.6}$ & $98.2_{\pm 1.3}$ & $98.5_{\pm 0.9}$ \\
Skill-curriculum & $100.0_{\pm 0.0}$ & $75.2_{\pm 30.1}$ & $52.2_{\pm 3.7}$ & $51.0_{\pm 4.6}$ & $54.4_{\pm 3.1}$ & $66.6_{\pm 7.7}$ \\
Skill-anticurriculum & $100.0_{\pm 0.0}$ & $90.2_{\pm 8.1}$ & $88.2_{\pm 8.3}$ & $73.2_{\pm 12.2}$ & $62.4_{\pm 9.4}$ & $82.8_{\pm 4.9}$ \\
\name & $100.0_{\pm 0.0}$ & $99.2_{\pm 0.8}$ & $99.0_{\pm 1.0}$ & $99.4_{\pm 0.5}$ & $99.6_{\pm 0.5}$ & $99.4_{\pm 0.2}$ \\
\bottomrule
\end{tabular}
\end{adjustbox}
\label{tab:lego_pt_acc}
\end{table}

\begin{table}[h] 
\footnotesize
\caption{Results on validation loss per skill for Addition pre-training experiment, averaged over $5$ random seeds.}
\begin{adjustbox}{center}
\begin{tabular}{c|cccc}
\toprule
& Skill 1 & Skill 2 & Skill 3 & Average \\
\midrule
Random & $0.008_{\pm 0.007}$ & $0.020_{\pm 0.019}$ & $0.005_{\pm 0.005}$ & $0.011_{\pm 0.014}$ \\
Curriculum & $0.009_{\pm 0.011}$ & $0.010_{\pm 0.008}$ & $0.008_{\pm 0.010}$ & $0.009_{\pm 0.010}$ \\
Anticurriculum & $0.007_{\pm 0.010}$ & $0.012_{\pm 0.013}$ & $0.008_{\pm 0.017}$ & $0.009_{\pm 0.014}$ \\
Skill-stratified & $0.012_{\pm 0.011}$ & $0.015_{\pm 0.015}$ & $0.010_{\pm 0.020}$ & $0.012_{\pm 0.016}$ \\
Skill-curriculum & $0.016_{\pm 0.013}$ & $0.019_{\pm 0.013}$ & $0.010_{\pm 0.003}$ & $0.015_{\pm 0.010}$ \\
Skill-anticurriculum & $0.005_{\pm 0.008}$ & $0.037_{\pm 0.028}$ & $1.141_{\pm 1.126}$ & $0.395_{\pm 0.371}$ \\
\name & $0.004_{\pm 0.003}$ & $0.009_{\pm 0.007}$ & $0.013_{\pm 0.017}$ & $0.009_{\pm 0.011}$ \\
\bottomrule
\end{tabular}
\end{adjustbox}
\label{tab:addition_pt_loss}
\end{table}

\begin{landscape}
\begin{table}
\footnotesize
\centering
\caption{Validation loss per skill for data selection in continual pre-training setting on a subset of the Natural Instructions Dataset.}
\centerline{\begin{tabular}{l|c|c|c|c|c|c|c}
\toprule
Skill & Random & Curriculum & Anticurriculum & Skill-stratified & Skill-curriculum & Skill-anticurriculum & \name \\
\midrule
Answer Verification & $2.297_{\pm 0.058}$ & $2.368_{\pm 0.055}$ & $2.391_{\pm 0.061}$ & $2.180_{\pm 0.059}$ & $2.249_{\pm 0.116}$ & $2.325_{\pm 0.085}$ & $2.158_{\pm 0.059}$ \\
Code to Text & $0.246_{\pm 0.021}$ & $0.203_{\pm 0.019}$ & $1.099_{\pm 0.115}$ & $0.178_{\pm 0.016}$ & $0.126_{\pm 0.009}$ & $1.232_{\pm 0.070}$ & $0.223_{\pm 0.017}$ \\
Discourse Connective Identification & $2.927_{\pm 0.069}$ & $3.084_{\pm 0.067}$ & $2.932_{\pm 0.058}$ & $2.805_{\pm 0.071}$ & $2.891_{\pm 0.001}$ & $2.925_{\pm 0.011}$ & $2.784_{\pm 0.068}$ \\
Entity Generation & $2.033_{\pm 0.421}$ & $2.012_{\pm 0.437}$ & $2.363_{\pm 0.234}$ & $1.803_{\pm 0.384}$ & $1.853_{\pm 0.483}$ & $2.068_{\pm 0.719}$ & $1.863_{\pm 0.418}$ \\
Entity Relation Classification & $1.020_{\pm 0.147}$ & $1.014_{\pm 0.140}$ & $1.533_{\pm 0.138}$ & $0.859_{\pm 0.131}$ & $0.825_{\pm 0.022}$ & $0.959_{\pm 0.009}$ & $0.908_{\pm 0.146}$ \\
Information Extraction & $2.154_{\pm 0.040}$ & $2.247_{\pm 0.037}$ & $2.352_{\pm 0.042}$ & $2.140_{\pm 0.037}$ & $2.286_{\pm 0.022}$ & $2.338_{\pm 0.025}$ & $2.073_{\pm 0.042}$ \\
Irony Detection & $3.024_{\pm 0.154}$ & $3.798_{\pm 0.095}$ & $2.942_{\pm 0.158}$ & $2.680_{\pm 0.146}$ & $3.889_{\pm 0.066}$ & $2.099_{\pm 0.152}$ & $2.797_{\pm 0.155}$ \\
Preposition Prediction & $0.979_{\pm 0.124}$ & $0.887_{\pm 0.147}$ & $1.488_{\pm 0.213}$ & $0.845_{\pm 0.152}$ & $0.941_{\pm 0.019}$ & $1.044_{\pm 0.029}$ & $0.876_{\pm 0.173}$ \\
Punctuation Error Detection & $2.950_{\pm 0.065}$ & $3.120_{\pm 0.052}$ & $2.961_{\pm 0.064}$ & $3.264_{\pm 0.061}$ & $3.019_{\pm 0.010}$ & $3.360_{\pm 0.013}$ & $3.216_{\pm 0.055}$ \\
Question Answering & $2.277_{\pm 0.005}$ & $2.367_{\pm 0.006}$ & $2.398_{\pm 0.006}$ & $2.542_{\pm 0.004}$ & $2.689_{\pm 0.001}$ & $2.707_{\pm 0.016}$ & $2.448_{\pm 0.008}$ \\
Question Generation & $2.617_{\pm 0.005}$ & $2.777_{\pm 0.015}$ & $2.695_{\pm 0.008}$ & $2.783_{\pm 0.021}$ & $3.062_{\pm 0.006}$ & $2.876_{\pm 0.032}$ & $2.666_{\pm 0.012}$ \\
Question Understanding & $1.965_{\pm 0.051}$ & $2.199_{\pm 0.059}$ & $2.060_{\pm 0.033}$ & $1.958_{\pm 0.051}$ & $2.385_{\pm 0.022}$ & $2.100_{\pm 0.054}$ & $1.895_{\pm 0.043}$ \\
Sentence Expansion & $2.501_{\pm 0.095}$ & $2.598_{\pm 0.097}$ & $2.583_{\pm 0.074}$ & $2.225_{\pm 0.095}$ & $2.311_{\pm 0.076}$ & $2.408_{\pm 0.074}$ & $2.236_{\pm 0.083}$ \\
Sentiment Analysis & $3.203_{\pm 0.012}$ & $3.415_{\pm 0.016}$ & $3.209_{\pm 0.010}$ & $3.278_{\pm 0.014}$ & $3.607_{\pm 0.012}$ & $3.308_{\pm 0.015}$ & $3.213_{\pm 0.012}$ \\
Stance Detection & $1.810_{\pm 0.100}$ & $1.775_{\pm 0.120}$ & $2.231_{\pm 0.128}$ & $1.385_{\pm 0.070}$ & $1.361_{\pm 0.114}$ & $1.823_{\pm 0.189}$ & $1.556_{\pm 0.125}$ \\
Summarization & $2.961_{\pm 0.015}$ & $3.149_{\pm 0.023}$ & $3.041_{\pm 0.014}$ & $2.960_{\pm 0.019}$ & $3.323_{\pm 0.028}$ & $3.021_{\pm 0.013}$ & $2.907_{\pm 0.012}$ \\
Text Categorization & $2.488_{\pm 0.023}$ & $2.692_{\pm 0.029}$ & $2.553_{\pm 0.006}$ & $2.570_{\pm 0.015}$ & $3.001_{\pm 0.007}$ & $2.635_{\pm 0.014}$ & $2.448_{\pm 0.017}$ \\
Text Matching & $2.177_{\pm 0.059}$ & $2.232_{\pm 0.055}$ & $2.316_{\pm 0.048}$ & $2.152_{\pm 0.061}$ & $2.324_{\pm 0.004}$ & $2.304_{\pm 0.035}$ & $2.093_{\pm 0.054}$ \\
Text Simplification & $2.155_{\pm 0.023}$ & $2.193_{\pm 0.039}$ & $2.325_{\pm 0.033}$ & $1.926_{\pm 0.026}$ & $2.037_{\pm 0.005}$ & $2.156_{\pm 0.011}$ & $1.952_{\pm 0.026}$ \\
Text to Code & $0.560_{\pm 0.037}$ & $0.495_{\pm 0.036}$ & $1.215_{\pm 0.052}$ & $0.490_{\pm 0.029}$ & $0.433_{\pm 0.014}$ & $1.455_{\pm 0.086}$ & $0.553_{\pm 0.042}$ \\
Toxic Language Detection & $3.106_{\pm 0.027}$ & $3.496_{\pm 0.017}$ & $3.058_{\pm 0.029}$ & $3.199_{\pm 0.024}$ & $3.758_{\pm 0.025}$ & $3.155_{\pm 0.050}$ & $3.129_{\pm 0.020}$ \\
Word Semantics & $2.092_{\pm 0.027}$ & $2.334_{\pm 0.034}$ & $2.156_{\pm 0.064}$ & $1.916_{\pm 0.043}$ & $1.784_{\pm 0.048}$ & $2.424_{\pm 0.038}$ & $1.952_{\pm 0.019}$ \\
Wrong Candidate Generation & $2.438_{\pm 0.021}$ & $2.606_{\pm 0.039}$ & $2.519_{\pm 0.027}$ & $2.506_{\pm 0.026}$ & $2.849_{\pm 0.029}$ & $2.574_{\pm 0.018}$ & $2.432_{\pm 0.025}$ \\
\midrule 
Average & $2.173_{\pm 0.028}$ & $2.307_{\pm 0.025}$ & $2.366_{\pm 0.026}$ & $2.115_{\pm 0.027}$ & $2.304_{\pm 0.031}$ & $2.317_{\pm 0.052}$ & $2.103_{\pm 0.032}$ \\
\bottomrule
\end{tabular}}
\label{tab:ni_pretraining}
\end{table}
\end{landscape}

\paragraph{Out-of-domain}

In Table~\ref{tab:ni_test}, we provide a breakdown of validation loss per evaluation skill under random sampling on the training data, skill-stratified sampling over prerequisite skills (e.g., the nonzero rows in Figure~\ref{fig:ni_test_heatmap}), and \name.

\begin{table}[htbp]
  \centering
  \caption{Validation loss per skill for data selection in out-of-domain setting over Natural Instructions train task split and test task split.}
  \label{tab:skill-results}
  \begin{tabular}{l |c |c |c}
    \toprule
    Skill & Random & Skill-stratified & \name \\
    \midrule
    Answerability Classification & $3.048_{\pm 0.003}$ & $3.076_{\pm 0.002}$ & $3.043_{\pm 0.003}$ \\
    Cause Effect Classification & $2.068_{\pm 0.004}$ & $2.101_{\pm 0.005}$ & $2.067_{\pm 0.006}$ \\
    Coreference Resolution & $3.101_{\pm 0.003}$ & $3.142_{\pm 0.004}$ & $3.099_{\pm 0.004}$ \\
    Data to Text & $2.363_{\pm 0.004}$ & $2.388_{\pm 0.005}$ & $2.359_{\pm 0.005}$ \\
    Dialogue Act Recognition & $2.329_{\pm 0.009}$ & $2.364_{\pm 0.010}$ & $2.320_{\pm 0.009}$ \\
    Grammar Error Correction & $2.399_{\pm 0.008}$ & $2.418_{\pm 0.009}$ & $2.389_{\pm 0.007}$ \\
    Keyword Tagging & $2.744_{\pm 0.005}$ & $2.760_{\pm 0.007}$ & $2.733_{\pm 0.006}$ \\
    Overlap Extraction & $2.749_{\pm 0.011}$ & $2.763_{\pm 0.012}$ & $2.733_{\pm 0.010}$ \\
    Question Rewriting & $2.591_{\pm 0.009}$ & $2.628_{\pm 0.011}$ & $2.586_{\pm 0.010}$ \\
    Textual Entailment & $2.472_{\pm 0.002}$ & $2.503_{\pm 0.003}$ & $2.468_{\pm 0.002}$ \\
    Title Generation & $3.027_{\pm 0.002}$ & $3.037_{\pm 0.002}$ & $3.015_{\pm 0.002}$ \\
    Word Analogy & $1.665_{\pm 0.016}$ & $1.682_{\pm 0.015}$ & $1.668_{\pm 0.016}$ \\
    \midrule 
    Average & $2.546_{\pm 0.003}$ & $2.572_{\pm 0.003}$ & $2.540_{\pm 0.003}$ \\
    \bottomrule
  \end{tabular}
  \label{tab:ni_test}
\end{table}

 In Table~\ref{tab:redpajama} we provide a breakdown of the RedPajama experiment's accuracy per evaluation skill, corresponding to the results in Figure~\ref{fig:redpajama}.

\begin{table}[ht]
\centering
\caption{Performance of model trained on RedPajama with uniform sampling and \name on LM evaluation harness. Unless otherwise noted, accuracy is reported for each task.}
\begin{tabular}{l|cc|cc|cc}
\toprule
& \multicolumn{2}{c|}{1 Billion Tokens} & \multicolumn{2}{c|}{2 Billion Tokens} & \multicolumn{2}{c}{3 Billion Tokens} \\
\cmidrule{2-3} \cmidrule{4-5} \cmidrule{6-7}
& Uniform & \name & Uniform & \name & Uniform & \name  \\
\midrule
ARC Challenge (acc norm) & $35.4$ & $34.6$ & $35.3$ & $34.9$ & $34.6$ & $34.8$ \\
ARC Easy (acc norm) & $62.2$ & $61.2$ & $62.4$ & $61.7$ & $62.5$ & $62.0$ \\
BoolQ & $68.9$ & $68.2$ & $67.7$ & $68.6$ & $67.2$ & $68.7$ \\
COPA & $81.0$ & $82.0$ & $80.0$ & $81.0$ & $81.0$ & $81.0$ \\
HellaSwag (acc norm) & $63.9$ & $63.7$ & $63.8$ & $63.9$ & $64.0$ & $63.9$ \\
LAMBADA OpenAI & $64.4$ & $67.0$ & $65.9$ & $66.7$ & $66.8$ & $66.0$ \\
PIQA (acc norm) & $74.8$ & $75.0$ & $75.5$ & $75.2$ & $75.0$ & $75.7$ \\
Winogrande & $62.8$ & $63.9$ & $63.9$ & $63.2$ & $63.4$ & $63.1$ \\
\midrule 
Average accuracy & $64.2$ & $64.4$ & $64.3$ & $64.4$ & $64.3$ & $64.4$ \\
\bottomrule
\end{tabular}
\label{tab:redpajama}
\end{table}

\subsubsection{Weight trajectories} \label{supp:weights}

\begin{figure}
    \centering
    \includegraphics[width=0.6\textwidth]{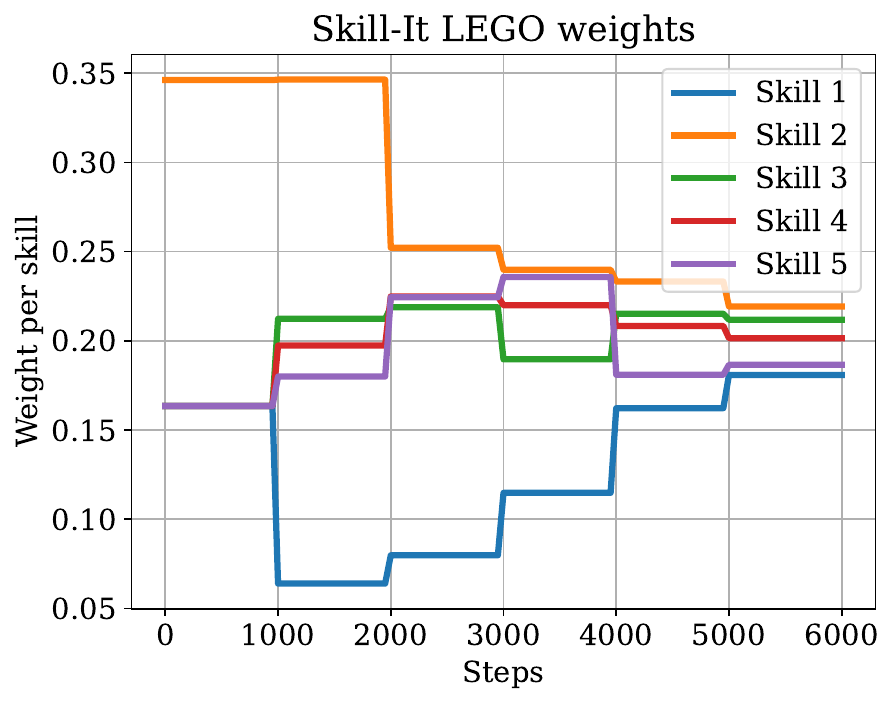}
    \caption{Weight per skill for LEGO pre-training experiment. \name initially allocates more weight to skill $2$, but eventually puts more weight on harder skills ($3, 4, 5$) before converging to uniform sampling when all losses converge roughly to $0$.}
    \label{fig:lego_weights}
\end{figure}

We provide \name's weight trajectories for each result. The weight per skill across training steps for the LEGO pre-training experiment corresponding to Figure~\ref{fig:lego_mw} (left) is shown in Figure~\ref{fig:lego_weights}. We see that \name initially allocates more weight to skill $2$ and less to $1, 3, 4, 5$. Since skill $1$ is learned quickly, the weight on skill $1$ immediately drops to below $0.1$ at $1000$ steps. The weight on skills $3, 4,$ and $5$ increase from around $0$ to $3000$ steps, during which their respective validation losses are higher than those of skills $1$ and $2$. Near the end of training, all losses are converging to $0$, and so the weight per skill is roughly uniform.

\begin{figure}
    \centering
    \includegraphics[width=0.6\textwidth]{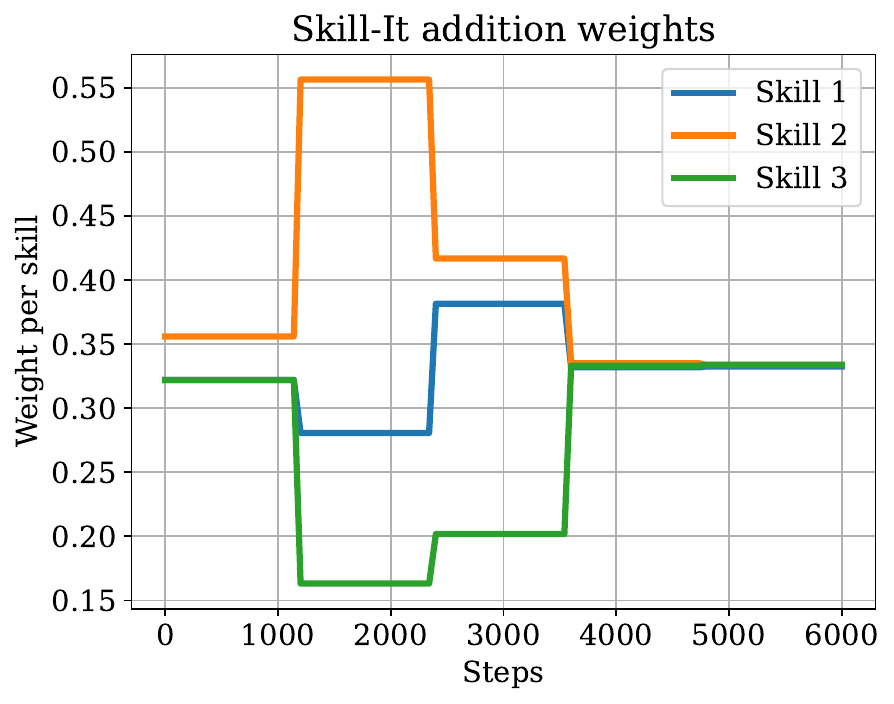}
    \caption{Weight per skill for addition pre-training experiment. \name initially allocates more weight to skill $2$, which has an edge to skill $1$, while allocating little weight to skill $3$ which is learned quickly. Eventually, \name puts more weight on the harder skill $1$ before converging to uniform sampling when all losses roughly approach $0$.}
    \label{fig:addn_weights}
\end{figure}

The weight per skill across training steps for the addition pre-training experiment corresponding to Figure~\ref{fig:lego_mw} (right) is shown in Figure~\ref{fig:addn_weights}. \name allocates more weight to skill $2$, which has an edge to skill $1$ as shown in Figure~\ref{fig:addition_heatmap}. It also allocates very little weight to skill $3$, which is learned faster than the other two skills. Eventually, it puts more weight on skill $1$, the hardest skill, and then converges to uniform sampling as all validation losses approach $0$.

\begin{figure}
    \centering
    \includegraphics[width=0.3\textwidth]{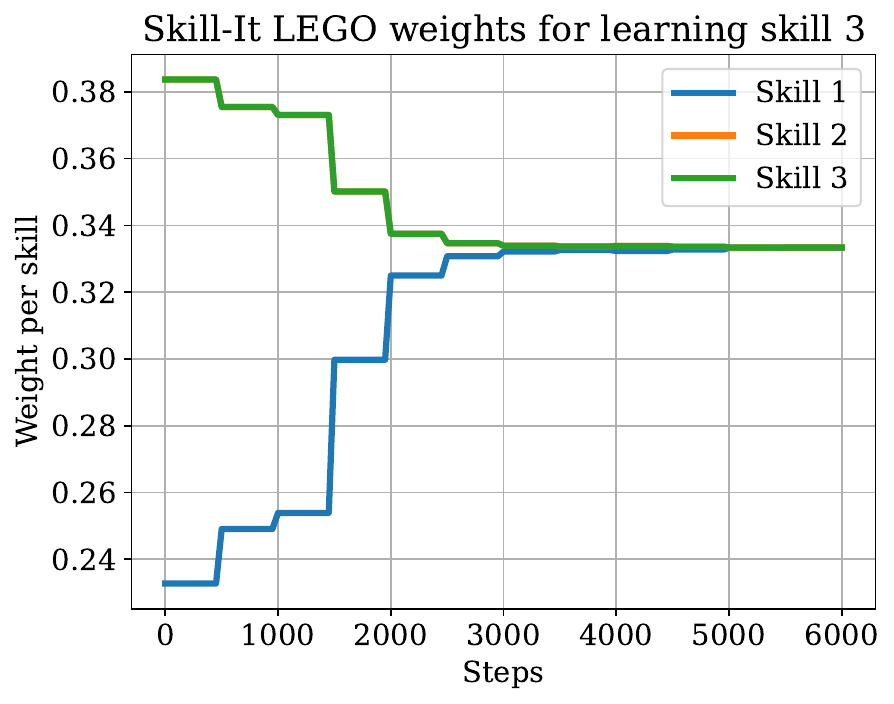}
    \includegraphics[width=0.3\textwidth]{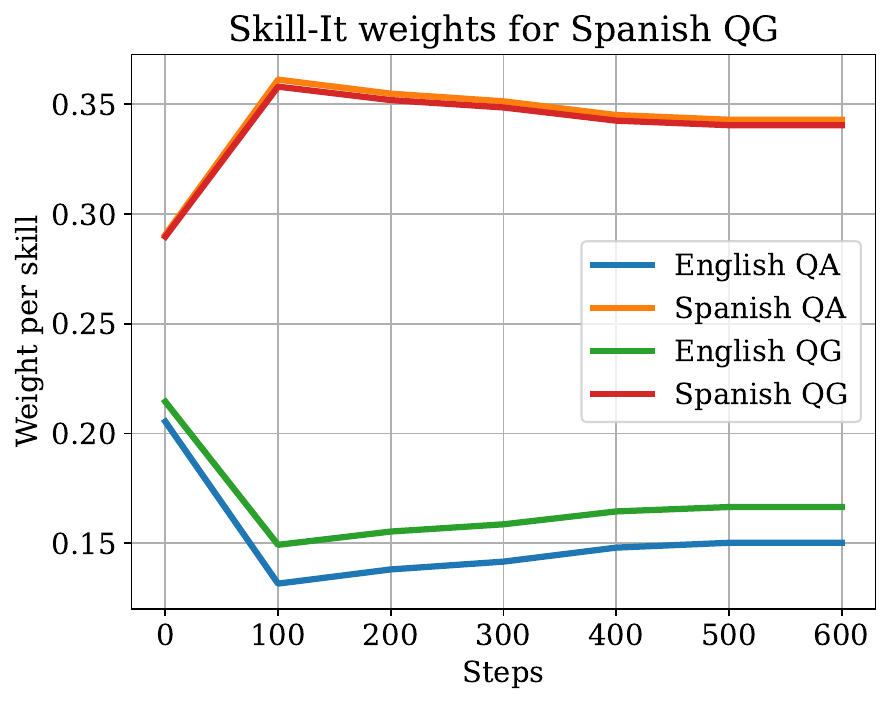}
    \includegraphics[width=0.3\textwidth]{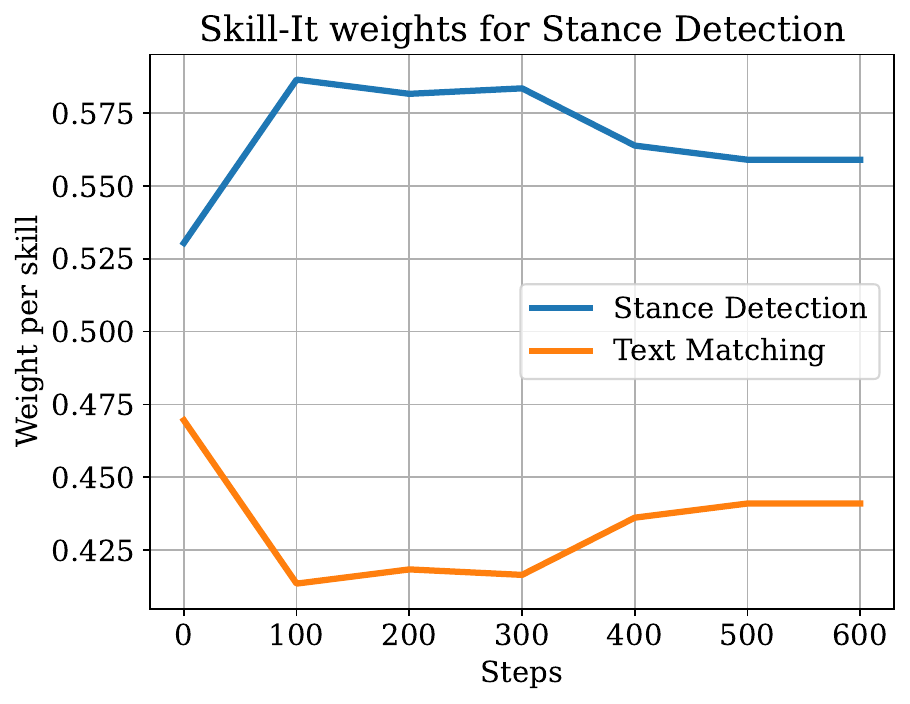}
    \caption{Weight per skill for fine-tuning experiments. Left: LEGO; Center: Spanish question generation; Right: stance detection.}
    \label{fig:ft_weights}
\end{figure}

The weight per skill across training steps for the LEGO fine-tuning experiment and the Spanish question generation and stance detection experiments corresponding to Figure~\ref{fig:ni_targeted} is shown in Figure~\ref{fig:ft_weights}. Since there is only one target skill in these experiments, the mixture of weights approaches uniform as the loss on the target skill approaches $0$. It is interesting to explore how to reduce edge weights and regularization so that the mixture approaches the target skill instead, although preliminary experiments where we decayed the edge weight and the strength of the Bregman divergence term did not appear better. We hypothesize that since training on a uniform mixture (as in Figure~\ref{fig:ni_lego_examples}) did strictly better than training on the target skill and their loss curves did not intersect during the training run, it is better to allocate non-negligible weight on all skills throughout the training run. 

\begin{figure}
    \centering
    \includegraphics[width=0.6\textwidth]{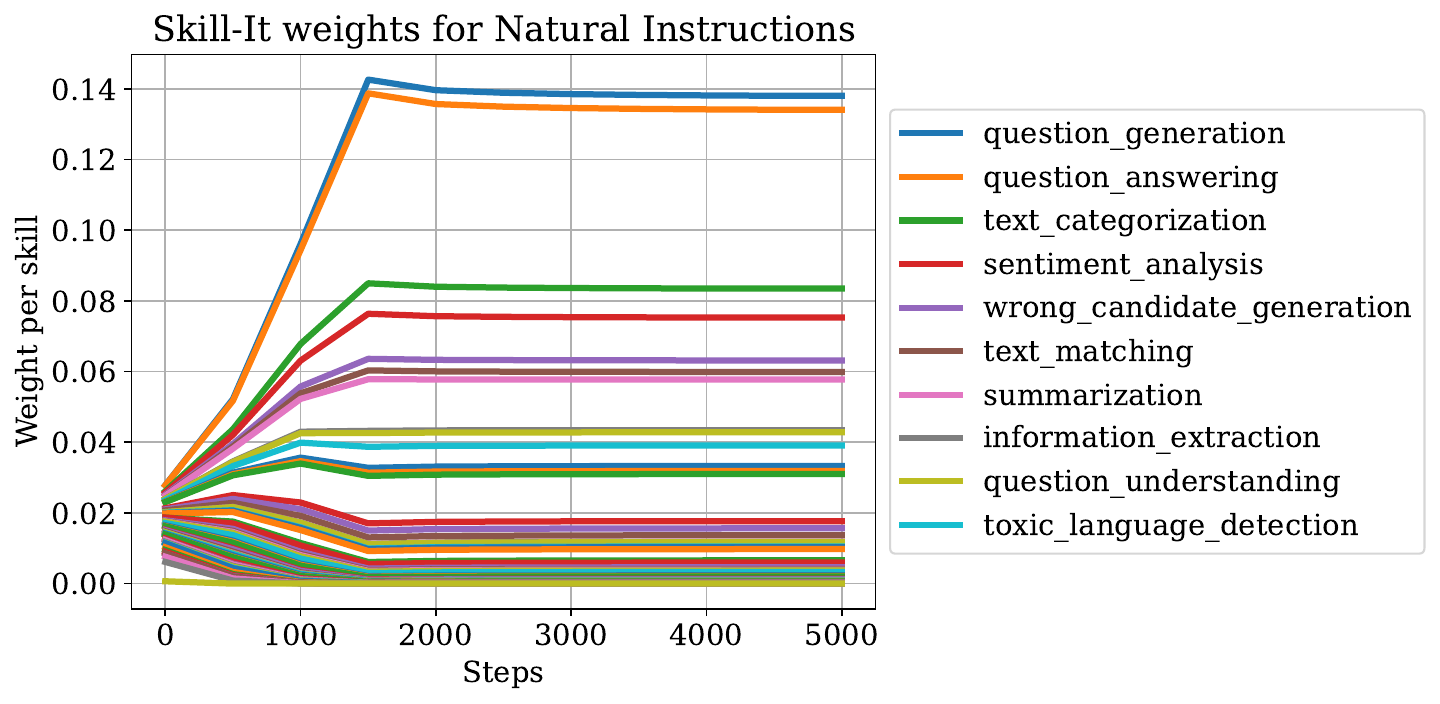}
    \caption{Weight per skill for Natural Instructions out-of-domain experiment. The legend shows the top $10$ skills with the largest weight. While the relative order of weight magnitude does not change significantly across training, the incorporation of loss dramatically increases the range of the weights, showing the importance of an online algorithm.}
    \label{fig:ni_test_weights}
\end{figure}

The weight per skill across training steps for the Natural Instructions out-of-domain experiment corresponding to Figure~\ref{fig:ni_test} is shown in Figure~\ref{fig:ni_test_weights}, where the legend is provided for the top $10$ task categories with the largest weights. While the initial weights based on the skills graph roughly establishes the order of weight magnitude, the differences among the losses on the evaluation skills increases the range of weights as training continues. As validation losses saturate, the weights also converge to fixed values.

\subsection{Experiments on 1.3B parameter model}\label{supp:larger_model}

We demonstrate that the skills graph learned on the 125M parameter model can be used for data selection with the GPT-Neo-1.3B model. We present results in the continual pre-training setting on the LEGO synthetic and Natural Instructions.

All results are reported over $3$ random seeds. For the LEGO experiment, we train for $1500$ steps with $\eta =0.5, T = 30, w = 3$. For the NI experiment, we train for $5000$ steps with $\eta = 0.2$, and $T = 1$. The skill graphs were learned using the 125M parameter model as described in section~\ref{supp:skill_graphs}.

\begin{figure}
    \centering
    \includegraphics[width=6in]{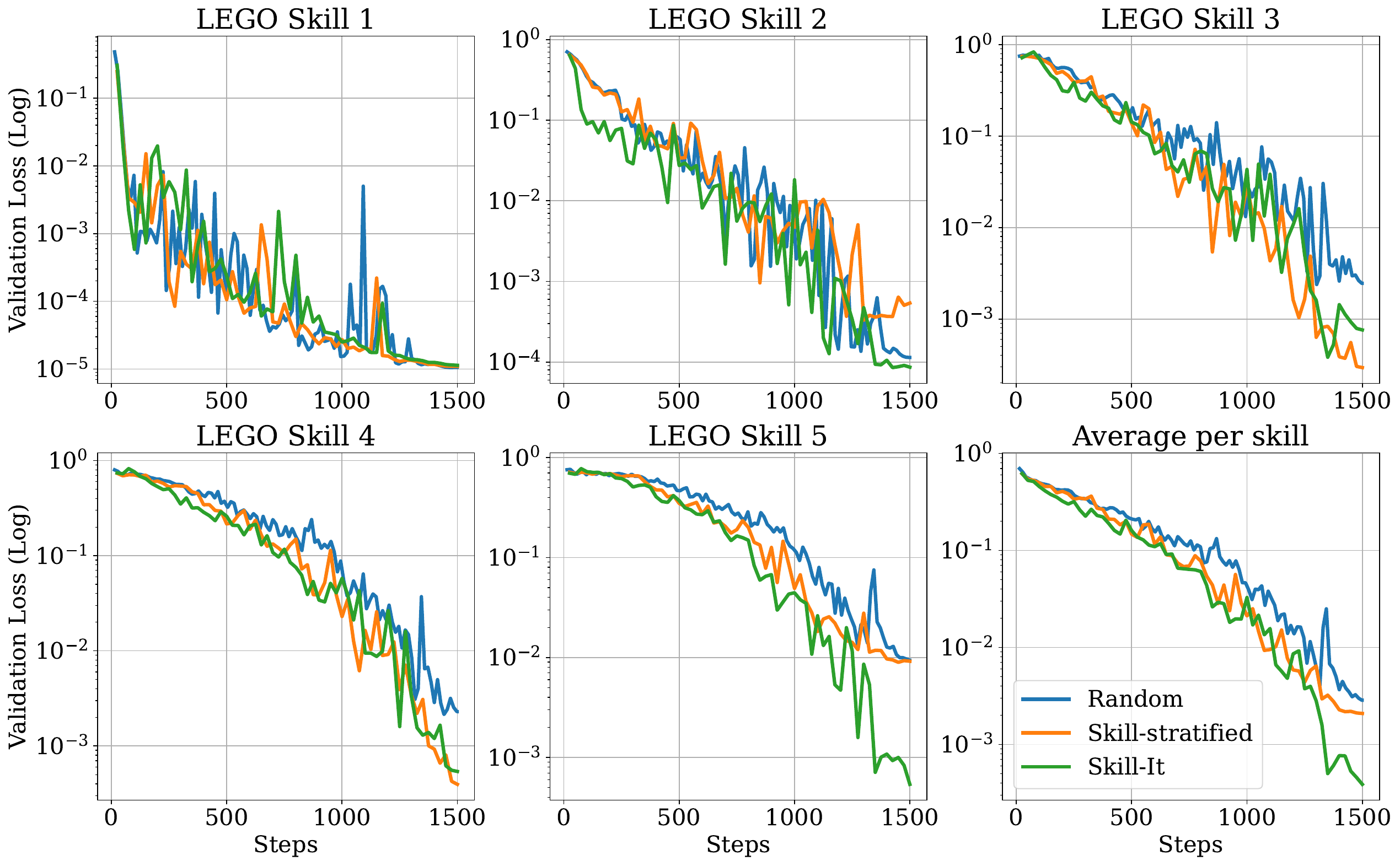}
    \caption{Performance of \name for LEGO pre-training setting when skills graph is learned on a 125M parameter model and used for data selection with a 1.3B model. \name on average still outperforms random and skill-stratified sampling, suggesting that findings on ordered skill sets can transfer from small models to large models.}
    \label{fig:lego_large}
\end{figure}

\begin{figure}
    \centering
    \includegraphics[width=0.5\textwidth]{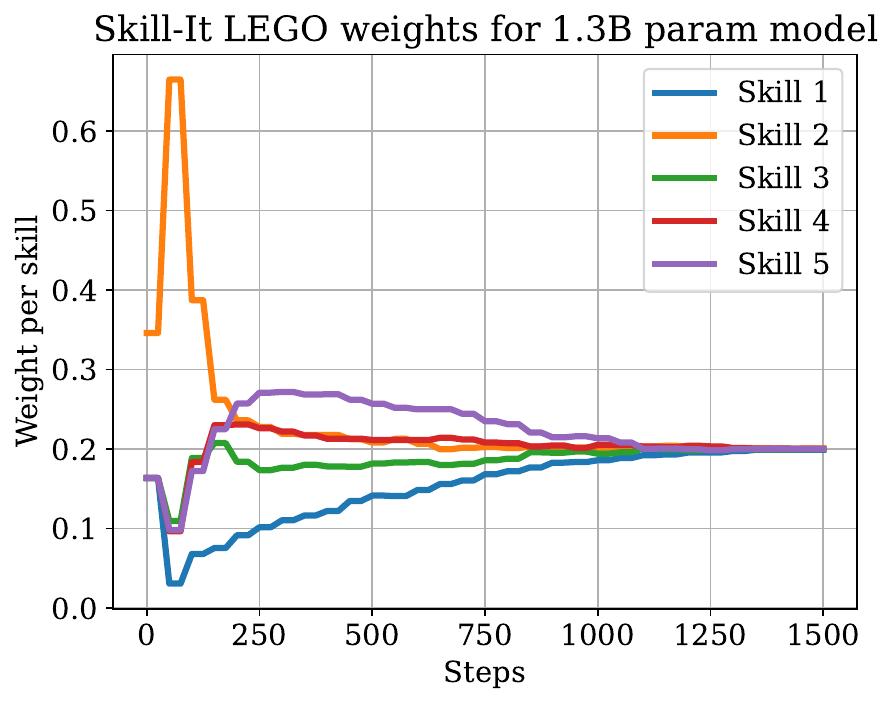}
    \caption{Weight per skill for LEGO pre-training experiment on 1.3B parameter model. The trajectories are similar to those of the 125M parameter model in Figure~\ref{fig:lego_weights}. \name initially allocates more weight to skill $2$, but eventually puts more weight on skills $4$ and $5$ before converging to uniform sampling when all losses converge to near $0$.}
    \label{fig:lego_weights_large}
\end{figure}

In Figure~\ref{fig:lego_large}, we train the $1.3$B model using \name for the LEGO synthetic and find that it still outperforms random and skill-stratified sampling on average. In particular, while performance across sampling methods is similar for early skills, the discrepancy is larger for skill $5$, for which \name allocates more weight to dynamically. In Figure~\ref{fig:lego_weights_large}, we provide the weight trajectories of \name. We observe that the weight trajectories are similar to that on the 125M parameter model, where initial weight is allocated towards skill $2$. Later on, more weight is allocated towards skills $4$ and $5$, whose losses are higher, and eventually the weight mixture converges to uniform as all losses converge to near $0$. 

In Table~\ref{tab:ni_pretraining_large}, we report performance of \name with the $1.3$B model on the Natural Instructions pre-training experiment and find that the trends from the smaller model hold---\name outperforms random and skill-stratified sampling on average.

\begin{table}[ht]
\footnotesize
\centering
\caption{Results when skills graph for Natural Instructions learned on 125M parameter model is used for data selection with a 1.3B model. We see that \name on average still outperforms random and skill-stratified sampling, even though the edges used by \name are not derived from the larger model.}
\begin{tabular}{l|c|c|c}
\toprule
Skill & Random & Skill-stratified & \name \\
\midrule
Answer Verification & $2.005_{\pm 0.059}$ & $1.903_{\pm 0.069}$ & $1.890_{\pm 0.072}$ \\
Code to Text & $0.302_{\pm 0.032}$ & $0.204_{\pm 0.022}$ & $0.269_{\pm 0.032}$ \\
Discourse Connective Identification & $2.529_{\pm 0.046}$ & $2.372_{\pm 0.054}$ & $2.393_{\pm 0.056}$ \\
Entity Generation & $2.108_{\pm 0.328}$ & $1.788_{\pm 0.429}$ & $1.885_{\pm 0.461}$ \\
Entity Relation Classification & $1.130_{\pm 0.048}$ & $0.836_{\pm 0.006}$ & $0.841_{\pm 0.010}$ \\
Information Extraction & $2.032_{\pm 0.013}$ & $1.992_{\pm 0.006}$ & $1.933_{\pm 0.013}$ \\
Irony Detection & $2.802_{\pm 0.125}$ & $2.528_{\pm 0.146}$ & $2.585_{\pm 0.149}$ \\
Preposition Prediction & $1.095_{\pm 0.040}$ & $0.686_{\pm 0.041}$ & $0.774_{\pm 0.029}$ \\
Punctuation Error Detection & $2.633_{\pm 0.027}$ & $3.188_{\pm 0.055}$ & $2.726_{\pm 0.025}$ \\
Question Answering & $1.947_{\pm 0.003}$ & $2.119_{\pm 0.003}$ & $2.073_{\pm 0.001}$ \\
Question Generation & $2.214_{\pm 0.007}$ & $2.345_{\pm 0.008}$ & $2.263_{\pm 0.010}$ \\
Question Understanding & $1.928_{\pm 0.020}$ & $1.837_{\pm 0.031}$ & $1.700_{\pm 0.042}$ \\
Sentence Expansion & $2.054_{\pm 0.018}$ & $1.828_{\pm 0.060}$ & $1.853_{\pm 0.058}$ \\
Sentiment Analysis & $2.771_{\pm 0.009}$ & $2.818_{\pm 0.006}$ & $2.774_{\pm 0.007}$ \\
Stance Detection & $1.814_{\pm 0.151}$ & $1.500_{\pm 0.117}$ & $1.628_{\pm 0.149}$ \\
Summarization & $2.531_{\pm 0.009}$ & $2.472_{\pm 0.012}$ & $2.440_{\pm 0.013}$ \\
Text Categorization & $2.289_{\pm 0.016}$ & $2.341_{\pm 0.021}$ & $2.231_{\pm 0.022}$ \\
Text Matching & $1.967_{\pm 0.008}$ & $1.913_{\pm 0.005}$ & $1.872_{\pm 0.005}$ \\
Text Simplification & $1.861_{\pm 0.003}$ & $1.692_{\pm 0.023}$ & $1.698_{\pm 0.022}$ \\
Text to Code & $0.614_{\pm 0.030}$ & $0.518_{\pm 0.030}$ & $0.585_{\pm 0.022}$ \\
Toxic Language Detection & $2.853_{\pm 0.020}$ & $2.911_{\pm 0.019}$ & $2.862_{\pm 0.018}$ \\
Word Semantics & $1.999_{\pm 0.023}$ & $1.870_{\pm 0.039}$ & $1.902_{\pm 0.024}$ \\
Wrong Candidate Generation & $2.187_{\pm 0.028}$ & $2.192_{\pm 0.023}$ & $2.140_{\pm 0.020}$ \\
\midrule 
Average & $1.985_{\pm 0.022}$ & $1.907_{\pm 0.027}$ & $1.883_{\pm 0.032}$ \\
\bottomrule
\end{tabular}
\label{tab:ni_pretraining_large}
\end{table}

\subsection{Ablations} \label{supp:ablations}

We report ablations on the skills graph and the online component of \name. Instead of using $A$ in Algorithm~\ref{alg:skillit}, we study the performance when the identity matrix is used instead; intuitively, this corresponds to a misspecified skills graph where no skill influences another skill. We refer this approach as ``No graph''. Note that the opposite case of a complete graph recovers skill-stratified sampling, which we already have as a baseline.

Second, instead of sampling over multiple rounds and weighting according to the loss of each skill, we study the effect of setting $T = 1$, which only uses a softmax on $A$ to yield static weights on the skills. We refer to this approach as ``Static''. We omit results on Natural Instructions continual pre-training, since \name uses $T = 1$ and using no graph with $T = 1$ recovers skill-stratified sampling. Intuitively, we expect the static version of \name to perform somewhat well unless there is significant discrepancy among the losses (e.g. in synthetics where the loss on one skill can be close to $0$ while the other is not, versus in Natural Instructions where all losses decrease consistently). For both ablations, we sweep over values of $\eta = [0.1, 0.2, 0.5, 0.8]$.

\begin{figure}
    \centering
    \includegraphics[width=5in]{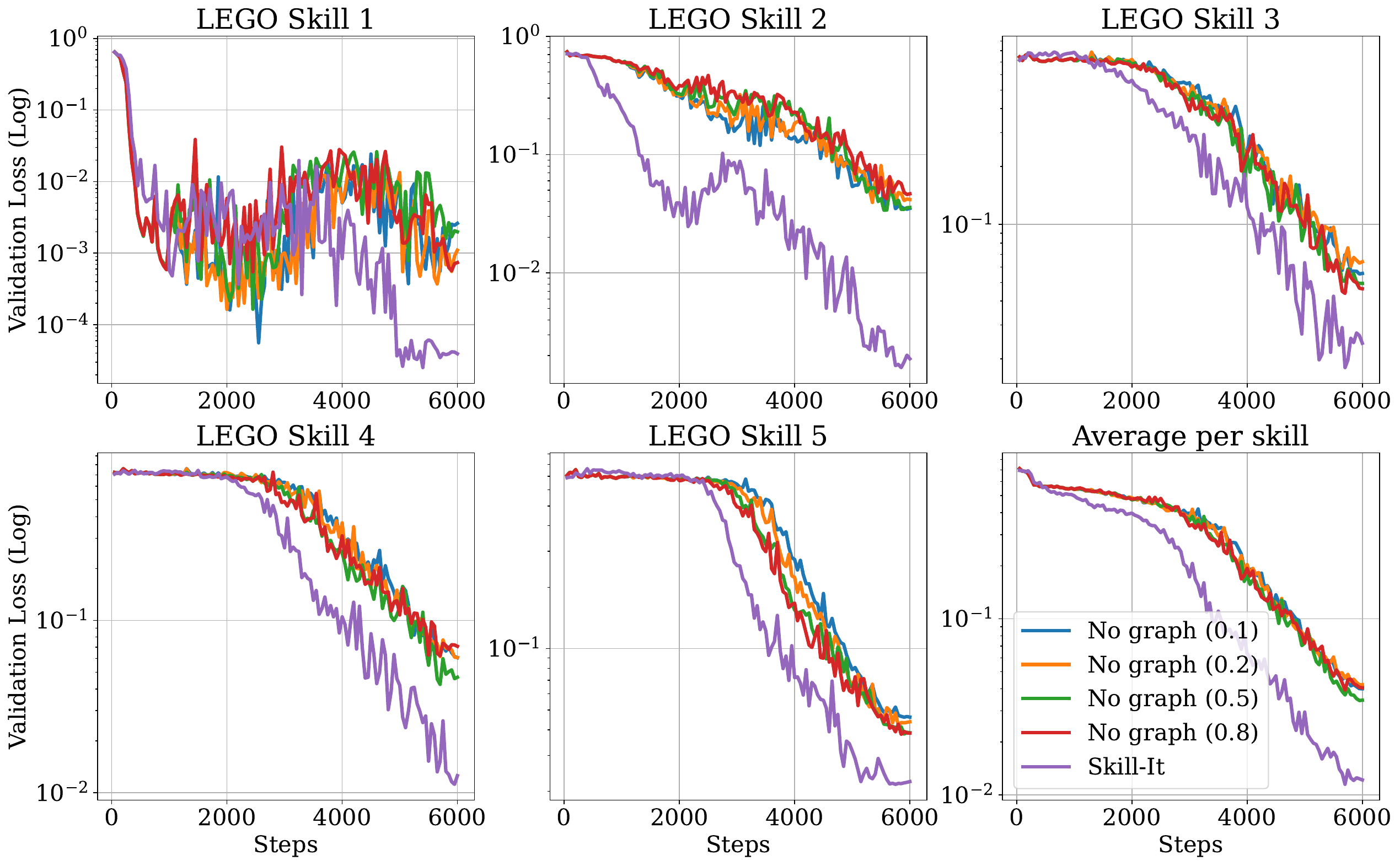}
    \caption{Comparison of \name versus using the identity adjacency matrix (no skills graph) with $\eta = 0.1, 0.2, 0.5, 0.8$ on the LEGO continual pre-training experiment. The latter does not capture the relationship between skills, and we find that \name attains lower loss on all skills.}
    \label{fig:lego_ablation_no_graph}
\end{figure}

\begin{figure}
    \centering
    \includegraphics[width=5in]{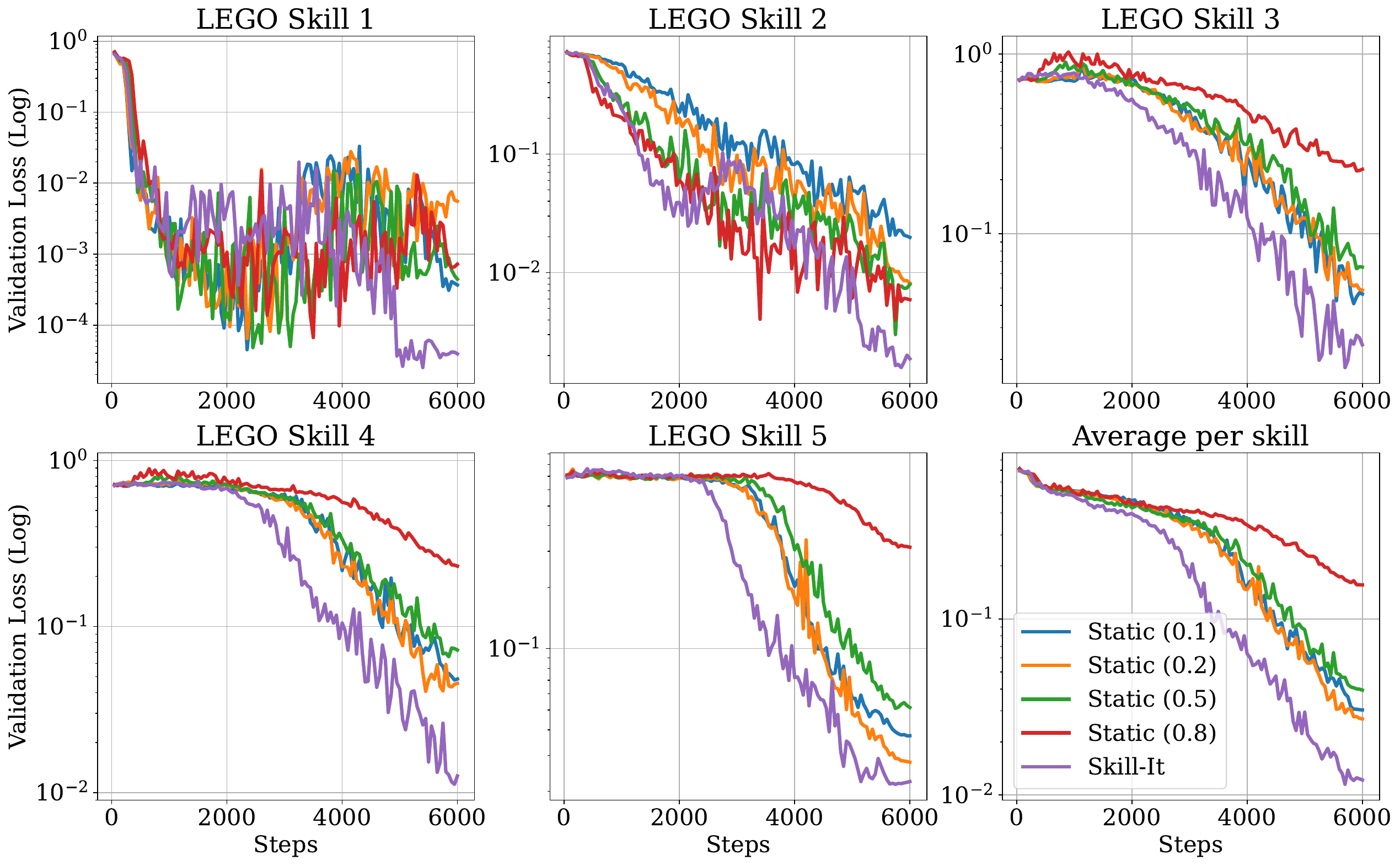}
    \caption{Comparison of \name versus using static data selection ($T = 1$) with $\eta = 0.1, 0.2, 0.5, 0.8$ on the LEGO continual pre-training experiment. While \name eventually allocates more weights to skills 3, 4, 5, which have higher loss, the static approach is not able to do this. We find that \name attains lower loss on all skills.}
    \label{fig:lego_ablation_static}
\end{figure}

Figure~\ref{fig:lego_ablation_no_graph} shows the comparison between \name and no graph on the continual pre-training LEGO experiment, and Figure~\ref{fig:lego_ablation_static} shows the comparison between \name and a static approach. We see that both the graph and the online dynamics of \name are important for its performance. In particular, using no graph results in allocating significant weight to harder skills early on, even though many of them have easier prerequisite skills (such as skill 3 having edges to skills 1 and 2). Using a static graph results in consistent allocation of significant weight to prerequisite skills even after their validation losses converge to near $0$, and thus the harder skills that have higher loss are not learned quickly afterwards.

\begin{figure}
    \centering
    \includegraphics[width=3in]{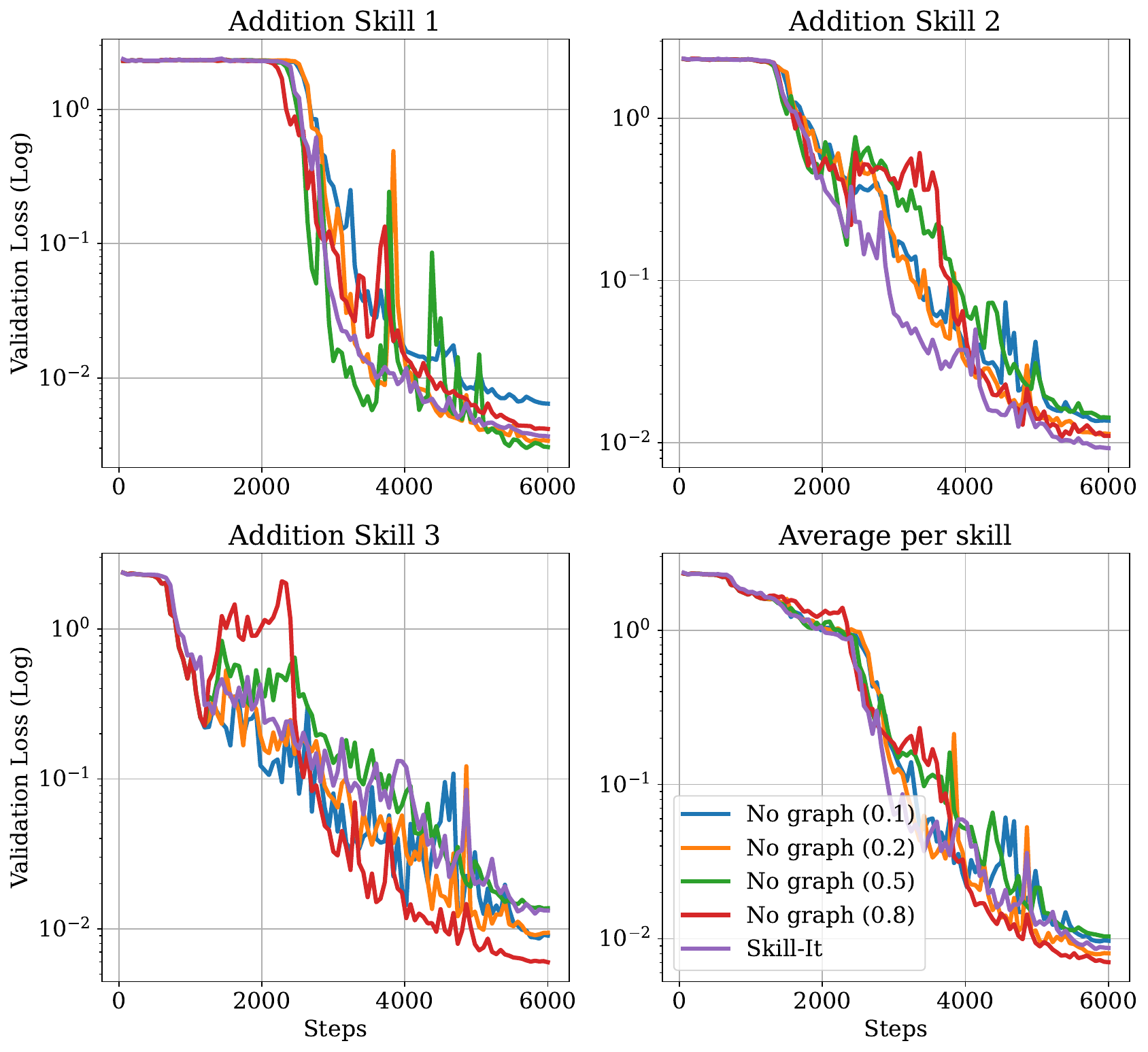}
    \caption{Comparison of \name versus using the identity adjacency matrix (no skills graph) with $\eta = 0.1, 0.2, 0.5, 0.8$ on the Addition continual pre-training experiment. 
    The latter does not capture the relationship between skills, and we find that \name attains lower loss on skill 2, but attains similar performance to methods that do not use the skills graph.}
    \label{fig:addn_ablation_no_graph}
\end{figure}

\begin{figure}
    \centering
    \includegraphics[width=3in]{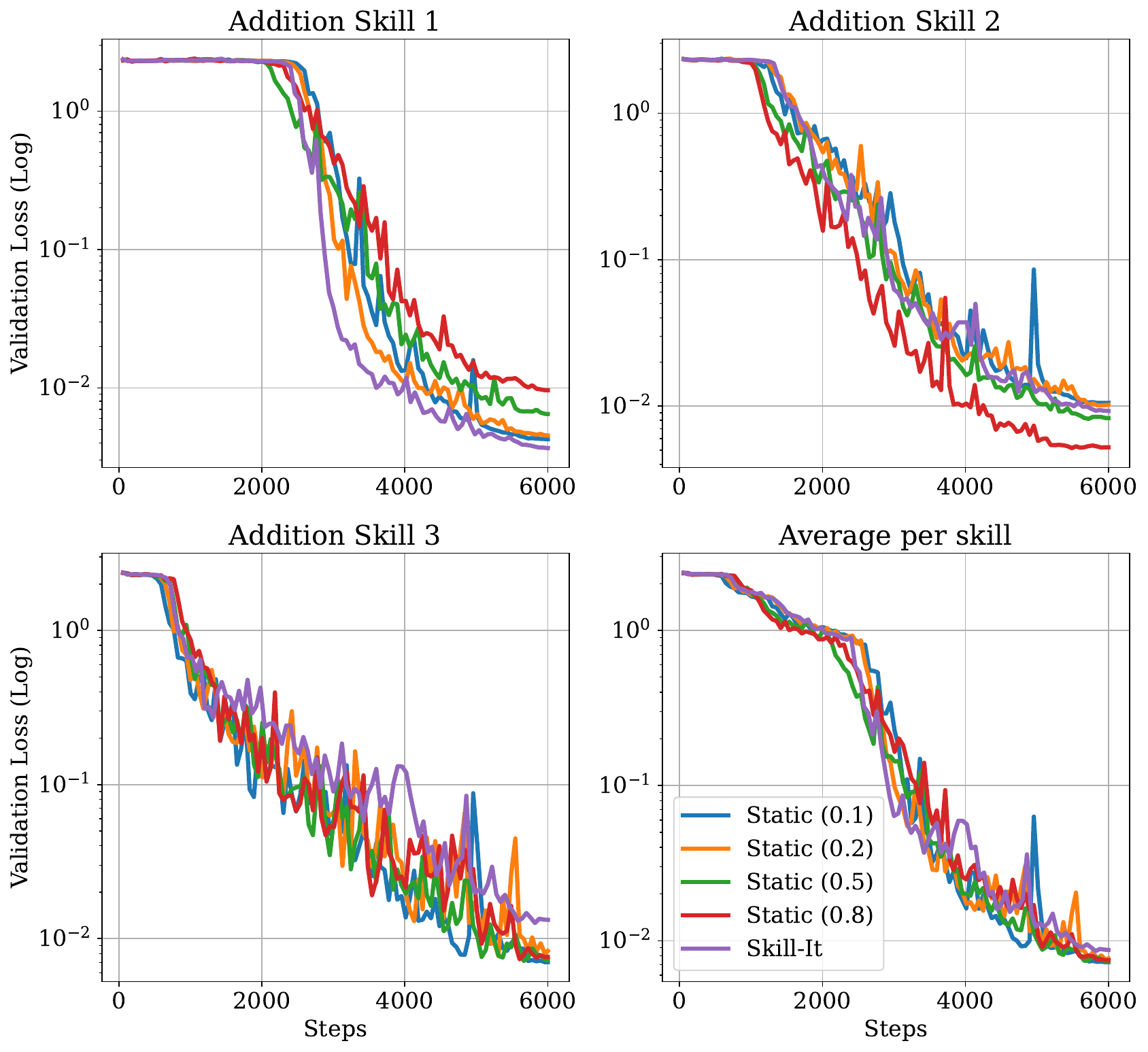}
    \caption{Comparison of \name versus using static data selection ($T = 1$) with $\eta = 0.1, 0.2, 0.5, 0.8$ on the Addition continual pre-training experiment. We find that \name attains lower loss on skill 1, but attains similar performance to the static methods.}
    \label{fig:addn_ablation_static}
\end{figure}

We perform the same ablation on the Addition dataset---the results for this are shown in Figures~\ref{fig:addn_ablation_no_graph} and~Figure~\ref{fig:addn_ablation_static}.
We find that these simple baselines, including using a static graph and no graph perform similarly to \name on average across all skills---while \name performs the best on skill 2 compared to vanilla multiplicative weights, and \name performs the best on skill 1 compared to a static graph. 
This suggests that Addition is somewhat easier than the other datasets that we consider, as \name still outperforms other baselines, as shown in Figure~\ref{fig:lego_mw}. 

\begin{figure}
    \centering
    \includegraphics[height=2in]{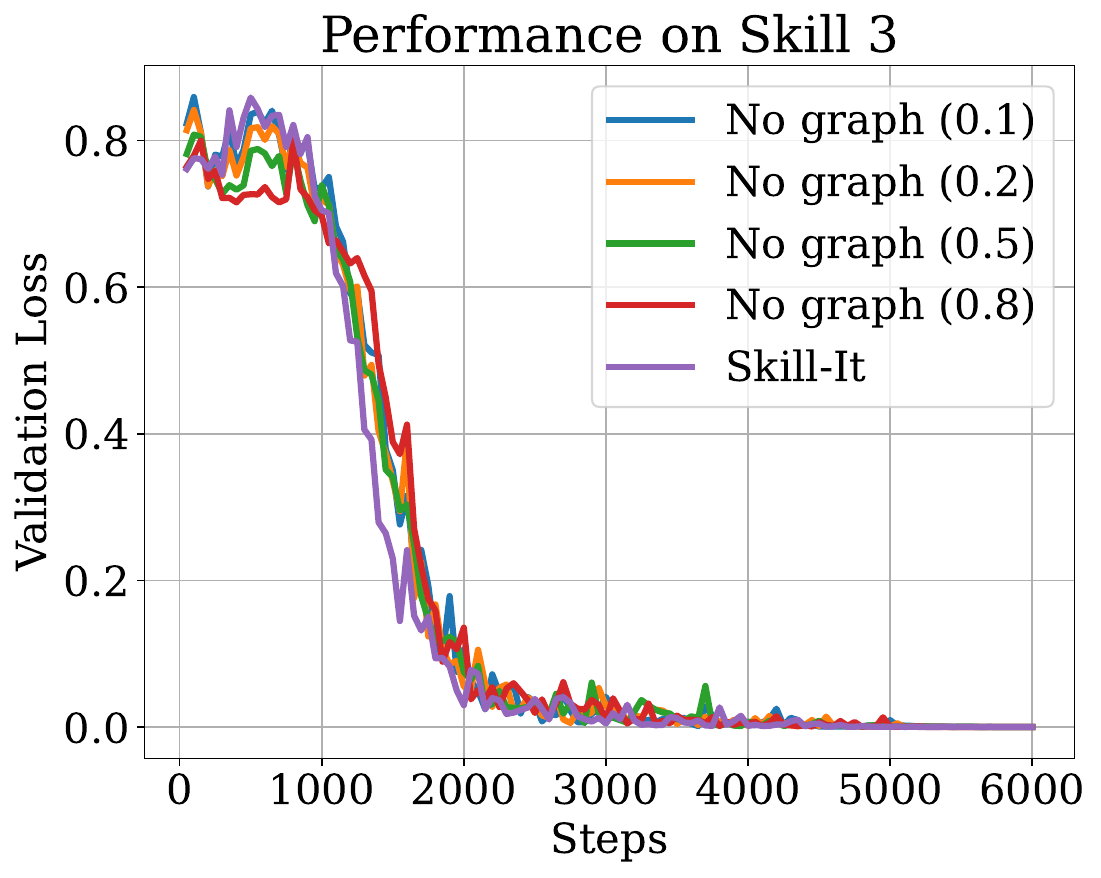}
    \includegraphics[height=2in]{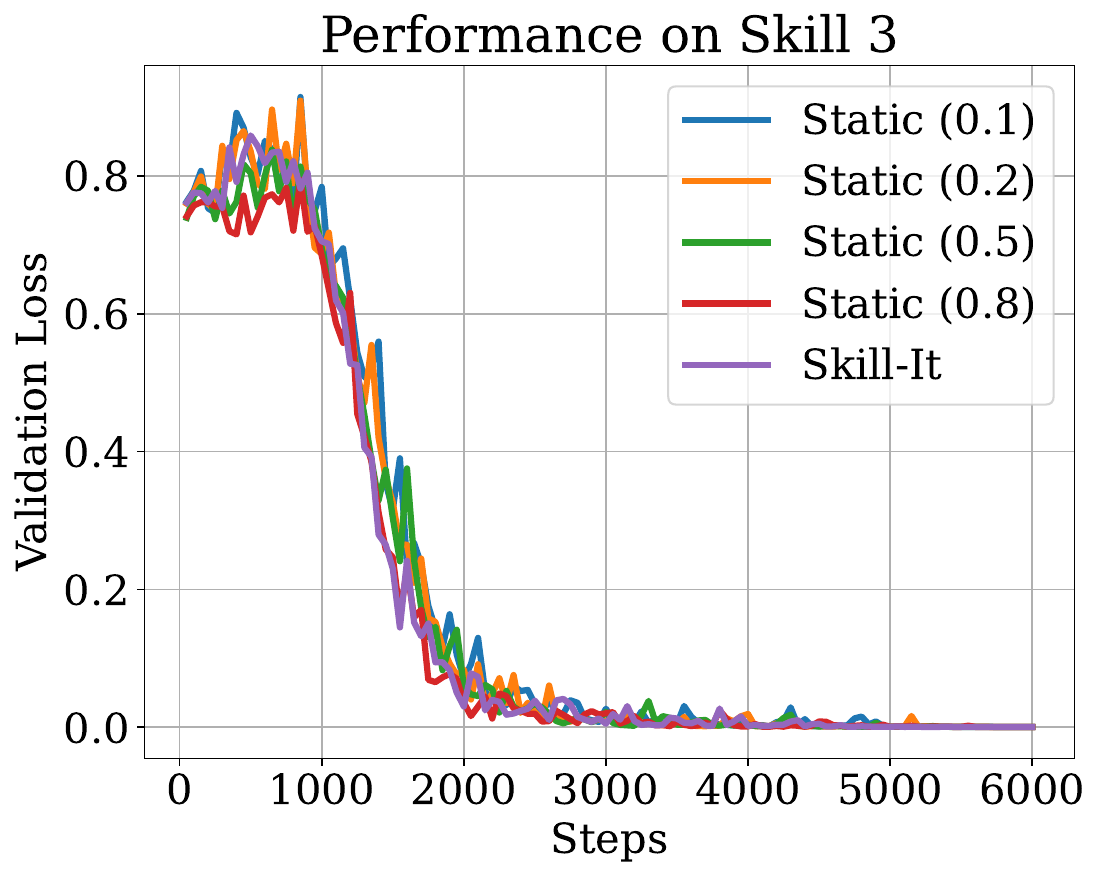}
    \caption{Comparison of \name versus using no graph (left) and static data selection (right) with $\eta = 0.1, 0.2, 0.5, 0.8$ on the LEGO fine-tuning experiment. All approaches have roughly the same loss trajectories, but \name is slightly lower than using no graph.}
    \label{fig:lego_one_skill_ablation}
\end{figure}

Figure~\ref{fig:lego_one_skill_ablation} compares \name, no graph, and static data selection for the LEGO fine-tuning experiment. No graph can be interpreted as allocating equal weight to all training skills not equal to the target skill, and varying this weight versus the weight on the target skill. While \name and setting $T = 1$ behave similarly, we see that \name is slightly better than using no graph. For instance, \name obtains a validation loss of $0.05$ in $2000$ steps, compared to $2050$-$2200$ steps when using no graph.

\begin{figure}
    \centering
    \includegraphics[height=2in]{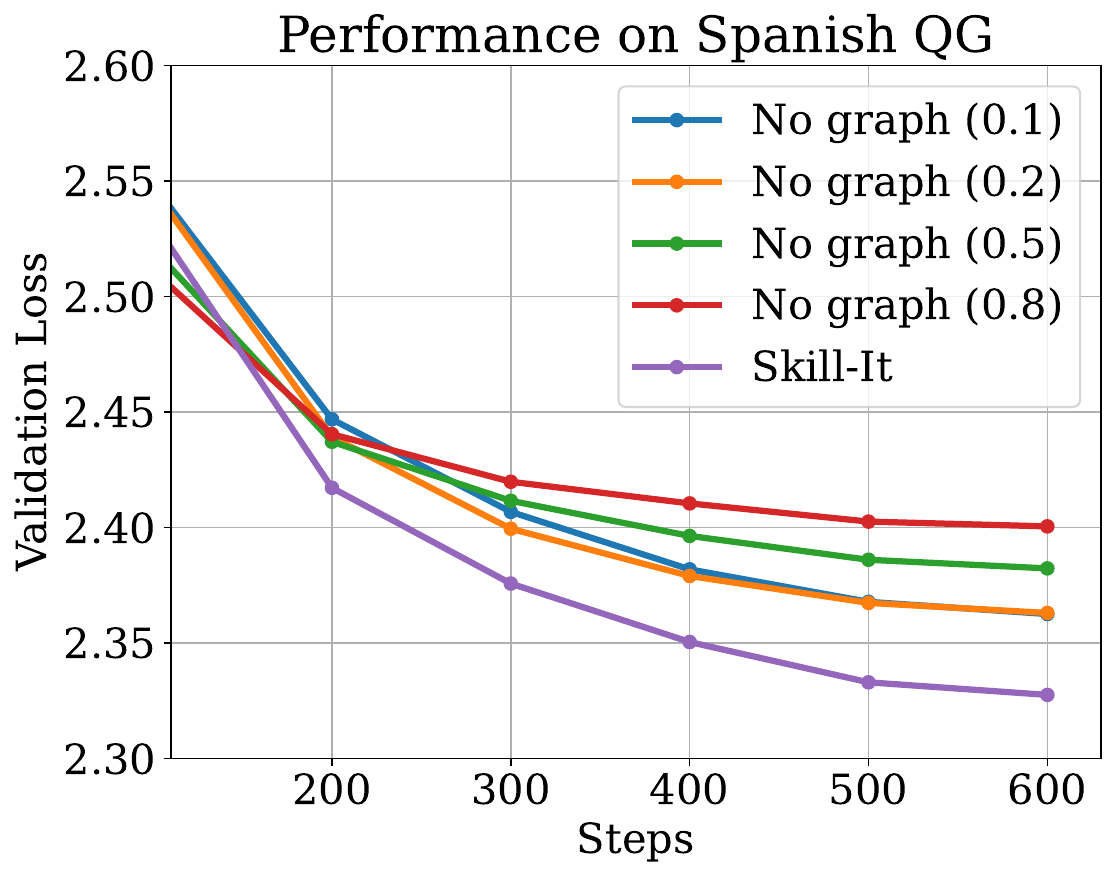}
    \includegraphics[height=2in]{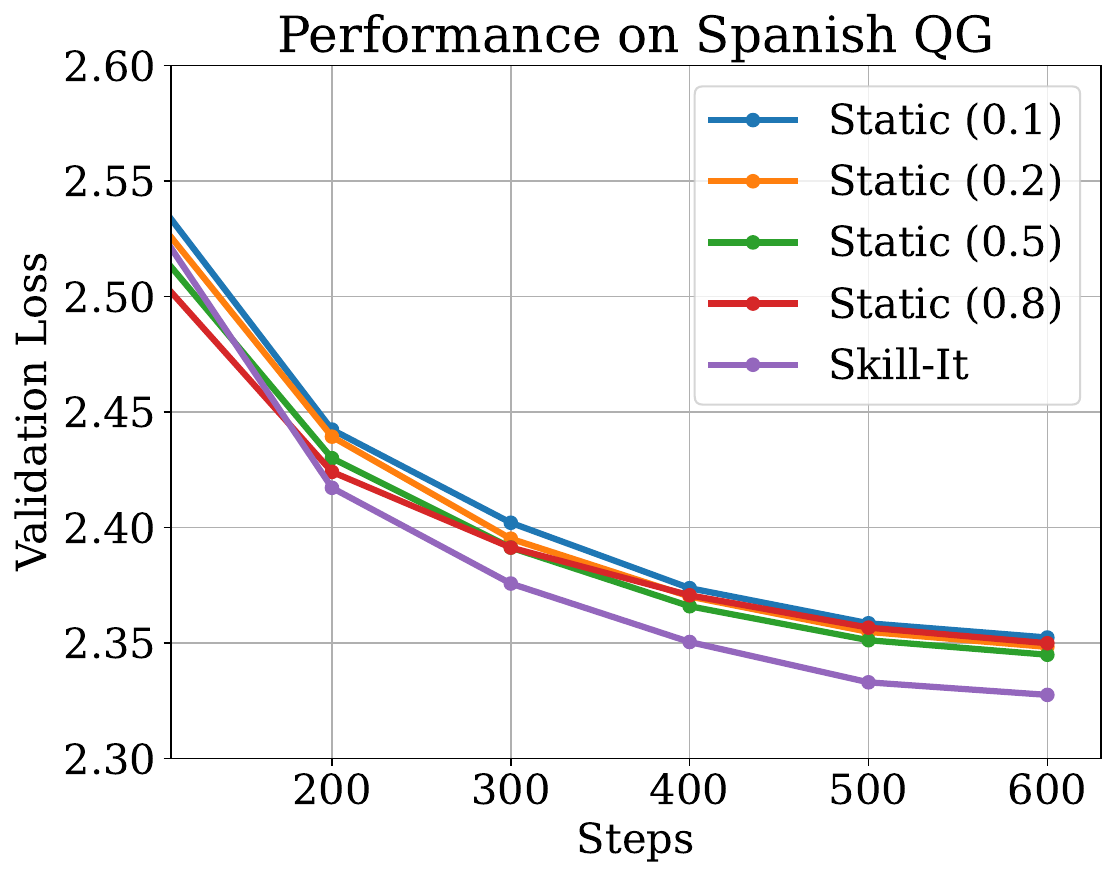}
    \caption{Comparison of \name versus using no graph (left) and static data selection (right) with $\eta = 0.1, 0.2, 0.5, 0.8$ on the Natural Instructions Spanish QG fine-tuning experiment. \name attains lower validation loss than both no graph and static data selection.}
    \label{fig:spanish_qg_ablation}
\end{figure}

\begin{figure}
    \centering
    \includegraphics[height=2in]{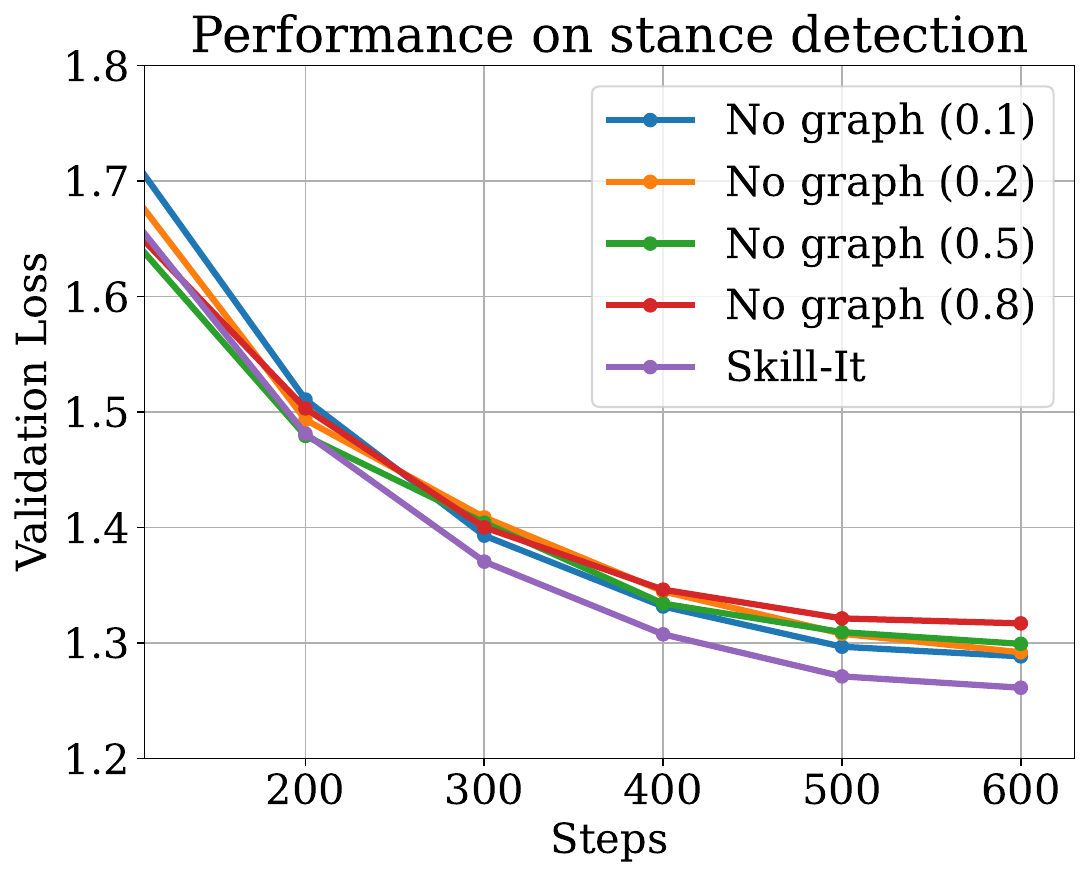}
    \includegraphics[height=2in]{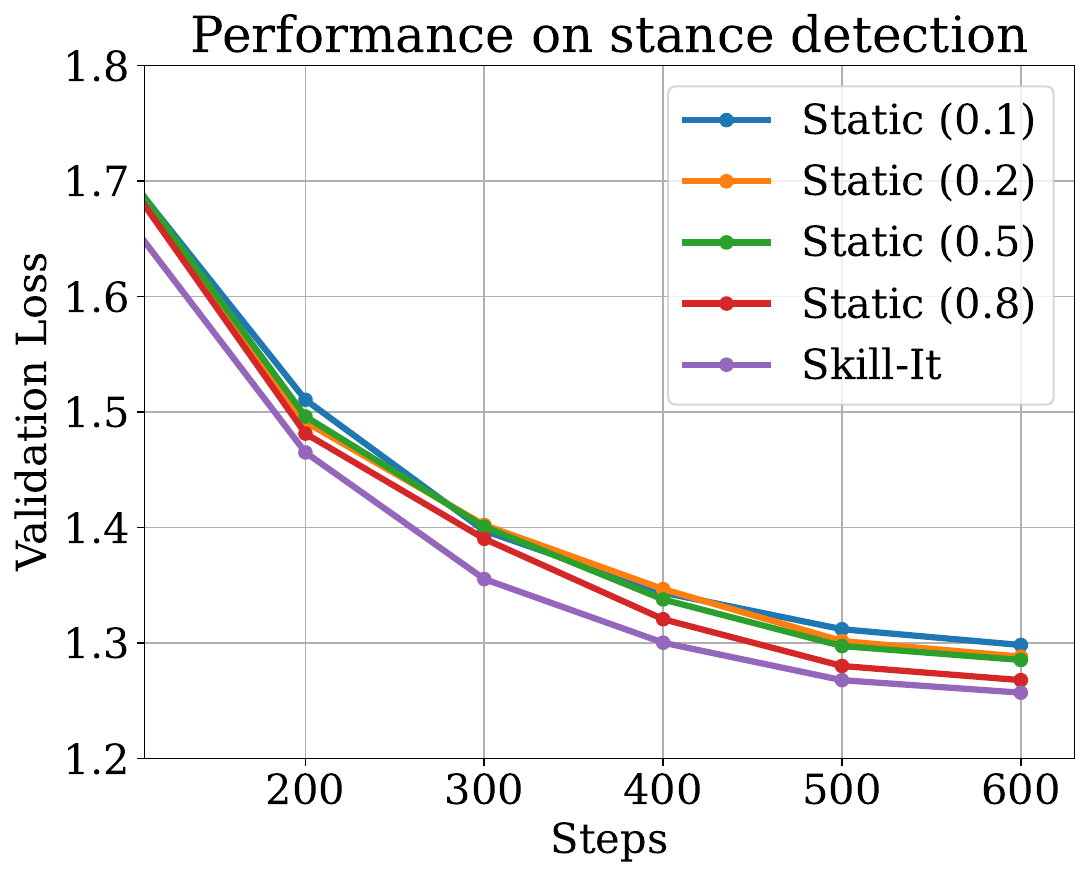}
    \caption{Comparison of \name versus using no graph (left) and static data selection (right) with $\eta = 0.1, 0.2, 0.5, 0.8$ on the Natural Instructions stance detection fine-tuning experiment. \name attains lower validation loss than both no graph and static data selection.}
    \label{fig:stance_detection_ablation}
\end{figure}

Figure~\ref{fig:spanish_qg_ablation} and~\ref{fig:stance_detection_ablation} compare \name, no graph, and static data selection for the Natural Instructions fine-tuning experiments. For both Spanish QG and stance detection, \name attains lower loss than using no graph or using $T = 1$ round.

\begin{figure}
    \centering
    \centerline{\includegraphics[width=6in]{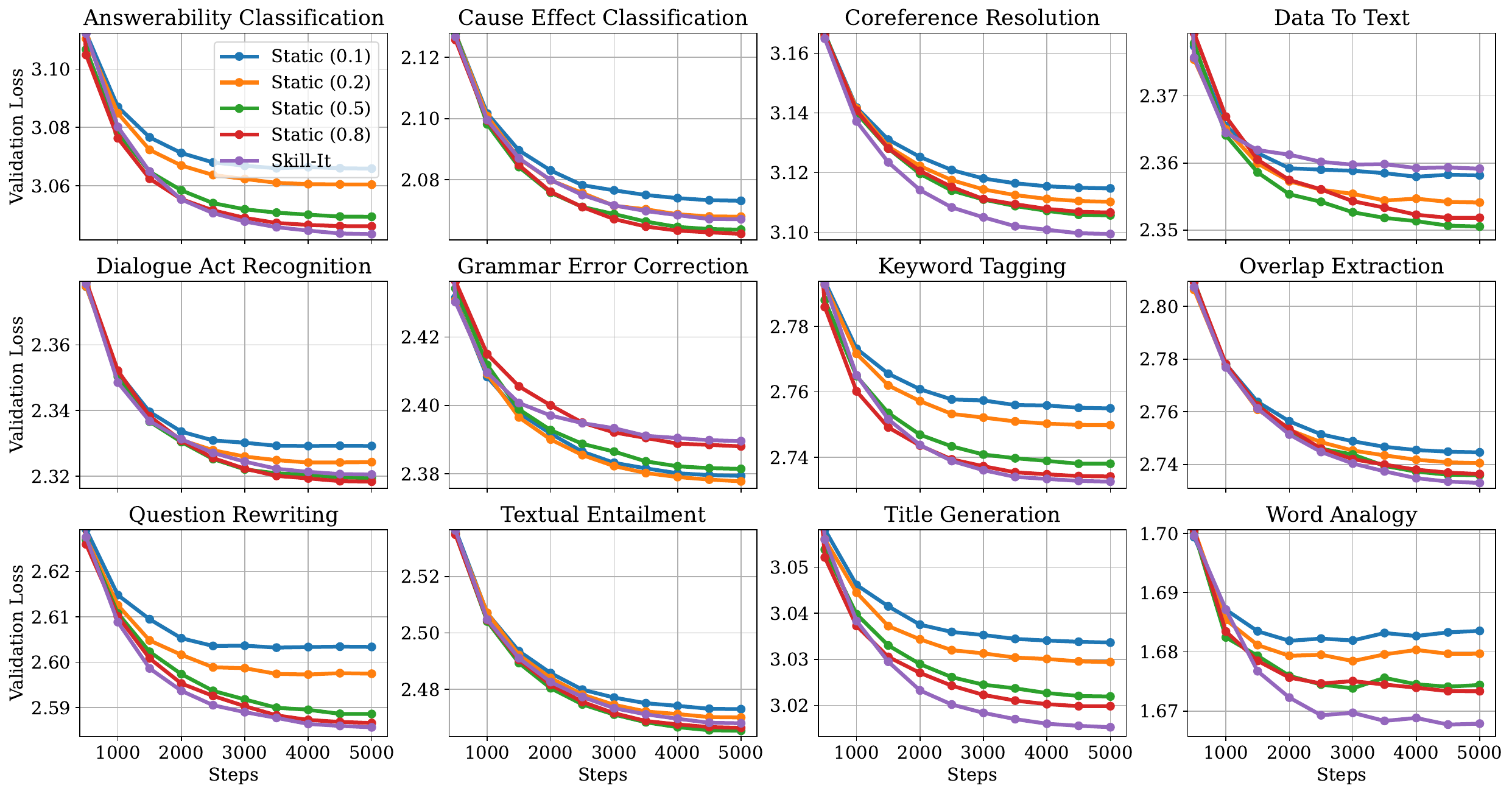}}
    \caption{Comparison of \name versus using static data selection with $\eta = 0.1, 0.2, 0.5, 0.8$ on the Natural Instructions out-of-domain experiment. \name attains the lowest validation loss on $7$ out of $12$ evaluation skills, and an average loss of $2.540$ compared to a range of $2.541$-$2.551$ for static data selection.}
    \label{fig:ni_test_ablation}
\end{figure}

Figure~\ref{fig:ni_test_ablation} compares \name and static data selection for the Natural Instructions out-of-domain experiment. \name attains the lowest validation loss on $7$ out of $12$ evaluation skills. It has an average loss of $2.540$ compared to a range of $2.541$-$2.551$ for static data selection.

\end{document}